\documentclass[preprint,12pt]{elsarticle}

\usepackage{times}
\usepackage{epsfig}
\usepackage{graphicx}
\usepackage{amsmath}
\usepackage{amssymb}
\usepackage{booktabs,siunitx}
\usepackage{subfig}
\usepackage{graphicx}
\usepackage{algorithm}
\usepackage{algpseudocode}
\usepackage{bm}
\usepackage{multirow}
\usepackage{makecell}
\usepackage[table]{xcolor}
\usepackage[normalem]{ulem}
\usepackage{color}
\newcommand{\vf}{\mathbf}
\newcommand{\revision}[1]{{\color{black}{#1}}}
\let\svthefootnote\thefootnote
\newcommand\freefootnote[1]{%
  \let\thefootnote\relax%
  \footnotetext{#1}%
  \let\thefootnote\svthefootnote%
}

\journal{Neural Networks}

\begin{document}

\begin{frontmatter}

\title{Beyond Simple Meta-Learning: Multi-Purpose Models for Multi-Domain, Active and Continual Few-Shot Learning}

\author[]{Peyman Bateni$^{a, 1}$}
\author[]{Jarred Barber$^{h, 2}$}
\author[]{Raghav Goyal$^{b, d}$}
\author[]{Vaden Masrani$^{b}$}
\author[]{Jan-Willem van de Meent$^{e, g}$}
\author[]{Leonid Sigal$^{a, b, d, i}$}
\author[]{Frank Wood$^{b, d, f, i}$\freefootnote{$^{a}$Beam AI $^{b}$University of British Columbia $^{c}$MILA $^{d}$Vector Institute $^{e}$University of Amsterdam $^{f}$Inverted AI $^{g}$Northeastern University $^{h}$Google Research $^{h}$CIFAR}\freefootnote{$^{1}$ Work completed while at the University of British Columbia and Inverted AI}\freefootnote{$^{2}$ Work completed while at Amazon}}

\begin{abstract}
Modern deep learning requires large-scale extensively labelled datasets for training. Few-shot learning aims to alleviate this issue by learning effectively from few labelled examples. In previously proposed few-shot visual classifiers, it is assumed that the feature manifold, 
where classifier decisions are made, has uncorrelated feature dimensions and uniform feature variance. In this work, we focus on addressing the limitations arising from this assumption by proposing a variance-sensitive class of models that operate in a low-label regime. The first method, Simple CNAPS, employs a hierarchically regularized Mahalanobis-distance based classifier combined with a neural adaptive feature extractor to achieve strong performance on Meta-Dataset, mini-ImageNet and tiered-ImageNet benchmarks. We further extend this approach to a transductive learning setting, proposing Transductive CNAPS. This transductive method combines a soft k-means parameter refinement procedure with a two-step task encoder to achieve improved test-time classification accuracy using unlabelled data. Transductive CNAPS achieves improved performance on Meta-Dataset. Finally, we explore the use of our methods (Simple and Transductive) for ``out of the box'' continual and active learning. Extensive experiments on large scale benchmarks illustrate robustness and versatility of this, relatively speaking, simple class of models. All trained model checkpoints and corresponding source codes have been made publicly available at {\tt github.com/plai-group/simple-cnaps}.
\end{abstract}

\begin{graphicalabstract}
\end{graphicalabstract}

\begin{highlights}
\item We describe ``Simple CNAPS'', a neural adaptive few-shot classifier with a regularized Mahalanobis based classifier that achieves strong performance on Meta-Dataset, mini- and tiered-ImageNet benchmarks.
\item We expand our work to transductive few-shot learning by developing ``Transductive CNAPS'', which extends our first method with a two-step transductive set-encoder and a soft-k means iterative procedure for refinement of class parameters. This method achieves further improved performance on Meta-Dataset, mini- and tiered-ImageNet benchmarks.
\item In addition to grounding our work in probabilistic mixture models, we provide an extensive discussion and interpretation of both methods as Riemannian metric learners. This analysis is consistent with our empirical results. 
\item We evaluate both methods on ``out of the box'' active learning using 3 label acquisition strategies. We also modify Simple and Transductive CNAPS for ``out of the box'' continual learning, proposing 3 approaches to continual estimation of the task-encoding that achieve competitive performance. 
\item Finally, through extensive experiments / ablations, we study importance of our design and architectural choices.
\end{highlights}


\end{frontmatter}


\section{Introduction}
\label{sec:introduction}

Deep neural networks have facilitated transformative advances in machine learning \cite{DBLP:journals/corr/abs-1907-09408-object-detection-survey,8441512-image-classification-survey, Hossain:2019:CSD:3303862.3295748-image-captioning-survey, guz-etal-2020-neural, Krizhevsky12_AlexNet, He15_ResNet, Redmon16_YOLO, Ren15_FasterRCNN, Goodfellow2014_GANS, Scibior2021_ITRA, bateni2022realtime}. However, much of their success depends on training using large-scale extensively-labelled datasets, and when an exhaustive set of training examples is not available, performance degrades significantly. \revision{Few-shot learning \cite{feyjie2020semisupervised-medical, DBLP:journals/corr/abs-1904-05046-survey-on-few-shot-learning, Wang:2019:SZL:3306498.3293318-survey-of-zero-shot-learning, bellet2013survey} seeks to address this issue by developing architectures for learning effectively from few labelled instances.} For a novel task, a few-shot visual classifier is derived using a small number of labelled ``support'' images per category, and is then evaluated on a set of unlabelled ``query'' images.

\begin{figure}
    \centering
    \subfloat[Euclidean]{{\includegraphics[width=0.33\textwidth]{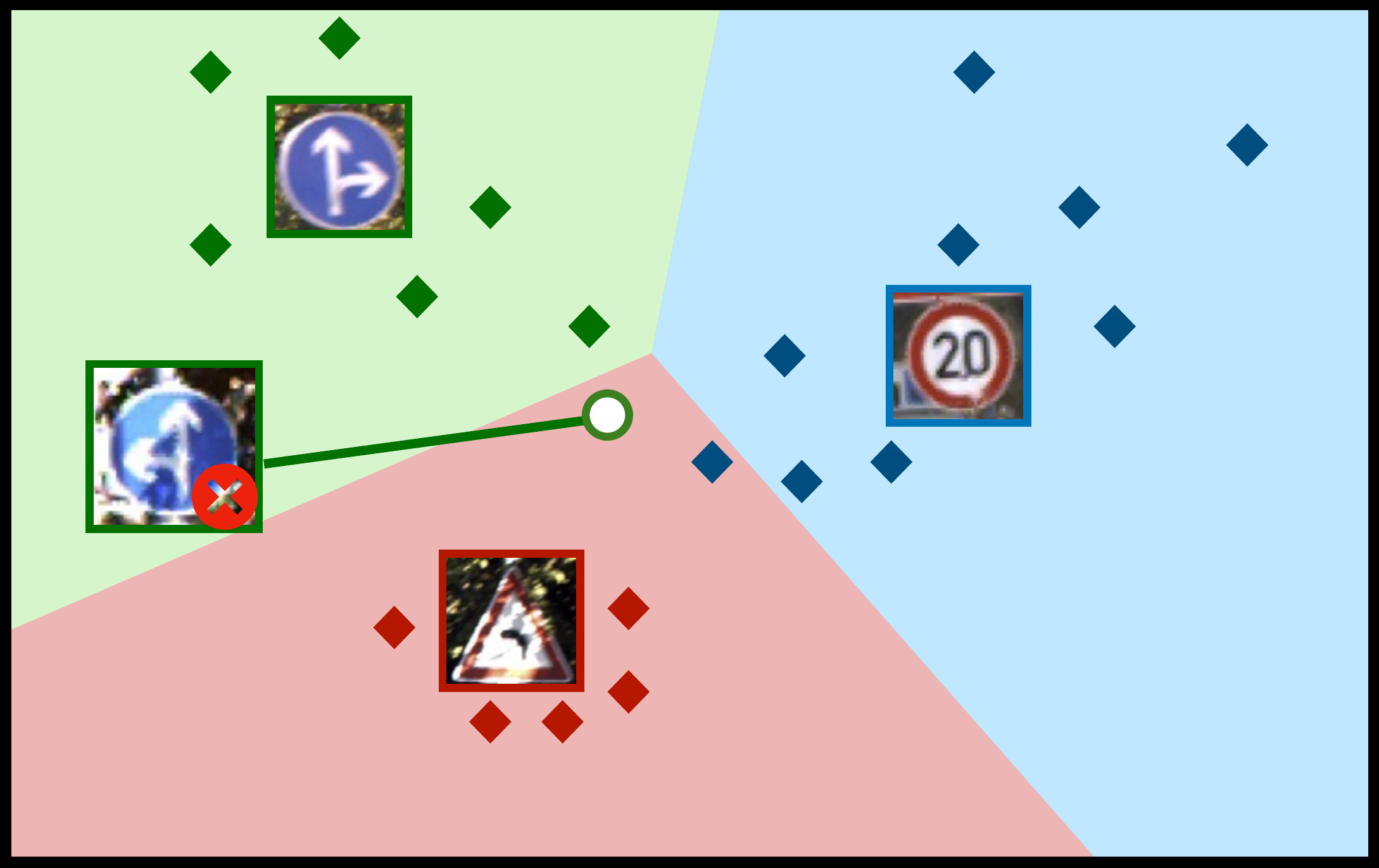} }}
    \subfloat[Mahalanobis]{{\includegraphics[width=0.33\textwidth]{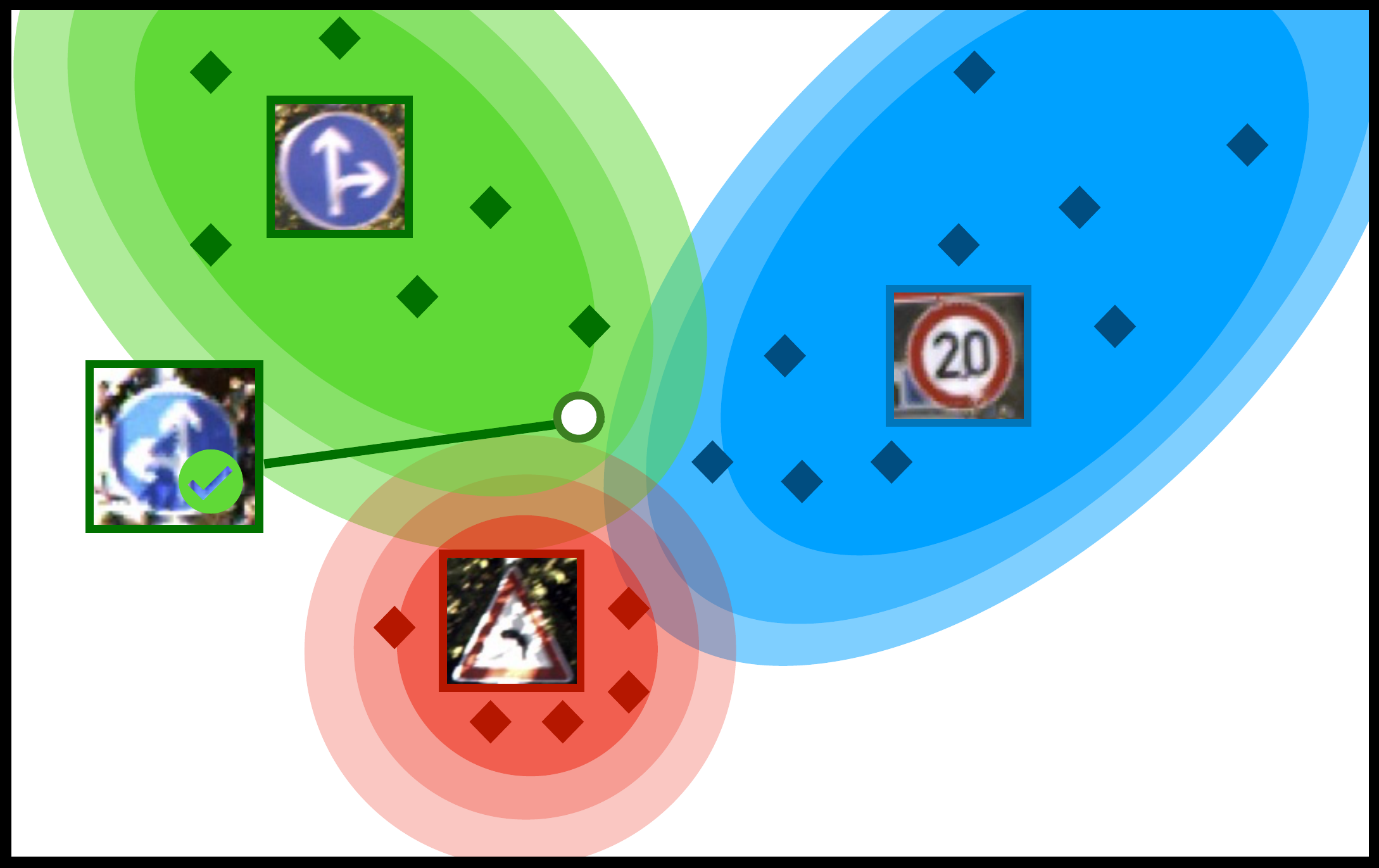} }}
    \subfloat[Transductive]{{\includegraphics[width=0.33\textwidth]{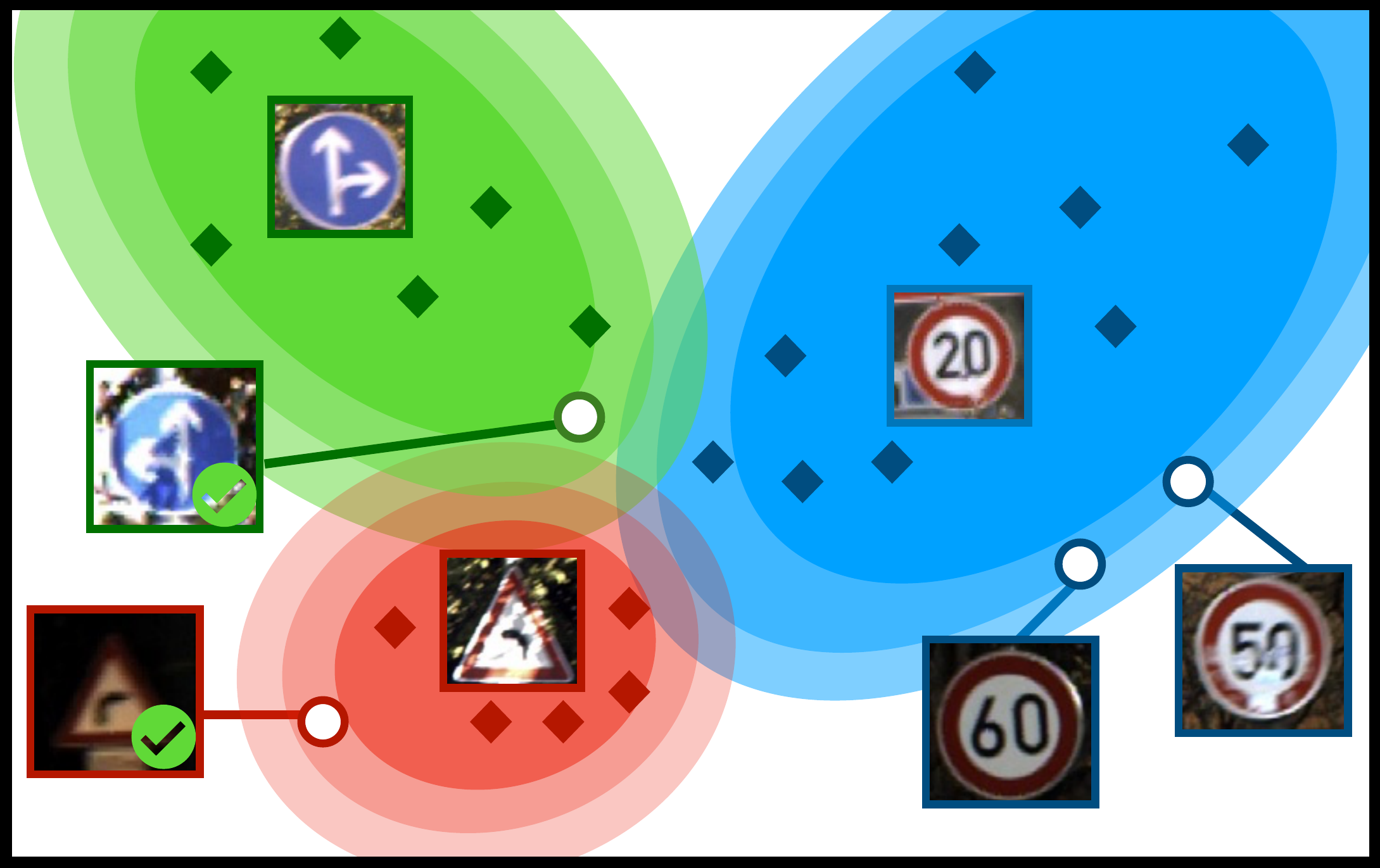} }} \\
    \subfloat[Active Learning]{{\includegraphics[width=\textwidth]{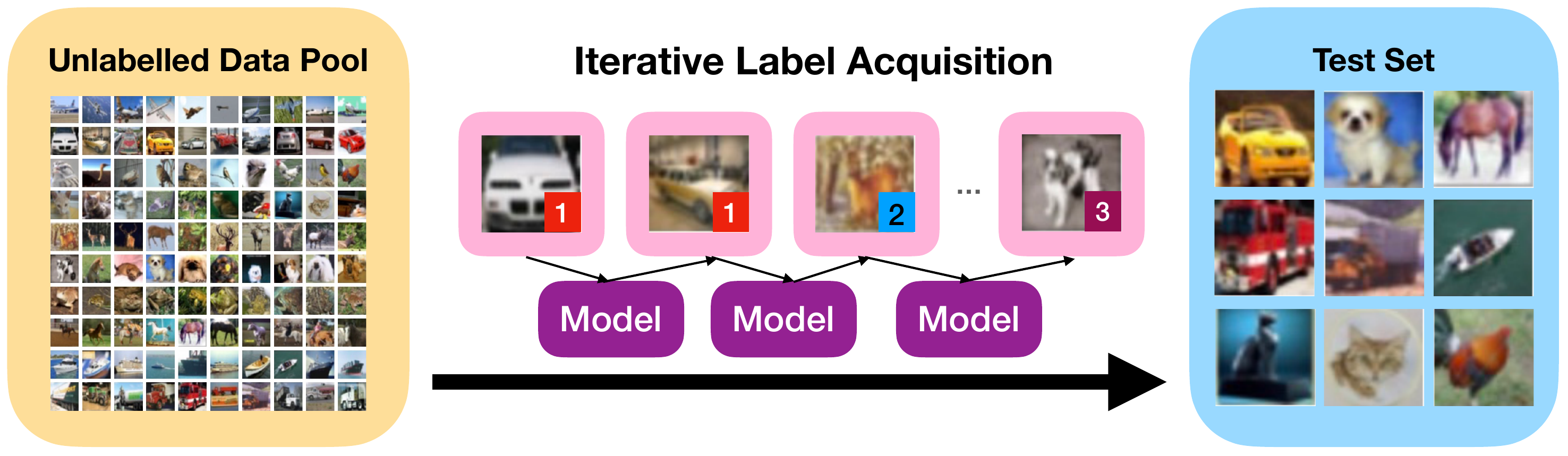} }} \\
    \subfloat[Continual Learning]{{\includegraphics[width=\textwidth]{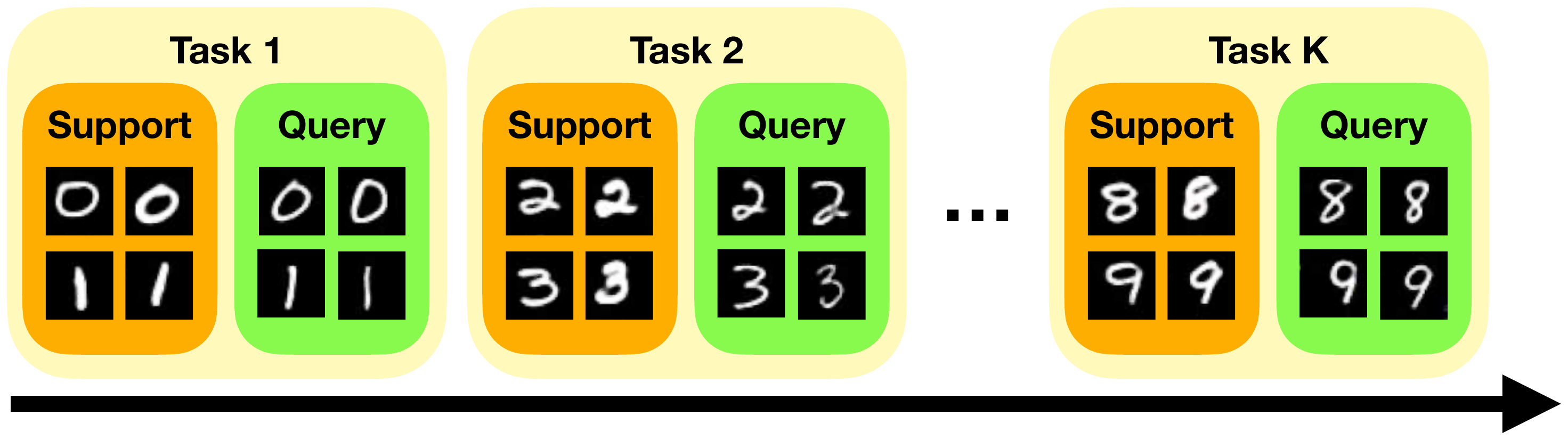} }} \\
    \caption{Two-dimensional illustration of task-adapted support image features are shown in the top row. When using Euclidean distance (a) as the metric, each cluster is assumed to have identity covariance, leading to the incorrect classification of query examples. Mahalanobis distance (b) resolves this issue. In the transductive setting (c), all query examples are labelled all at once, permitting semi-supervised refinement of clusters. During active learning (d), methods can iteratively request labels from a pool of unlabelled examples. In continual learning (e), models see a new task at each iteration, aiming to perform well without forgetting previous tasks.}
\end{figure}

Existing few-shot classification approaches can be divided into two categories. First, there are approaches based on nearest-neighbour classification \cite{vinyals2016matching}, either directly in the feature space \cite{DBLP:journals/corr/abs-1905-01436-edge-labelling-gnn,koch2015siamese, garcia2018fewshot} or the mapping of the feature vectors onto the natural language space \cite{Frome-NIPS2013_5204}. The second group consists of methods that distill examples into class-wise vector prototypes that are either learned \cite{gidaris2019generating, Requeima19_CNAPS} or mathematically derived by mean-pooling the examples \cite{Snell17_Proto}. The prototypes are usually defined in the feature or the natural language space (e.g. word2vec \cite{DBLP:journals/corr/abs-1902-07104-elementai}).  Most research in this area has focused on learning non-linear mappings (typically in the form of deep multi-layer neural networks) from the images \revision{to a high-dimensional vector space, namely the embedding space.} A pre-specified metric within the embedding space is then used for final nearest-class classification (such as the cosine similarity between the feature embedding of the query image and the class embedding). Significant work has focused on effective task-conditioned adaptation of these non-linear mappings at different levels of granularity \cite{finn2017model, Requeima19_CNAPS, DBLP:conf/iclr/RaviL17-meta-lstm, DBLP:journals/corr/abs-1803-02999-reptile}. Recently, Conditional Neural Adaptive Processes (CNAPS) \cite{Requeima19_CNAPS} achieved high few-shot visual classification accuracy through the use of sparse FiLM \cite{perez2018film} layers for partial network adaptation to prevent over-fitting problems that arise from adapting the entire embedding neural network using few support examples.

Overall, far less attention has been given to the metric used to compute distances for classification within the embedding space. Snell et al. \cite{Snell17_Proto} study the underlying distance function in order to justify the use of sample means as prototypes. They argue that Bregman divergences \cite{banerjee2005clustering} are the theoretically sound family of metrics to be used in the few-shot setting, but only utilize a single instance within this family — the squared Euclidean distance, which they find to perform better than the more traditional cosine metric. However, the choice of the Euclidean metric involves making two assumptions: 1) that feature dimensions are un-correlated and 2) that they have uniform variance. In addition, the Euclidean distance is insensitive to the distribution of within-class samples with respect to their prototype and recent results \cite{NIPS2018_7352-tadam,Snell17_Proto} suggest that this is problematic. Modelling this distribution (in the case of \cite{banerjee2005clustering} using extreme value theory) is, as we observe, a key to achieving better accuracy.

Starting from this intuition, we develop our first method, the ``Simple CNAPS'' architecture that achieves a 6.1\% improvement over CNAPS \cite{Requeima19_CNAPS} while removing 788,485 parameters (3.2\% of the total) from the original CNAPS architecture, replacing them with fixed, closed-form, deterministic covariance estimation and Mahalanobis distance computations. We find that surprisingly, we are able to generate useful high-dimensional estimates of covariance even in the few-shot classification setting, where the number of available support examples per class is in theory far too small to estimate the required class-specific covariances.

In a standard few-shot learning setting, the classifier is adapted using labelled examples in the support set. Performance can be improved further by exploiting additional unlabelled support data (semi-supervised few-shot learning) \cite{DBLP:journals/corr/abs-1803-00676-tieredimagenet}, or examples in the query set (transductive few-shot learning) \cite{DBLP:journals/corr/abs-1805-10002-tpn, DBLP:journals/corr/abs-1905-01436-edge-labelling-gnn}. Existing transductive few-shot methods reason about unlabelled examples by performing k-means clustering with Euclidean distance \cite{DBLP:journals/corr/abs-1803-00676-tieredimagenet} or message passing in graph convolutional networks \cite{DBLP:journals/corr/abs-1805-10002-tpn, DBLP:journals/corr/abs-1905-01436-edge-labelling-gnn}. These methods, while improving performance, lack the expressibility of the Mahalanobis-distance based classification framework used in Simple CNAPS. Furthermore, they primarily focus on tranductive adaptation within the classification space, disregarding tranductive adaptation of the feature extractor all together.

\begin{figure}[t]
    \centering
    \includegraphics[width=\textwidth]{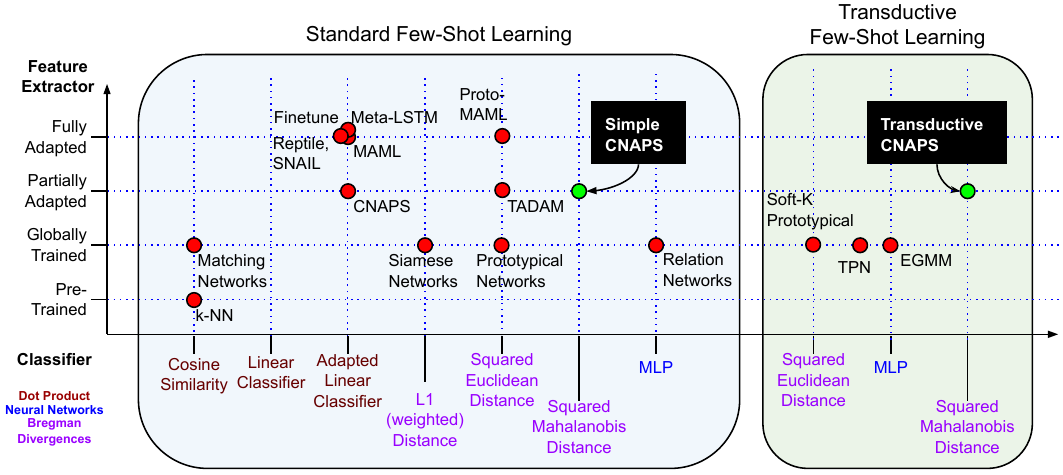}
    \caption{Overview of research on few-shot image classification, organized by image feature extractor adaptation scheme (\emph{vertical axis}) versus final classification methodology (\emph{horizontal axis}).}
    \label{fig:related-work-overview}
\end{figure}

Motivated by these observations, we develop a transductive variant of the Simple CNAPS architecture. In this variant, we infer labels for query set examples, which allows us to make use of these examples in class parameter estimates. The resulting architecture, namely ``Transductive CNAPS'', extends Simple CNAPS with a transductive two-step task-encoder, and an iterative soft k-means procedure for refining class parameter estimates (mean and covariance). We demonstrate that this approach achieves high accuracy on Meta-Dataset \cite{triantafillou2019meta}. \revision{In addition, Transductive CNAPS achieves notable performance on mini-ImageNet \cite{Snell17_Proto} and tiered-ImageNet \cite{DBLP:journals/corr/abs-1803-00676-tieredimagenet} benchmarks.}

\revision{Furthermore, Requeima et al. \cite{Requeima19_CNAPS} recently proposed to evaluate pre-trained few-shot classifiers for active learning and continual learning ``out of the box" without any additional problem specific fine-tuning. Following their work, we also explore Simple and Transductive CNAPS \cite{bateni2019improved,  bateni2021improving, Bateni20_TransductiveCNAPS, Bateni2021_Thesis} in the context of ``out of the box" active learning and continual learning.} In the active learning setting, a meta-trained few-shot classifier is presented with a set of unlabelled examples, from which it can acquire the label for a single example at each iteration. The goal is to select the examples to maximally improve performance on a separate set of test examples. We evaluate both methods without any additional training specific to active learning, and demonstrate that uncertainty-driven selection methodologies outperform random selection in our models, thus demonstrating their abilities in producing effective measures of uncertainty.

In the continual learning problem domain, a few-shot learning method is presented with a new classification task at each iteration. The goal is to continuously learn to classify accurately on these new tasks while maintaining good accuracy on the previous tasks. Continual learning requires this to be achieved without explicitly saving every data point to memory. In our work, we propose and explore three continual learning strategies within the framework of our methods. 

\begin{figure*}[!ht]
    \centering
    \includegraphics[width=\textwidth]{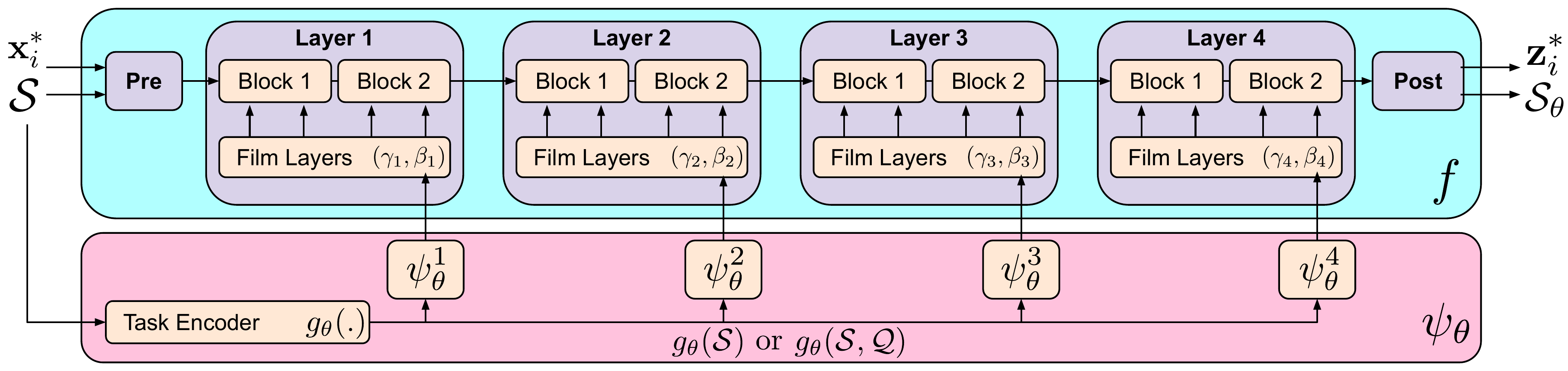}
    \caption{Overview of neural adaptive feature extraction in Simple, Transductive and "original" CNAPS. Figure adapted from \cite{Bateni20_TransductiveCNAPS}.}
    \label{fig:feature-extraction-procedure-overview}
\end{figure*}

\revision{
\vspace{0.075in}
\noindent
\textbf{Our contributions:} \textbf{(1)} We describe ``Simple CNAPS'', a neural adaptive few-shot classifier with a regularized Mahalanobis based classifier that achieves strong performance on Meta-Dataset. \textbf{(2)} We expand our work to transductive few-shot learning by developing ``Transductive CNAPS'', which extends our first method with a two-step transductive set-encoder and a soft-k means iterative procedure for refinement of class parameters. This method achieves further improved performance on Meta-Dataset. \textbf{(3)} In addition to grounding our work in probabilistic mixture models, we provide an extensive discussion and interpretation of both architectures as Riemannian metric learners. This analysis is consistent with our empirical results. \textbf{(4)} We evaluate both methods on ``out of the box'' active learning using three active label acquisition strategies. We also modify Simple and Transductive CNAPS for ``out of the box'' continual learning, proposing three approaches to continual estimation of the task-encoding that achieve competitive performance. 
\textbf{(5)} Finally, through extensive experiments / ablations, we study importance of our design choices.}

\section{Related Work}

\subsection{Few-Shot Learning with Labelled Data}

Past research on few-shot classification \cite{DBLP:journals/corr/abs-1904-05046-survey-on-few-shot-learning} can be differentiated along two major axes: 1) how images are transformed into vectorized embeddings, and 2) how ``distances'' are computed between vectors in order to assign labels. This is illustrated in Figure~\ref{fig:related-work-overview}.

Siamese networks \cite{koch2015siamese}, an early approach to few-shot learning, employed a shared feature extractor to produce vector embeddings for both the support and the query images. Classification was then done by picking the smallest weighted L1 distance between query and labelled image embeddings.
Relation networks \cite{sung2018learning}, and recent GCNN variants \cite{DBLP:journals/corr/abs-1905-01436-edge-labelling-gnn, garcia2018fewshot}, extended this by parameterizing and learning the classification metric using a Multi-Layer Perceptron (MLP).  Matching networks \cite{vinyals2016matching} learned distinct feature extractors for support and query images which were then used to compute cosine similarities for classification.

The feature extractors used by these models were, notably, not adapted to test-time classification tasks. It has become established that adaptation of the feature extractor to new tasks at test time is generally a good thing to do.  Fine tuning transfer-learned networks \cite{DBLP:journals/corr/YosinskiCBL14-finetune} did this by fine-tuning the feature extraction network using the task-specific support images but found limited success due to problems related to overfitting to, the generally very few, support examples. MAML \cite{finn2017model} (and its various extensions \cite{ DBLP:journals/corr/MishraRCA17-snail, DBLP:journals/corr/abs-1803-02999-reptile,DBLP:conf/iclr/RaviL17-meta-lstm}) mitigated this problem by learning a set of meta-parameters that specifically enabled feature extractors to be adapted to new tasks given few support examples using few gradient descent steps.

The two supervised few-shot learning approaches most directly related to this work are CNAPS \cite{Requeima19_CNAPS} (and the related TADAM \cite{NIPS2018_7352-tadam}) and Prototypical Networks \cite{Snell17_Proto}. CNAPS is a few-shot adaptive visual classifier based on conditional neural processes (CNPs) \cite{DBLP:journals/corr/abs-1807-01613-cnp}. It is a high-performing approach for few-shot image classification \cite{Requeima19_CNAPS} that uses a pre-trained feature extractor augmented with  FiLM layers \cite{perez2018film} that are adapted for each task using the support images specific to that task. CNAPS uses a dot-product distance in a final linear classifier; the parameters of which are also adapted at test-time to each new task (see Section \ref{sec:cnaps}).

Prototypical Networks \cite{Snell17_Proto} employ a non-adaptive feature extractor with a simple mean-pooling operation to form class ``prototypes'' from support instances in a meta-learned feature space. They employ episodic training to learn said feature extractor, where at each training iteration, a few-shot task is provided with a ``support'' set for adaptation and a ``query'' set for evaluation and calculating a classification loss that is then used for gradient descent. Squared Euclidean distances to these prototypes are calculated and subsequently used for nearest-class classification. Their choice of the distance metric was motivated by the theoretical properties of Bregman divergences \cite{banerjee2005clustering}, a family of functions of which the squared Euclidean distance is a member of. These properties allow for a correspondence between the use of the squared Euclidean distance in a Softmax classifier and performing density estimation. Expanding on the work of Snell et al.\cite{Snell17_Proto}, we also exploit similar properties of the squared Mahalanobis distance as a Bregman divergence \cite{banerjee2005clustering} to draw theoretical connections to Bregman soft-clustering. We additionally provide an extensive alternative theoretical grounding of both methods in Riemannian metric learning.

Our work differs from CNAPS \cite{Requeima19_CNAPS} and Prototypical Networks \cite{Snell17_Proto} in the following ways. First, while CNAPS has demonstrated the importance of adapting the feature extractor to a specific task, we show that adapting the classifier is actually unnecessary to obtain good performance. Second, we demonstrate that an improved choice of Bregman divergence can significantly impact accuracy. Specifically, we demonstrate that regularized class-specific covariance estimation from task-specific adapted feature vectors allows for use of the Mahalanobis distance for classification, achieving significant improvements in performance.

\revision{
Our methods also relate to recent works \cite{DBLP:journals/corr/abs-1708-02735-fort, Yang21_DC} in modelling per-class clusters with Gaussian means and covariance estimates. Fort et al. \cite{DBLP:journals/corr/abs-1708-02735-fort} extended Prototypical Networks to use the Mahalanobis distance by incorporating class-wise diagonal covariance estimates generated by a learned network. Unlike their work, our methods produce per-class closed-form full covariance estimates within a conditional neural adapted feature space, all trained end-to-end, making them especially effective empirically. Yang et al. \cite{Yang21_DC} produce class-wise Gaussian distributions within a fixed pre-trained feature space. These distributions are first fine-tuned using mixtures of Gaussian statistics from visually similar classes. They are then used to sample new support examples within the feature space, resulting in an augmented support set that is then used to train a logistic regression classifier for the task at hand. Although we similarly regularize the class-wise covariance estimates using the task-level covariance, our methods do not retain any Gaussian statistics about previous tasks/classes, relying solely on the support examples provided. Furthermore, we adapt the underlying feature space through an end-to-end learned adaptation procedure, resulting in empirically useful higher-dimensional mean and covariance estimates from the very few available support examples. Lastly, we use the Mahalanobis distance within the distribution space itself to perform classification, incorporating inter/intra-class variances within the decision boundaries.}

\subsection{Few-Shot Learning with Unlabelled Data}

The use of unlabelled instances for transductive few-shot learning has also been the subject of study by several methods \cite{DBLP:journals/corr/abs-1905-01436-edge-labelling-gnn, DBLP:journals/corr/abs-1805-10002-tpn, DBLP:journals/corr/abs-1803-00676-tieredimagenet}. EGNN \cite{DBLP:journals/corr/abs-1905-01436-edge-labelling-gnn} employs a graph convolutional edge-labelling network for iterative propagation of labels from support to query instances. Similarly, TPN \cite{DBLP:journals/corr/abs-1805-10002-tpn} learns a graph construction module for neural propagation of soft labels between elements of the query set. These methods rely on neural parameterizations of distance in the feature space. TEAM \cite{Qiao_2019_ICCV-team} uses an episode-wise transductively adaptable metric for performing inference on query examples using a task-specific metric. Song et al. \cite{Song_2020_CVPR-cross-attention} use a cross attention network with a transductive iterative approach for augmenting the support set using the query examples.

\begin{figure}
    \centering
    \includegraphics[width=0.95\textwidth]{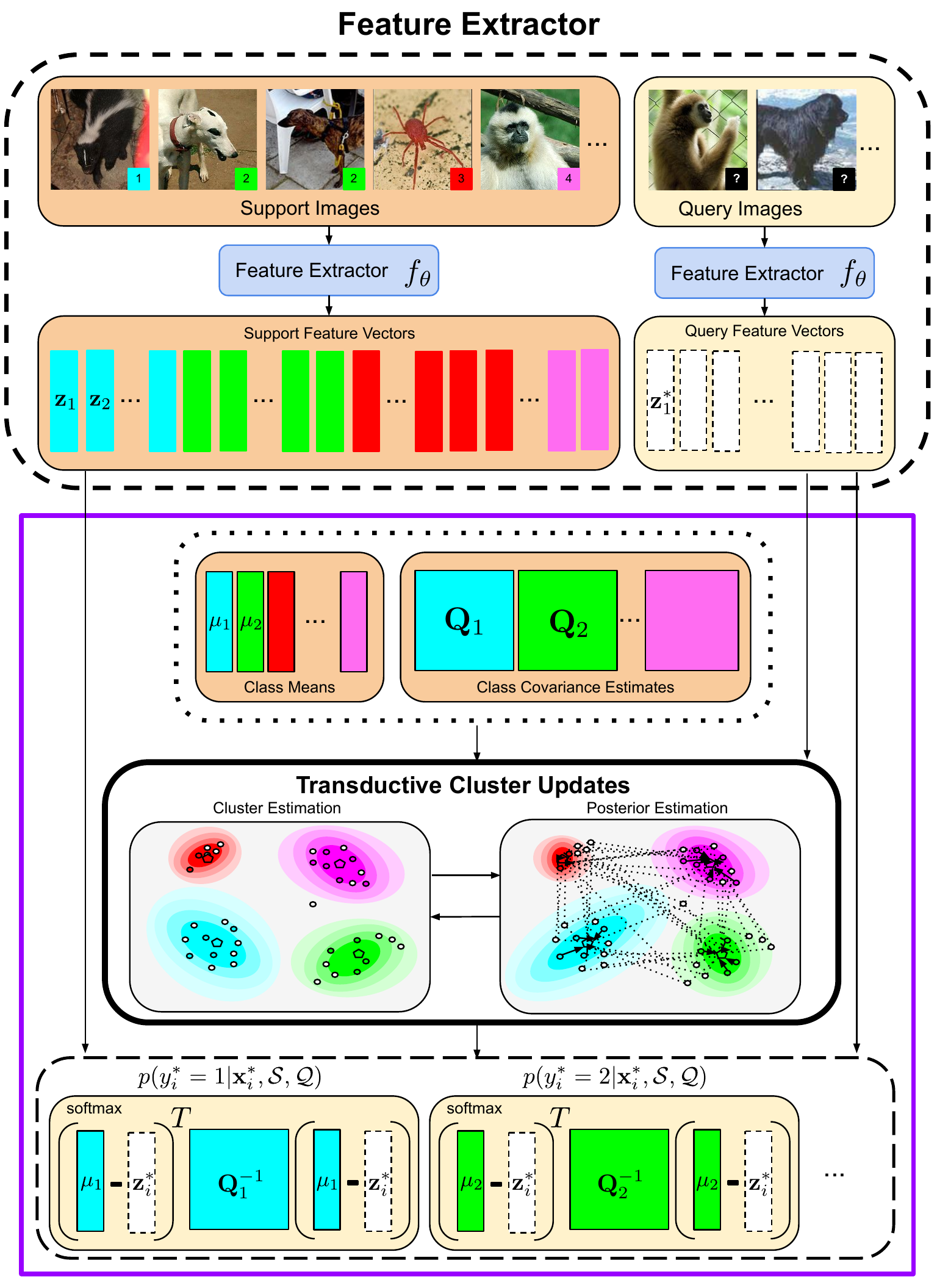}
    \caption{Transductive CNAPS extends the Mahalanobis-distance based classifier in Simple CNAPS through transductive soft k-means clustering of the visual space.}
    \label{fig:transductive-cnaps-vs-simple-cnaps}
\end{figure}

The closest approach to our transductive work, namely Transductive CNAPS, is that of Ren et al. \cite{DBLP:journals/corr/abs-1803-00676-tieredimagenet}. Their method extends Prototypical Networks \cite{Snell17_Proto} through incorporating a single further soft-labelled weighted estimation of class prototypes. Transductive CNAPS, on the other hand, differs in three major ways. First, we produce soft-label estimates of both the mean and covariance. Second, we use an expectation-maximization (EM) inspired algorithm that performs a dynamic number of soft-label updates, depending on the task at hand. Lastly, we employ a neural-adaptive procedure for feature extraction that is conditioned on a two-step learned transductive task representation, as opposed to a fixed feature-extractor. This novel task-representation encoder is responsible for significant performance gains on out-of-domain tasks (Sec. \ref{exp:feot-vs-cot}).

\subsection{Active Learning}

\revision{Active learning is a major research paradigm in machine learning that focuses on requesting labels for unlabelled examples such that performance gain is maximized. Specifically, active learning aims to make data-labelling part of the learning process itself, such that samples are chosen for labelling by the model. 

Existing methods focus on 3 major label selection strategies. First, there are uncertainty-based methods \cite{Beluch2018PowerEnsemblesActive, Joshi2010Multiclassbatch, Lewis1994sequentialalgorithmtraining, Ranganathan2017Deep, Seung1992Querycommittee, Tong2002Supportvectormachine} where sample selection is performed using class-wise probabilities as measure for model uncertainty on said samples. Second are diversity-based methods \cite{Bilgic2009LinkbasedActive, Gal2017DeepBayesianActiveLearningwithImageData, Guo2010ActiveInstanceSampling, Nguyen2004Activelearningusing} where selection is performed such that diversity among categories and their labelled examples is maximized. Third, approaches have also been proposed \cite{Freytag2014SelectingInfluentialExamples, Roy2001OptimalActiveLearning, Settles2007MultipleInstanceActive} to use expected model change as the criterion for sample selection, where unlabelled examples are ranked on the basis of expected parametric change to the model. Dataset-independent sample selection based on uncertainty, where the uncertainty of any unlabelled example is not dependent on the rest of the set, can lead to sampling bias. However, focusing on strategies that promote diversity may result in limited performance gain relative to the number of labels obtained. Motivated by these observations, hybrid instance selection strategies \cite{Ash2019DeepBatchActive, Shui2019DeepActiveLearning, Yin2017DeepSimilarityBased} have been proposed that use mixtures of uncertainty-based, diversity-based and expected-update criteria for active learning, leveraging each of their strengths while minimizing their weaknesses.

Although active learning has been the subject of much study, ``out of the box'' use of few-shot classifiers for active learning has only been explored recently \cite{Requeima19_CNAPS, Pezeshkpour20_ActiveFewShot}. Requeima et al.~demonstrate that CNAPS \cite{Requeima19_CNAPS}, when used for ``out of the box'' active learning, is able to outperform Prototypical Networks \cite{Snell17_Proto}. The margins of  improvement is even greater when uncertainty-based approaches for label acquisition is employed as oppose to random selection. Pezeshkpour et al. \cite{Pezeshkpour20_ActiveFewShot} study the effects of various active few-shot learning strategies within the context of Simple CNAPS and Prototypical Networks, ultimately concluding that better strategies need to be developed for effective active few-shot learning.}

\subsection{Continual Learning}
Continual learning focuses on developing methods that can learn to perform new tasks without ``forgetting'' previous ones. 
In particular, it studies the problem of learning from an infinite stream of tasks, where the goal is to gradually acquire knowledge, using it for future learning without loss of existing knowledge. Most previous work propose different criterions and strategies for gradually updating sections of the learned network as new tasks, and by consequence, labelled examples are seen. Zenke et al.~\cite{Zenke17_ContinualSI} propose Synaptic Intelligence (SI), a network of synapses with complex three-dimensional state-spaces, tracking the past and the current parameter values. Online estimates of each synapse's ``importance'' is employed to consolidate important synapses by preventing them from future learning, while allocating unimportant synapses to learn from future tasks. EWC \cite{Kirkpatrick16_EWC} slows down learning on select parameters based on how important they are to previously seen tasks, thus preserving the learned knowledge. VCL \cite{Nguyen18_VCL} fuses online variational inference (VI) with Monte Carlo VI to train deep discriminative and generative models in complex continual learning settings. Chaudhry et al.~propose RWalk \cite{Chaudhry18_RWalk}, a generalized and more efficient EWC with a theoretically grounded KL-divergence based perspective that achieves superior accuracy on a range of continual learning image classification benchmarks.

CNAPS \cite{Requeima19_CNAPS} presents the first instance of ``out of the box'' use of few-shot image classifiers for continual learning. Here, the few-shot classifier is first trained on Meta-Dataset \cite{triantafillou2019meta} and then used for continual learning without any additional training. They employ context-weighted estimates of class mean for computing class means from old/new tasks. This is performed with current-task only adaptation of the feature space, thus resulting in feature manifolds that differ from each other. We refer to this strategy as the ``Moving Encoding'', and explain it in detail when discussing continual learning in Simple and Transductive CNAPS.

\section{Problem Definition}

\begin{figure}[t]
    \centering
    \includegraphics[width=0.92\textwidth]{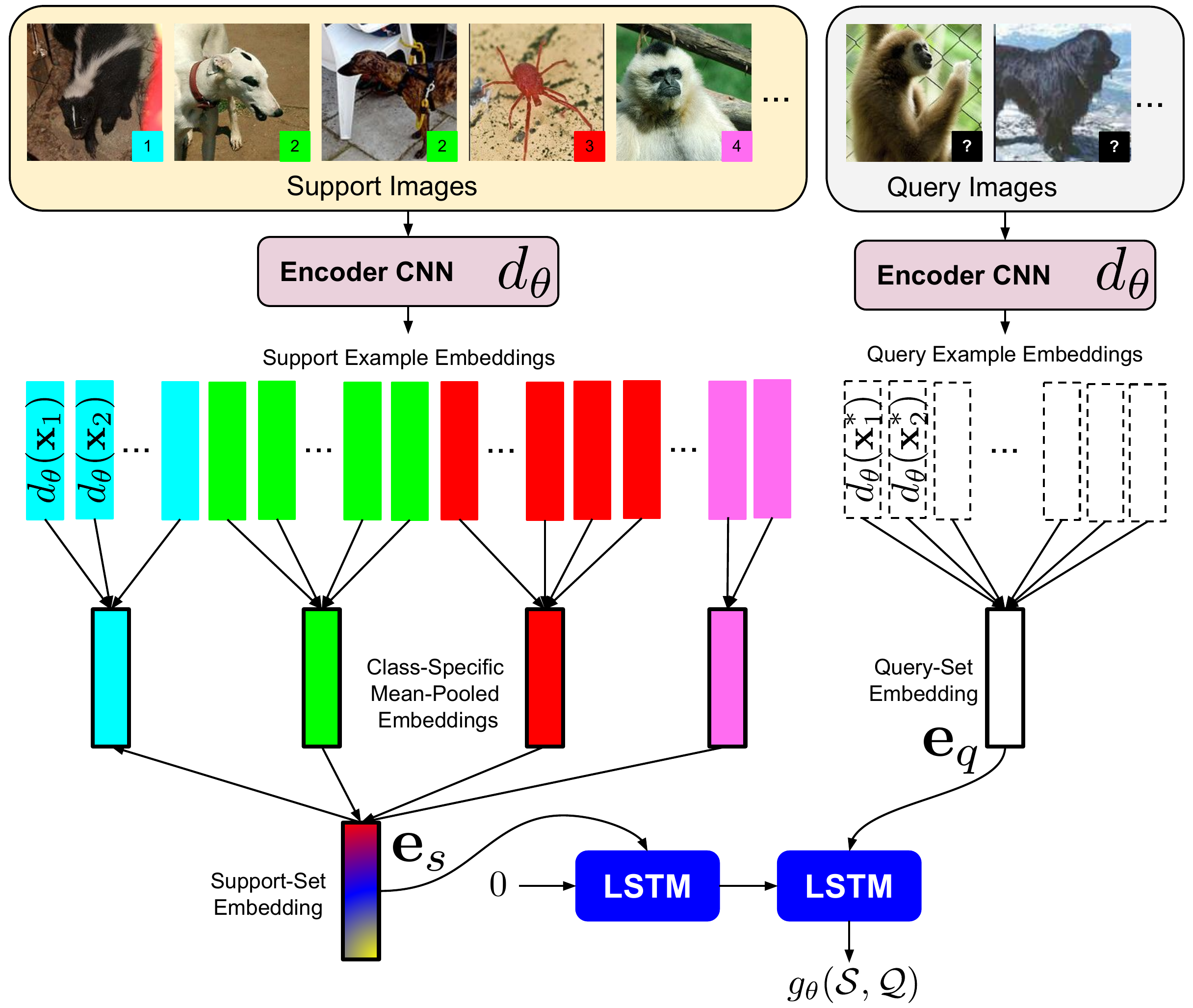}
    \caption{Overview of the transductive task-encoder, $g_\theta(\mathcal{S}, \mathcal{Q})$, used for task-adaptation in Transductive CNAPS.}
    \label{fig:transductive-task-encoder}
\end{figure}

Following previous work \cite{Snell17_Proto, bateni2019improved, Requeima19_CNAPS, finn2017model}, we focus on a few-shot learning setting where a distribution $D$ over image classification tasks $(\mathcal{S}, \mathcal{Q})$ is available for episodic training. Each task $(\mathcal{S}, \mathcal{Q}) \sim D$ consists of a support set $\mathcal S = \{(\mathbf{x}_i, y_i)\}_{i=1}^n$ of labelled images and a query set $\mathcal Q = \{\mathbf{x}_i^*\}_{i=1}^{m}$ of unlabelled images; the objective is to predict labels for these query examples, given the (typically small) support set. Each query image $\mathbf{x}_i^* \in \mathcal Q$ has a corresponding ground truth label $y_i^*$ available at training time. A model will be trained by minimizing, over some parameters $\theta$ (which are shared across tasks), the expected query set classification loss over tasks. We minimize $\mathbb{E}_{(\mathcal{S}, \mathcal{Q}) \sim D}[-\sum_{\mathbf{x}_i^* \in \mathcal Q}\log p_\theta(y^*_i|\mathbf x^*_i, \mathcal{S})]$ for Simple CNAPS and $\mathbb{E}_{(\mathcal{S}, \mathcal{Q}) \sim D}[-\sum_{\mathbf{x}_i^* \in \mathcal Q}\log p_\theta(y^*_i|\mathbf x^*_i, \mathcal{S}, \mathcal{Q})]$ to train Transductive CNAPS. Note that the inclusion of the dependence on all of $\mathcal{Q}$ in the transductive case allows for joint prediction of labels for the query set, all at once. At test time, a separate distribution of tasks generated from previously unseen images and classes is used to evaluate performance. Let us also define \textit{shot} as the number of support examples per class, and \textit{way} as the number of classes within the task.

\section{Methods}

\label{sec:method}

\subsection{CNAPS}
\label{sec:cnaps}
Conditional Neural Adaptive Processes (CNAPS) \cite{Requeima19_CNAPS} consist of two modules: a feature extractor and a linear classifier, which are each task-dependent. Adaptation is performed by meta-trained network adaptation modules that condition on the support set.

The method uses a ResNet18 \cite{He15_ResNet} architecture (Figure \ref{fig:feature-extraction-procedure-overview}) as the feature extractor. This ResNet18 is trained separately prior to episodic training of the adaptation networks. Within each residual block, Feature-wise Linear Modulation (FiLM) layers are inserted to compute a scale factor $\gamma$ and shift $\beta$ for each output channel, using block-specific adaptation networks $\psi_\theta$ that are conditioned on a task encoding. The task encoding $g_\theta(\mathcal{S})$ consists of the mean-pooled feature vectors of support examples produced by $d_\theta$, a separate but end-to-end learned Convolution Neural Network (CNN). This produces a task-adapted feature extractor $f_\theta$ (which implicitly depends on the support set $\mathcal{S}$) that maps support/query images onto the corresponding adapted feature space.  We will use $\mathcal{S}_\theta, \mathcal{Q}_\theta$ to denote versions of the support/query sets where each image $\mathbf x$ is mapped into its 512-dimensional feature vector representation $\mathbf z = f_\theta(\mathbf x)$.

Classification in CNAPS is performed by a task-adapted linear classifier where the class probabilities for a query image $\mathbf{x}_i^*$ are computed as $\text{softmax}(\mathbf{W} \mathbf{z}_i^* + \mathbf{b})$.
The classification weights $\mathbf{W}$ and biases $\mathbf{b}$ are produced by the classifier adaptation network $\psi_\theta^c$ forming $[\mathbf{W}, \mathbf{b}] = [\psi_\theta^c(\pmb{\mu}_1)\;\psi_\theta^c(\pmb{\mu}_2)\; \ldots\; \psi_\theta^c(\pmb{\mu}_K)]^T$
where for each class $k$ in the task, the corresponding row of classification weights is produced by $\psi_\theta^c$ from the class mean $\pmb{\mu}_k$. The class mean $\pmb{\mu}_k$ is obtained by mean-pooling the feature vectors of the support examples for class $k$ extracted by the adapted feature extractor $f_\theta$.

\subsection{Simple CNAPS}

\label{method:simple-cnaps}

\begin{figure}
    \centering
    \includegraphics[width=0.98\textwidth]{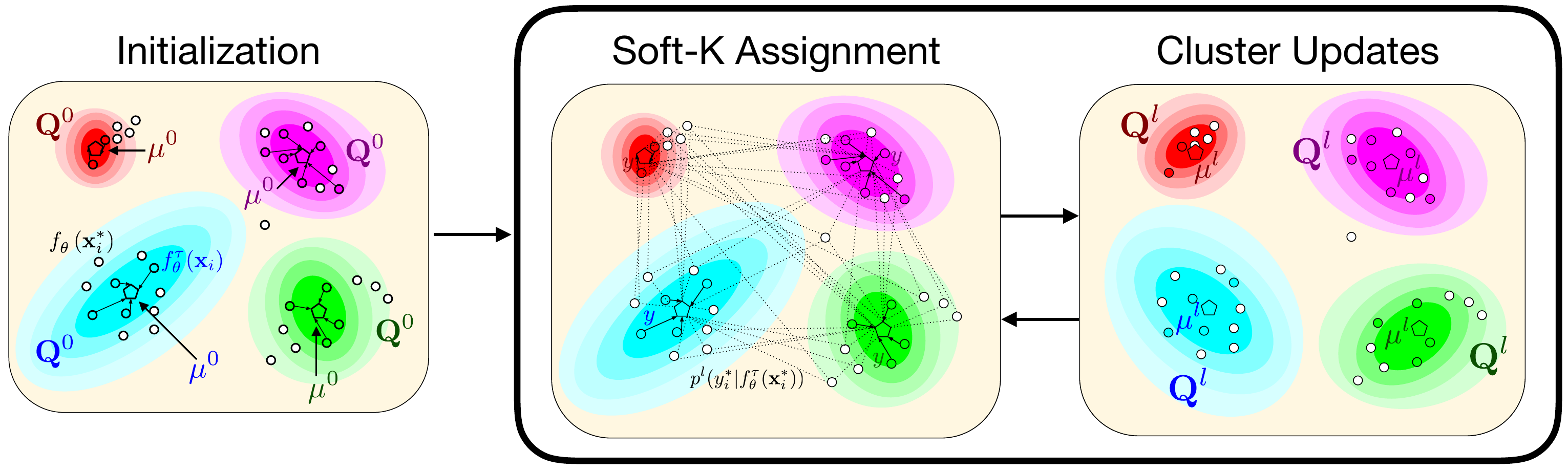}
    \caption{Transductive CNAPS' soft k-means Mahalanobis-distance based clustering procedure. First, cluster parameters are initialized using the support examples. Then, during cluster update iterations, the query examples are assigned class probabilities as soft labels and subsequently, both soft-labelled query examples and labelled support examples are used to estimate new cluster parameters.}
    \label{fig:clustering-method}
\end{figure}

In Simple CNAPS, we employ the same ResNet18 for feature extraction with the same adaptation module $\psi_\theta$. However, because of the novel classification architecture we use, it becomes trained to do something different than it does in CNAPS. This choice, like for CNAPS, allows for a task-dependent adaptation of the feature extractor. Unlike CNAPS, we undertake the classification step in Simple CNAPS by computing softmax of a Mahalanobis distance\footnote{Note that we explicitly do not use the $\frac{1}{2}$ coefficient as we experimentally verified doing so to decrease performance.},
\begin{equation}
    p(y^* = k \mid \mathbf z^*) \propto \exp
    \big(
      -(\mathbf z^* - \pmb{\mu}_k)^T\mathbf{Q}_k^{-1}(\mathbf z^* - \pmb{\mu}_k)
    \big)
    ,
    \label{simple-cnaps-eq:simple-cnaps-mahalanobis-classifier}
\end{equation}
of feature vector $\mathbf{z}^*_i$
relative to each class $k$ by estimating a mean $\pmb{\mu}_k$ and regularized covariance $\mathbf{Q}_k$ in the adapted feature space, using the support instances,
\begin{align}
    \pmb{\mu}_k
    &=
    \frac{1}
         {n_k}
    \sum_{i}
    \:
    \mathbb{I}[y_i=k] \:
    \mathbf z_i
    ,\textbf{}
    \\
    \mathbf{Q}_k
    &=
    \lambda_k \,
    \pmb{\Sigma}_k
    +
    (1 - \lambda_k) \,
    \pmb{\Sigma}
    +
    \beta I,
    &
    \lambda_k
    &=
    \frac{n_k}
         {n_k + 1}
    .
    \label{simple-cnaps-eq:mean-cov-calculation}
\end{align}

Here $\mathbb{I}[y_i=k]$ is the indicator function and $n_k = \sum_i \mathbb I[y_i=k]$ is the number of examples with class $k$ in the support set $\mathcal{S}$. The ratio $\lambda_k$ balances a task-conditional sample covariance $\pmb{\Sigma}$ and a class-conditional sample covariance $\pmb{\Sigma}_k$,
\begin{align}
    \pmb{\Sigma}
    &=
    \frac{1}{n}
    \sum_{i}
    \!
    \big( \mathbf z_i \!-\! \pmb{\mu}\big)
    \big( \mathbf z_i \!-\! \pmb{\mu} \big)^T,
    \\
    \pmb{\Sigma}_k
    &=
    \frac{1}{n_k}
    \sum_{i}
    \mathbb{I}[y_i=k]\:
    \big( \mathbf z_i \!-\! \pmb{\mu}_k \big)
    \big( \mathbf z_i \!-\! \pmb{\mu}_k \big)^T.
\end{align}
where $\pmb{\mu} = \frac{1}{n}\sum_{i} \mathbf z_i$ is the task-level mean. When few support examples are available for a particular class, $\lambda_k$ is small, and the estimate is regularized towards the task-level covariance $\pmb{\Sigma}$. As the number of support examples for the class increases, the estimate tends towards the class-conditional covariance $\pmb{\Sigma}_k$. Additionally, a regularizer $\beta I$ (we set $\beta=1$ in our experiments) is added to ensure invertibility. 

\subsection{Transductive CNAPS}

\label{sec:transductive-cnaps}

Transductive CNAPS extends Simple CNAPS by making use of query set, both for feature adaptation and classification. First, the task encoder $g_\theta$ is extended to incorporate both a support-set embedding $\mathbf{e}_s$ and a query-set embedding $\mathbf{e}_q$ such that,

\begin{align}
    \mathbf{e}_s
    &=
    \frac{1}{K} \sum_{k} \frac{1}{n_k} \sum_{i}
    \mathbb{I}[y_i=k]\: d_\theta(\mathbf{x}_i),
    \\
    \mathbf{e}_q
    &=
    \frac{1}{n_q} \sum_{i*} d_\theta(\mathbf{x}_i^*),
    \label{transductive-cnaps-eq:encoder}
\end{align}
where $d_\theta$ is a learned CNN. The support embedding $\mathbf{e}_s$ is formed by an average of the (encoded) support examples, with weighting inversely proportional to their class counts to prevent bias from the class imbalance. The query embedding $\mathbf{e}_q$ uses simple mean-pooling; both embeddings $\mathbf{e}_s$ and $\mathbf{e}_q$ are invariant to permutations of the corresponding support/query instances. We then process $\mathbf{e}_s$ and $\mathbf{e}_q$ through two steps of a Long Short Term Memory (LSTM) network in the same order to generate a final transductive task representation $g_\theta(\mathcal{S}, \mathcal{Q})$ to be used for adaptation. This process is visualized in Figure \ref{fig:transductive-task-encoder}. 

Second, we can interpret Simple CNAPS as a form of ``supervised clustering'' in feature space; each cluster (corresponding to a class $k$) is parameterized with a centroid $\bm{\mu}_k$ and a metric $\mathbf Q_k^{-1}$, and we interpret \eqref{simple-cnaps-eq:simple-cnaps-mahalanobis-classifier} as class assignment probabilities based on the distance to each centroid. With this viewpoint in mind, a natural extension to consider is to use the estimates of the class assignment probabilities on unlabelled instances to refine the class parameters $\bm{\mu}_k$ and $\mathbf Q_k$ in a soft $k$-means framework based on per-cluster Mahalanobis distances \cite{melnykov2014k}.  In this framework, as shown in Figure \ref{fig:clustering-method}, we alternate between computing updated assignment probabilities using \eqref{simple-cnaps-eq:simple-cnaps-mahalanobis-classifier} on the query set and using those assignment probabilities to compute updated class parameters.

We define $\mathcal{R}_\theta = \mathcal{S}_\theta \sqcup \mathcal{Q}_\theta$ as the disjoint union of the support set and the query set. For each element of $\mathcal{R}_\theta$, which we index by $j$, we define responsibilities $w_{jk}$ in terms of their class predictions when it is part of the query set and in terms of the label when it is part of the support set,
\begin{equation}
    w_{jk} = \begin{cases}
        p\big(y_j'=k \mid \mathbf z_j'\big)
        & \mathbf{z}_j'\in \mathcal{Q}_\theta,
        \\
        \mathbb{I}[y'_j = k]
        &
        (\mathbf{z}_j', y_j') \in \mathcal{S}_\theta.
       \end{cases}
    \label{eq:class_weights}
\end{equation}
Using these responsibilities we can incorporate unlabelled samples from the support set by defining weighted estimates $\bm{\mu}'_k$ and $\mathbf{Q}'_k$:
\begin{align}
    \label{eq:update_first}
    \bm{\mu}'_k
    &=
    \frac{1}{n'_k}
    \sum_{j}
    w_{jk} \:
    \mathbf z_j',
    \\
    \mathbf{Q}'_k
    &=
    \lambda'_k
    \bm{\Sigma}'_k
    +
    (1 - \lambda'_k)
    \bm{\Sigma}'
    + \beta I
    ,
\end{align}
where $n'_k = \sum_{j} w_{jk}$ defines $\lambda'_k = n'_k / (n'_k + 1)$. The covariance estimates $\bm{\Sigma}'$ and $\bm{\Sigma}_k'$ are
\begin{align}
    \label{eq:update_last}
    \bm{\Sigma}'
    &=
    \frac{1}{\sum_k n'_k}
    \sum_{jk}
    \!
    w_{jk}
    \big( \mathbf z_j' \!-\! \bm{\mu}' \big)
    \big( \mathbf z_j' \!-\! \bm{\mu}' \big)^T,
    \\
    \bm{\Sigma}'_k
    &=
    \frac{1}{n'_k}
    \sum_{j}
    \!
    w_{jk}
    \big( \mathbf z_j' \!-\! \bm{\mu}'_k \big)
    \big( \mathbf z_j' \!-\! \bm{\mu}'_k \big)^T.
\end{align}
where $\bm{\mu}' = \left(\sum_{k} n_k'\right)^{-1}\sum_{jk}w_{jk}\mathbf z_j'$ is the task-level mean.

\begin{algorithm*}[t]
    \caption{Iterative Refinement in Transductive-CNAPS}
    \label{algo:soft-kmeans-cnaps}
    \begin{algorithmic}[1] 
        \Procedure{compute\_query\_labels}{$\mathcal{S}_\theta,\mathcal{Q}_\theta, N_\text{iter}$}
            \State For $j$ ranging over support and query sets, $w_{jk} \gets \begin{cases}
                1 &\text{if}~ (\mathbf{z}'_j,y'_j) \in \mathcal{S}_\theta \text{~and~} y_j=k
                \\ 0 &\text{otherwise}
            \end{cases}$
            \For{iter = $0 \cdots N_\text{iter}$}\Comment The first iteration is equivalent to Simple CNAPS;
                \State Compute class parameters $\bm{\mu}_k, \mathbf Q_k$ according to update equations \eqref{eq:update_first}-\eqref{eq:update_last}
                \State Compute class weights using class parameters according to \eqref{eq:class_weights}
                \State \textbf{break} if the most probable class for each query example hasn't changed
            \EndFor
            \State \Return class probabilities $w_{jk}$ for $j$ corresponding to $\mathcal{Q}_\theta$
        \EndProcedure
    \end{algorithmic}
    \label{alg:iterative}
\end{algorithm*}

These update equations are weighted versions of the original Simple CNAPS estimators from Section \ref{method:simple-cnaps}, and reduce to them exactly in the case of an empty query set.

Algorithm \ref{alg:iterative} summarizes the soft k-means procedure based on these updates. We initialize class weights using only the labelled support data. We use those weights to compute class parameters, then compute updated weights using both the support and query sets. At this point, the weights associated with the query set $\mathcal{Q}$ are the same class probabilities as estimated by Simple CNAPS.  However, in Transductive CNAPS, we repeat this procedure iteratively until we reach either reach a maximum number of iterations, or until class assignments $\text{argmax}_k \, w_{jk}$ stop changing.

Unlike the transductive task-encoder, this second extension in Transductive CNAPS, namely the soft k-mean iterative estimation of class means and covariance estimates, is used at test time only. During training, a single estimation is produced for both mean and covariance using only the support examples. This, as we discuss more in Section \ref{exp:feot-vs-cot}, was shown to empirically perform better. See Figure \ref{fig:transductive-cnaps-vs-simple-cnaps} for a high-level comparison of classification in Simple CNAPS vs. Transductive CNAPS.

\subsection{Active Learning}

Following past work \cite{Requeima19_CNAPS}, our main objective is to evaluate if Simple and Transductive CNAPS perform better with uncertainty-based active label aquisition methods as oppose to random selection when used in ``out of the box'' active learning.

Here, both models are first trained on a large few-shot learning benchmark, namely Meta-Dataset \cite{triantafillou2019meta}. Then, without any further training, they are evaluated for active learning, where given a set of unlabelled images, a label can be acquired at each iteration. We consider three standard label acquisitions methods: random selection, and two uncertainty-based approaches, specifically predictive entropy \cite{Cohn96_PredictiveEntropy} and variation ratios \cite{Cohn96_PredictiveEntropy}.

For uncertainty-based label acquisition, unlabelled examples are first processed by the adapted network where class probabilities are produced. These class probabilities are then used to rank examples. In predictive entropy, the entropy of class probabilities are calculated and examples with maximal predictive entropy are selected for label acquisition. In variation ratios, $\text{MaxProb}(\mathbf{y}_i)$ is calculated as defined by the maximum probability assigned to any of the classes. Examples are then ranked based on this score, where the lowest scoring examples are selected for label acquisition.

\subsection{Continual Learning}

To perform ``out of the box'' continual learning, we modify Simple and Transductive CNAPS in two major ways. After both methods have been episodically meta-trained on a large few-shot learning benchmark (e.g. Meta-Dataset \cite{triantafillou2019meta}), each architecture is deployed in a continuous learning problem setting where at each time $t$, a new task consisting of $(\mathcal{S}^t, \mathcal{Q}^t)$ is presented for undertaking.

\begin{figure*}
    \centering
    \includegraphics[width=0.95\textwidth]{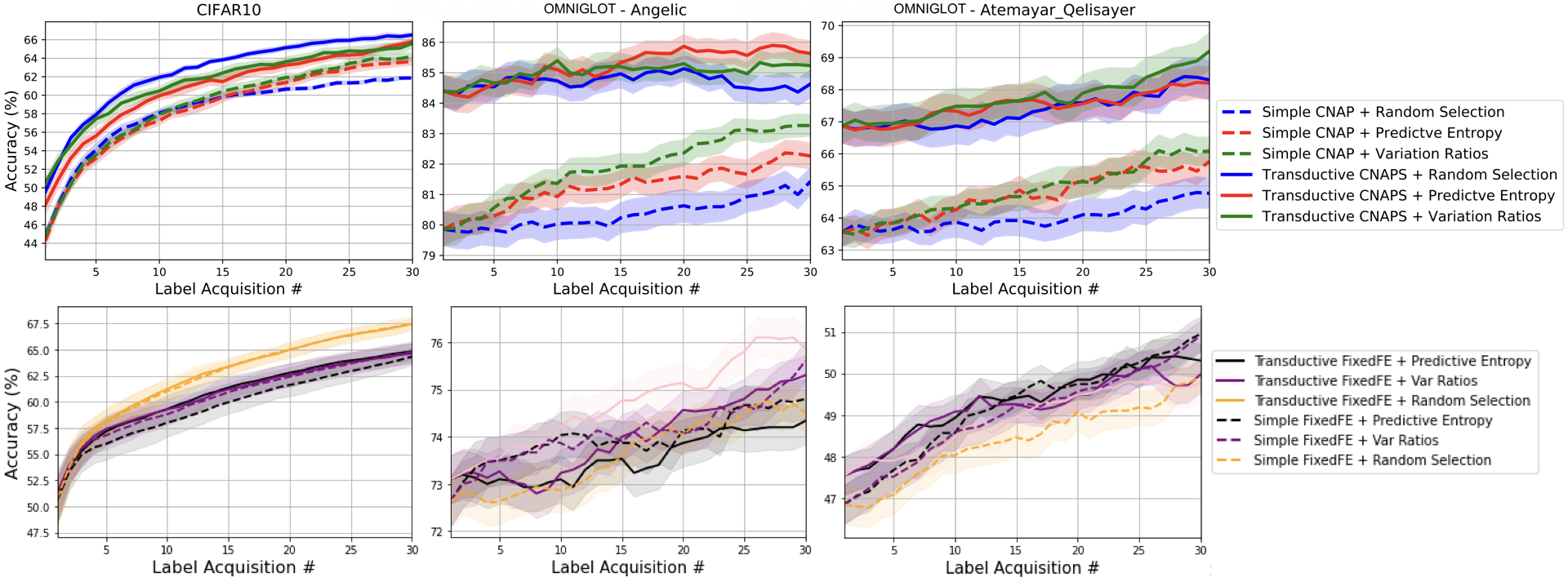}
    \caption{``Out of the box'' active learning performance of Simple CNAPS and Transductive CNAPS on CIFAR 10 and select OMNIGLOT languages. For each method, we use 3 label acquisition methods. Results on other OMNIGLOT languages can be found in the Appendix.}
    \label{fig:active-learning-figure}
\end{figure*}

Each task at time $t$ can consist of both previously seen classes and entirely novel categories. For novel classes, we produce and save class-wise estimates of mean and covariance using equations 1 and \ref{simple-cnaps-eq:mean-cov-calculation} for Simple CNAPS and their corresponding transductive extensions for Transductive CNAPS. For previously seen classes, weighted estimates are produced for both means and covariances based on the number of class examples present in the task and the class instances previously seen, 
\begin{align}
    \bm{\mu}_k &= \frac{n_k^t}{n_k^t + n_k^{1:t-1}}\times \bm{\mu}_k^t + \frac{n_k^{1:t-1}}{n_k^t + n_k^{1:t-1}}\times \bm{\mu}_k', \\
    \mathbf{Q}_k &= \frac{n_k^t}{n_k^t + n_k^{1:t-1}}\times \mathbf{Q}_k^t + \frac{n_k^{1:t-1}}{n_k^t + n_k^{1:t-1}}\times \mathbf{Q}_k'.
\end{align}
where $n_k^t$ indicates the number of support examples for class $k$ within the new task seen at time $t$ and $n_k^{1:t-1}$ is the number of support instances seen for class $k$ prior to the task. $\bm{\mu}_k'$ and $\mathbf{Q}_k'$ specify saved mean and covariance estimates prior to the current task whereas $\bm{\mu}_k^t$ and $\mathbf{Q}_k^t$ are task-parameters estimated using the task at time $t$.

The second modification to both methods for continual learning involves the continuous adaptation of the feature extractor $f_\theta$. This is primarily accomplished through developing strategies for continual updating of the task-encoding. Let us denote this task-encoding at time $t$ as $\mathbf{d}^t$. We consider three update strategies. 

First, we use the ``Moving Encoding'', where $\mathbf{d}^t = g_\theta(\mathcal{S}^t)$ for Simple CNAPS and $\mathbf{d}^t = g_\theta(\mathcal{S}^t, \mathcal{Q}^t)$ for Transductive CNAPS. That is, the encoding is set to the task-encoding of the most recent task. This naturally results in different feature space manifolds at different time steps $t$ while saved class representations are based on manifolds at the time of their respective tasks.

Second, we focus on the case of fixing the feature space to the task-encoding of the first task, $t=1$. We refer to this variation at ``First Encoding'', where $\mathbf{d}^t = g_\theta(\mathcal{S}^1)$ for Simple CNAPS and $\mathbf{d}^t = g_\theta(\mathcal{S}^1, \mathcal{Q}^1)$ for Transductive CNAPS. This assures that all class-wise and query embeddings are generated within the same feature manifold. However, that manifold has been task-adapted for maximal performance on only the first task observed during continual learning.

Third, we consider the ``Averaging Encoding'' strategy, with

\begin{align}
    \mathbf{d}^t &= \frac{1}{t} g_\theta(\mathcal{S}^t) + \frac{t-1}{t} d^{t-1} \text{ (Simple CNAPS)}, \\
    \mathbf{d}^t &= \frac{1}{t} g_\theta(\mathcal{S}^t, \mathcal{Q}^t) + \frac{t-1}{t} d^{t-1} \text{ (Transductive CNAPS)}.
\end{align}

where a $t$-scaled weighted convex combination of the current task-encoding $g_\theta(\mathcal{S}^t)$/$g_\theta(\mathcal{S}^t, \mathcal{Q}^t)$ and the task-encoding up to the previous time step $d^{t-1}$. This update strategy strikes a hybrid balance between maintaining feature space manifolds that don't vary considerably between time-steps, while partially adapting the feature space to new tasks as they come.

\section{Theoretical Motivation}

\begin{figure}[t]
    \centering
    \subfloat[Multi-Head + Moving Encoding]{{\includegraphics[width=0.5\textwidth]{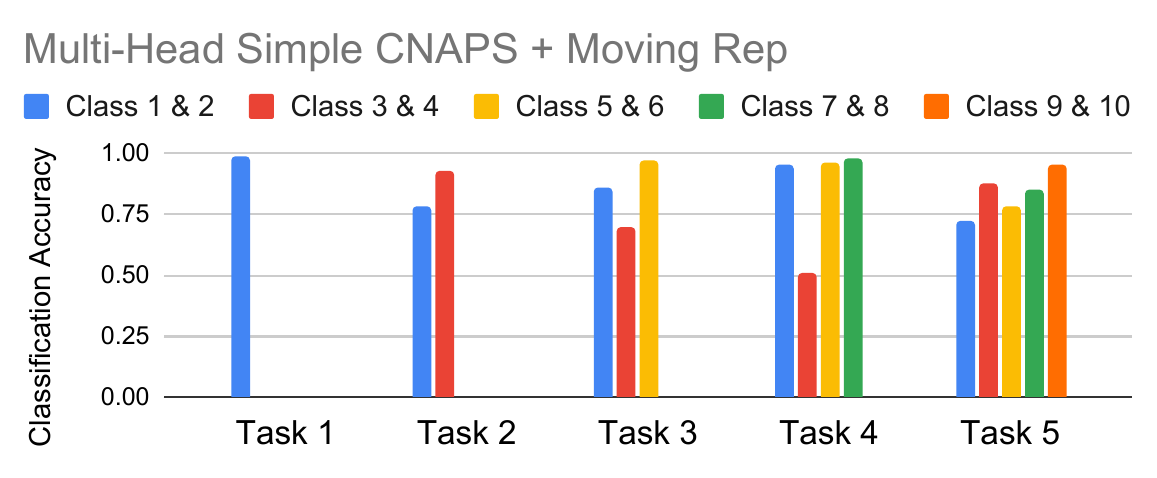} }}
    \subfloat[Single-Head + Moving Encoding]{{\includegraphics[width=0.5\textwidth]{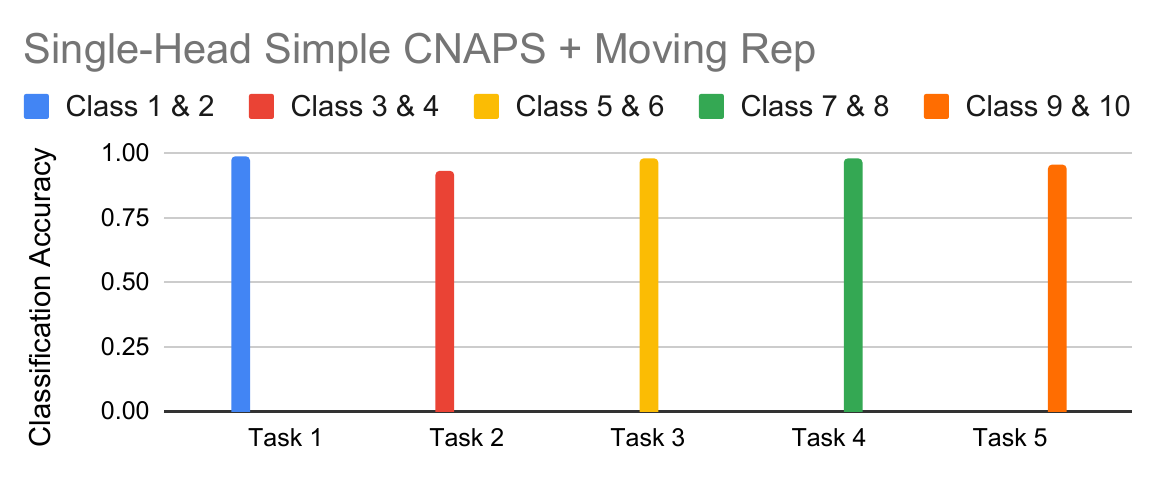} }} \\
    \subfloat[Multi-Head + First Encoding]{{\includegraphics[width=0.5\textwidth]{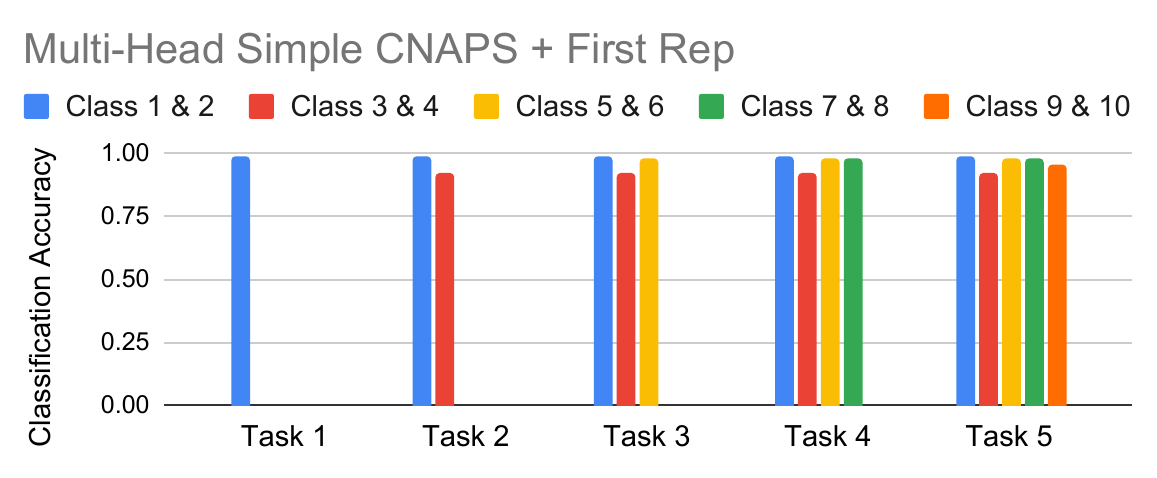} }}
    \subfloat[Single-Head + First Encoding]{{\includegraphics[width=0.5\textwidth]{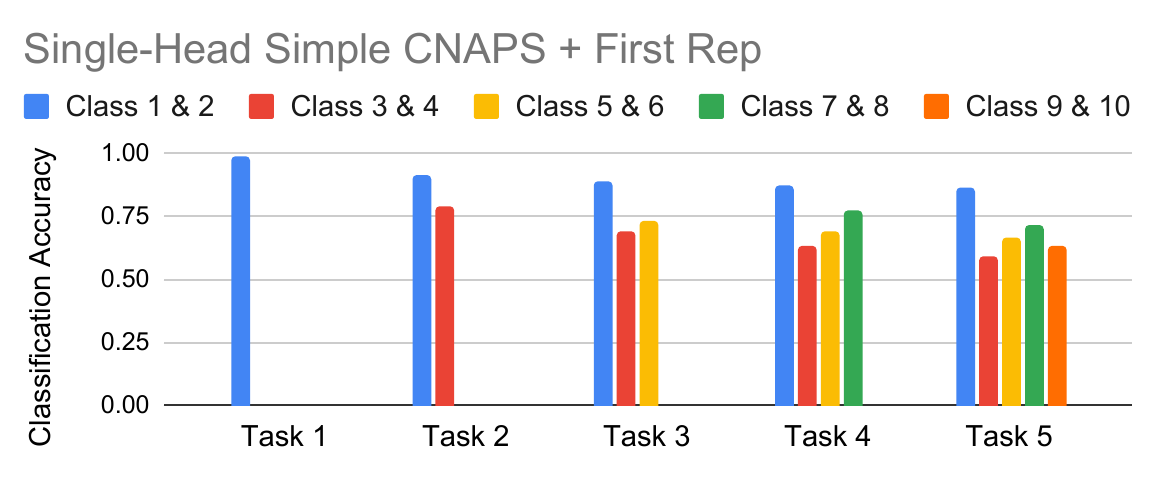} }} \\
    \subfloat[Multi-Head + Averaging Encoding]{{\includegraphics[width=0.5\textwidth]{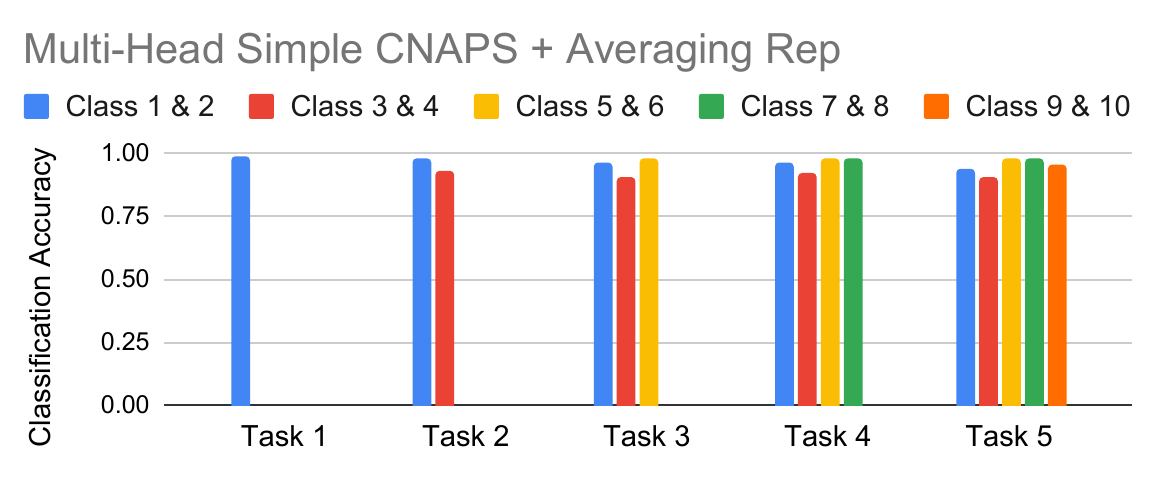} }}
    \subfloat[Single-Head + Averaging Encoding]{{\includegraphics[width=0.5\textwidth]{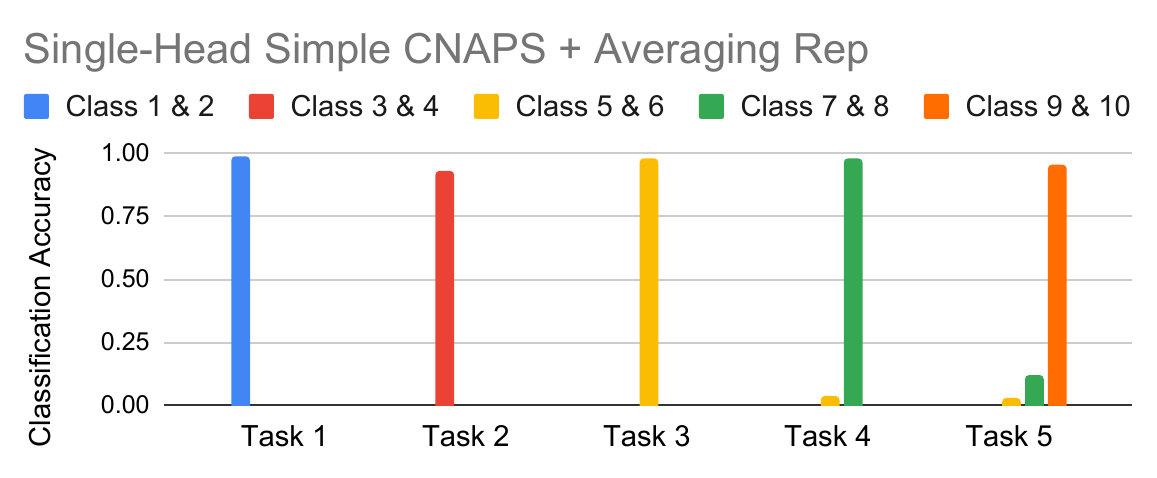} }}\\
    \caption{Task-wise classification accuracy of ``out of the box'' continual learning on MNIST. Here, consecutive pairs of digits are grouped together one task and iteratively shown to the models. X-axis specifies the iteration in the continual learning process while the coloured bars signify classification accuracy on the query test set of each task.}
    \label{fig:continual-learning-charts}
\end{figure}

\subsection{Relationship to Bregman Soft Clustering}

The procedure in Algorithm \ref{algo:soft-kmeans-cnaps} resembles the Bregman clustering algorithms of Banerjee et al. \cite{banerjee2005clustering}. Specifically, the updates to soft assignments $w_{jk}$ in Equation \ref{simple-cnaps-eq:simple-cnaps-mahalanobis-classifier} are the semi-supervised equivalent of those in Bregman soft clustering, in which the divergence is based on the Mahalanobis distance  $F(\textbf{z}) = \textbf{z}^T \textbf{Q}^{-1} \textbf{z}$,
\begin{align}
    D_F(\textbf{z}, \textbf{z}') = F(\textbf{z}) - F(\textbf{z}') - \nabla F(\textbf{z}')^T(\textbf{z} - \textbf{z}').
\end{align}
However, Algorithm~\ref{algo:soft-kmeans-cnaps} differs in that it updates both $\bm{\mu}'_k$ and $\textbf{Q}'_k$ at each iteration, rather than just $\bm{\mu}'_k$.

In general, any (regular) exponential family can be associated with a Bregman divergence and vice versa, which gives rise to a correspondence between EM-based clustering and Bregman soft clustering algorithms \cite{banerjee2005clustering}. Standard Bregman soft clustering corresponds to EM in which the likelihood is a Gaussian with unknown mean and a known covariance $\textbf{Q}$ that is shared across clusters. The case where the covariance is unknown corresponds to Gaussian mixture models (GMMs), but the function $F(\textbf{z})$ is not simply the Mahalanobis distance in this case.

The updates for $\bm{\mu}'_k$ and $\textbf{Q}'_k$ in Algorithm~\ref{algo:soft-kmeans-cnaps} are equivalent to those in a GMM that incorporates regularization for the covariances. However, GMM clustering differs in the calculation of the assignment probabilities
\begin{align}
    p(y^* &= k \mid \mathbf z^*) \propto \pi_k\\ &\exp
    \left(
      -\frac{1}{2}(\mathbf z - \bm{\mu_k})^T\mathbf{Q}_k^{-1}(\mathbf z - \bm{\mu}_k)-\frac{1}{2}\log |\mathbf Q_k|\right).
\end{align}
These probabilities incorporate a term $\pi_k = p(y^* = k)$, which defines a prior probability of assignments to a cluster, and a term $\exp(- \log |\textbf{Q}_k|)$, which reflects the fact that GMMs employ a likelihood with unknown covariance.

In short, our clustering procedure employs an update to soft assignments $w_{jk}$ that is similar to that of soft Bregman clustering, but employs an updates to $\bm{\mu}'_{k}$ and $\textbf{Q}'_k$ that are similar to those in a (regularized) Gaussian Mixture Model (GMM). In Section \ref{exp:gaussian-mixture-models} we demonstrate through ablations that this combination of updates improves empirical performance relative to baselines that perform GMM-based clustering.

\subsection{Connections with Riemannian Metric Learning}

By dropping the log-determinant of the class covariances $\mathbf Q_k$, we lose the ability to interpret the model as a Gaussian mixture model, since each mixture component is no longer a normalized conditional density $p(\bm{z}^* \mid \bm{\mu}_k, \bm{Q}_k, \bm{y}^*\!=\!k)$.
We show that by postulating that the feature space is best described as a Riemannian manifold, our relative class scores \eqref{simple-cnaps-eq:simple-cnaps-mahalanobis-classifier} approximate the squared \emph{geodesic} distance between a test point $\mathbf z^*$ and the centroid $\bm{\mu}_k$.

\begin{figure}%
    \centering
    \subfloat[Transductive]{{\includegraphics[width=0.5\textwidth]{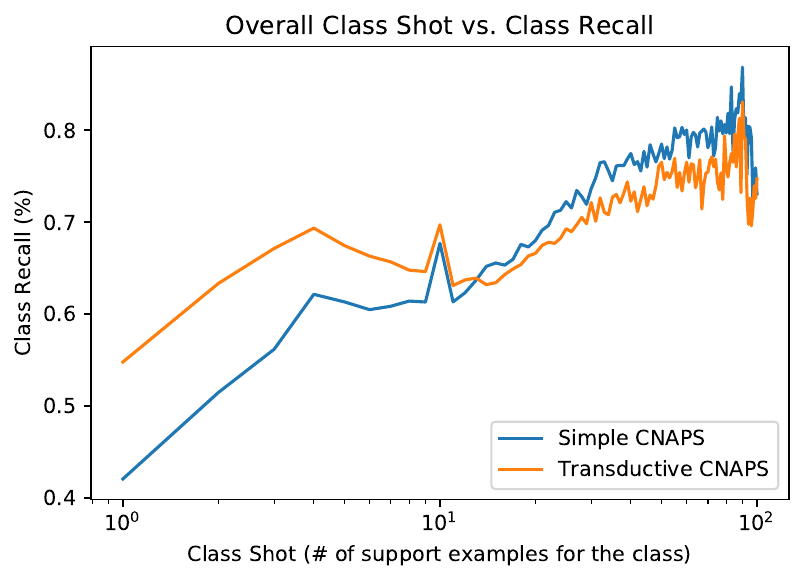} }}
    \subfloat[Non-Transductive]{{\includegraphics[width=0.44\textwidth]{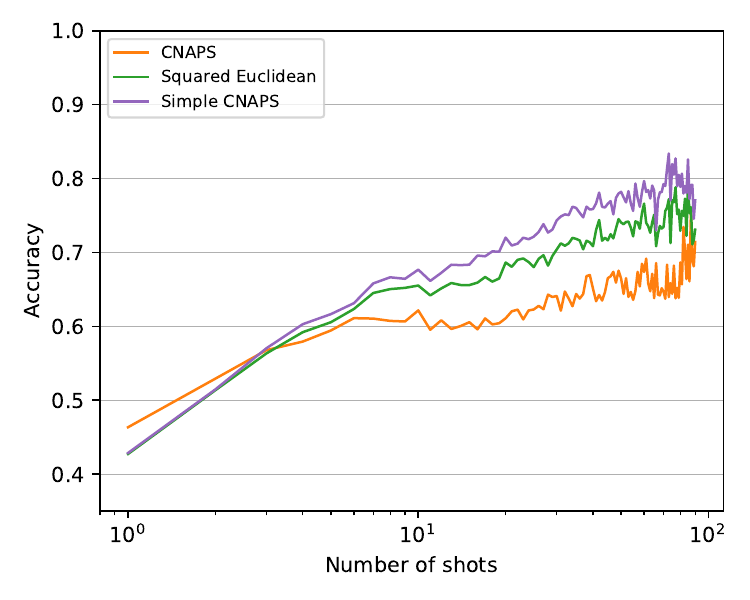} }}
    \vspace{-0.1in}
    \caption{Class recall averaged between classes across Meta-Dataset.}
    \vspace{-0.2in}
    \label{fig:ratios-and-shots-all-ranges-10max}
\end{figure}

The geometry of a Riemannian manifold is defined by a local \emph{metric tensor} ($d\times d$ positive-definite matrix defined for each point in $\mathbb{R}^d$), $\vf g(\vf x)$, in a data or feature space $\mathbb{R}^d$ (more general topologies are not considered in this discussion).
The metric tensor defines the geometry of the underlying space; particularly, we can employ it to define a notion of length. The distance along a path $\gamma : [0, 1] \rightarrow \mathbb{R}^d$ is computed in terms of this metric tensor via the arclength functional:
\begin{equation}
    \mathcal{L}[\gamma] = \int_0^1  \sqrt{\dot\gamma(\lambda)^T \vf g(\gamma(\lambda)) \dot\gamma(\lambda)} d\lambda
\end{equation}
From this, we can derive a global distance (the geodesic distance) between points (at least in the $\mathbb{R}^d$ case) as the length of the shortest path between $\vf x$ and $\vf y$:
\begin{equation}
    d(\vf x, \vf y) = \inf_\gamma \mathcal{L}[\gamma]~\text{for $\gamma$ where}~\gamma(0) = \vf x, \gamma(1) = \vf y
\end{equation}
The arclength functional is difficult to analyze, but we can instead analyze the related \emph{energy functional} \cite{carmo1992riemannian},
\begin{equation}
    E[\gamma] = \int_0^1 \dot\gamma(\lambda)^T \vf g(\gamma) \dot \gamma(\lambda) d\lambda
\end{equation}
Both $E$ and $\mathcal{L}$ yield the same local minimizers; these are called geodesics, and are equivalent to straight lines within the geometry defined by $\vf g(\vf x)$.

In metric learning, our goal is to estimate the metric tensor $\vf g(\vf x)$ from data. This is an underdetermined task, since its only constraints are smoothness and positive definiteness. To reduce the space of metric tensors under consideration, we treat the class centroids $\bm \mu_k$ as local inducing points for a metric that is locally (near $\mu_k$) the Mahalanobis distance defined by $\vf Q_k$, and model the global metric tensor as some smooth interpolation of these local Mahalanobis metrics:
\begin{equation}
    \vf g(\vf x) = \sum_k w_k(\vf x - \mu_k)\vf Q_k^{-1}
\end{equation}

Here, $\{w_k(\vf r)\}_k$ is a smooth partition of unity, which satisfies
\begin{align}
    w_k(\vf 0) &= 1 &
    w_k(\vf x) &\ge 0 &
  \sum_k w_k(\vf x) &= 1
\end{align}
The existence of such functions is guaranteed \cite{lee2013smooth}.

Even with these simplifying assumptions, the global geodesic distance is extremely challenging to compute. Since the geodesic distance is the minimum path length over all paths, we can upper bound it by a specific path \cite{ramanan2010local}:
\begin{equation}
    d(\vf x, \vf y) \le  \int_0^1 \sqrt{ \dot{\vf c}(\lambda)^T\vf g(\vf c(\lambda))\dot{\vf c}(\lambda)} d\lambda
\end{equation}
where $\vf c(\lambda) = (1 - \lambda) \vf x + \lambda \vf y$ is a straight line (in the coordinate space) interpolation. The time derivative of $\vf c$ is easily computed to be $\dot{\vf c} = \vf y - \vf x$. The corresponding energy functional upper bounds the squared distance,
\begin{equation}
    d(\vf x, \vf y)^2 \le \tilde{E}(\vf x, \vf y) = \int_0^1 \dot{\vf c}(\lambda)^T\vf g(\vf c(\lambda))\dot{\vf c}(\lambda) d\lambda.
\end{equation}

For purposes of classification, computing the exact distance to each centroid is not necessary; we only need to reason about the \emph{difference} in distance between a test point $\vf x$ and two class centroids $\bm \mu_i, \bm \mu_j$, which corresponds to their relative (log) probabilities. Substituting the energy along the straight-line paths yields
\begin{align}
    \Delta \tilde E_{ij}
    &= \tilde E(\vf x, \mu_i) - \tilde E(\vf x, \mu_j), \\
    &= \int_0^1\Delta_i^T \vf g_i(\lambda) \Delta_i - \Delta_j^T \vf g_j(\lambda) \Delta_j  d\lambda,
\end{align}
where $\Delta_k$ and and $g_k(\lambda)$ denote,
\begin{align}
    \Delta_k &= (\vf x - \mu_k), \\
    \vf g_k(\lambda) &= \vf g(\lambda \mu_k + (1 - \lambda)\vf x).
\end{align}
We can write $\Delta E_{ij}$ in terms of a path parameter $\tau \in [0, \frac{1}{2}]$ as
\begin{align*}
    \Delta \tilde E_{ij}(\tau) = &\int_0^{\tau} \Delta_i^T \vf g_i(\lambda) \Delta_i - \Delta_j^T \vf g_j(\lambda) \Delta_j d\lambda \\ &+ \int_{1-\tau}^1 \Delta_i^T \vf g_i(\lambda) \Delta_i - \Delta_j^T \vf g_j(\lambda) \Delta_j d\lambda.
\end{align*}
As $T$ increases, we simultaneously grow the path inward from each class centroid towards $\vf x$ and outward from $\vf x$ to the centroids.

A first-order Taylor expansion of this around $T=0$ causes the first integral to drop out (since $g_i(0) = g_j(0)$), and the second integral yields (when evaluated at $T=\frac{1}{2}$):
\begin{align*}
    \Delta \tilde E_{ij} \approx \frac{1}{2}[&(\vf x - \mu_i)^T\vf Q_i^{-1}(\vf x - \mu_i) \\ &- (\vf x - \mu_j)^T\vf Q_j^{-1}(\vf x - \mu_j)].
\end{align*}
The higher order terms can be controlled to some extent by forcing the partition functions $w_k(\vf x)$ to be flat near the class centroids.

This allows us to think of the Simple CNAPS classifier logit function (the squared Mahalanobis distance to the class centroids with per-class metrics) as a coarse approximation of the (squared) geodesic distance between a query test point and a class centroid; this coarse approximation is improved slightly by noting that other low-order terms that could be considered drop out when examining the difference in squared geodesic distance between a test point and two class centroids. 

There are a few implications of this viewpoint, which point to directions for future work. First, the performance gains that we see from using per-class Mahalanobis metrics suggests that modeling the local geometry of the adapted features is important. By taking the manifold viewpoint more seriously, we could consider more principled geometric algorithms such as manifold-regularized SVMs \cite{belkin2006manifold} to further exploit the local geometry of the adapted feature space.  Second, while the inverse weighted sample covariance works in practice as a local metric, and makes intuitive sense, we could consider techniques specifically designed for estimating local Riemannian metrics \cite{hauberg2012geometric} to provide better (or at least more principled) estimates of our local metrics $\mathbf Q_k^{-1}$.

\section{Experiments}

\subsection{Few-Shot Learning Benchmarks}
\label{experiments:benchmarks}

\noindent
\textbf{Meta-Dataset} \cite{triantafillou2019meta} is a few-shot visual classification benchmark consisting of 10 widely used datasets: ILSVRC-2012 (ImageNet) \cite{russakovsky2015imagenet}, Omniglot \cite{lake2015human}, FGVC-Aircraft (Aircraft) \cite{maji2013fine}, CUB-200-2011 (Birds) \cite{wah2011caltech}, Describable Textures (DTD) \cite{cimpoi2014describing}, QuickDraw \cite{jongejan2016quick}, FGVCx Fungi (Fungi)  \cite{fungi2018schroeder}, VGG Flower (Flower) \cite{nilsback2008automated}, Traffic Signs (Signs) \cite{houben2013detection} and MSCOCO \cite{lin2014microsoft}. Consistent with past work \cite{Requeima19_CNAPS, bateni2019improved}, we train our model on the official training splits of the first 8 datasets and use the test splits to evaluate in-domain performance. We use the remaining two datasets as well as three external benchmarks, namely MNIST \cite{lecun-mnisthandwrittendigit-2010}, CIFAR10 \cite{Krizhevsky09learningmultiple} and CIFAR100 \cite{Krizhevsky09learningmultiple}, for out-of-domain evaluation.

Task generation in Meta-Dataset follows a complex procedure. The \textit{task} is defined as an instance of classification problem that 
can be of different \textit{ways} (number of classes) and individual classes can be of varying \textit{shots} (number of samples per class) even within the same task. 
For example, an instance of MNIST classification task, may require classification of digits `1', `2' and `4', each of which contains 7, 2 and 9 training support instances respectively. This would constitute a 3-way task with 7-, 2- and 9-shots.
In task formation, the task \textit{way} is first sampled uniformly between 5 and 50 and \textit{way} classes are selected at random from the corresponding class/dataset split. Then, for each class, 10 instances are sampled at random and used as query examples for the class, while of the remaining images for the class, a \textit{shot} is sampled uniformly from [1, 100] and \textit{shot} number of images are selected at random as support examples with total support set size of 500. 

Additional dataset-specific constraints are enforced, as discussed in Section 3.2 of \cite{triantafillou2019meta}, and since some datasets have fewer than 50 classes and fewer than 100 images per class, the overall \textit{way} and \textit{shot} distributions resemble Poisson distributions where most tasks have fewer than 10 classes and most classes have fewer than 10 support examples (see Appendix-\ref{appendix:benchmark-and-training:meta-dataset}). Following \cite{bateni2019improved} and \cite{Requeima19_CNAPS}, we first train our ResNet18 feature extractor on the Meta-Dataset defined training split of ImageNet following the procedure in Appendix-\ref{appendix:benchmark-and-training:meta-dataset-training-and-testing}. The ResNet18 parameters are then kept fixed while we train the adaptation network for a total of sampled 110K tasks using Episodic Training \cite{Snell17_Proto, finn2017model} (see Appendix-\ref{appendix:benchmark-and-training:meta-dataset-training-and-testing}).

\vspace{0.05in}
\noindent
\textbf{Mini/tiered-ImageNet} \cite{vinyals2016matching, DBLP:journals/corr/abs-1803-00676-tieredimagenet} are two benchmarks for few-shot learning. Both datasets employ subsets of ImageNet \cite{russakovsky2015imagenet} with a total of 100 classes and 60K images in mini-ImageNet and 608 classes and 779K images in tiered-ImageNet. Unlike Meta-Dataset, tasks across these datasets have pre-defined \textit{shots} and \textit{ways} that are uniform across every task in the specified setting.

Following \cite{DBLP:journals/corr/abs-1803-02999-reptile, DBLP:journals/corr/abs-1805-10002-tpn, Snell17_Proto}, we report performance on the 1/5-\textit{shot} 5/10-\textit{way} settings across both datasets with 10 query examples per class. We first train the ResNet18 on the training set of the corresponding benchmark at hand following the procedure noted in Appendix-\ref{appendix:benchmark-and-training:minitiered-training-and-testing}. We also consider a more feature-rich ResNet18 trained on the larger ImageNet dataset. However, we exclude classes and examples from test sets of mini/tiered-ImageNet to address potential class/example overlap issues, resulting in 825 classes and 1,055,494 images remaining. Then, with the ResNet18 parameters fixed, we train episodically for 20K tasks (see Appendix-\ref{appendix:benchmark-and-training:minitiered} for details).

\subsection{Results}

\begin{table*}
    \centering
    \scriptsize
    \tabcolsep=0.1cm
    \begin{tabular}{lccccccccccc}
        & \multicolumn{8}{c}{In-Domain Accuracy (\%)} \\
        \cmidrule(lr){2-9}
        Model & \rotatebox{45}{ImageNet} & \rotatebox{45}{Omniglot} & \rotatebox{45}{Aircraft} & \rotatebox{45}{Birds} & \rotatebox{45}{DTD} & \rotatebox{45}{QuickDraw} & \rotatebox{45}{Fungi} & \rotatebox{45}{Flower} \\
        \midrule
        RelationNet \cite{Sung2018_RelationNet} & 30.9\textpm0.9 & 86.6\textpm0.8 & 69.7\textpm0.8 & 54.1\textpm1.0 & 56.6\textpm0.7 & 61.8\textpm1.0 & 32.6\textpm1.1 & 76.1\textpm0.8 \\
        MatchingNet \cite{Vinyals16_MatchingNet} & 36.1\textpm1.0 & 78.3\textpm1.0 & 69.2\textpm1.0 & 56.4\textpm1.0 & 61.8\textpm0.7 & 60.8\textpm1.0 & 33.7\textpm1.0 & 81.9\textpm0.7 \\
        MAML \cite{Finn17_MAML} & 37.8\textpm1.0 & 83.9\textpm1.0 & 76.4\textpm0.7 & 62.4\textpm1.1 & 64.1\textpm0.8 & 59.7\textpm1.1 & 33.5\textpm1.1 & 79.9\textpm0.8 \\
        ProtoNet \cite{Snell17_Proto} & 44.5\textpm1.1 & 79.6\textpm1.1 & 71.1\textpm0.9 & 67.0\textpm1.0 & 65.2\textpm0.8 & 64.9\textpm0.9 & 40.3\textpm1.1 & 86.9\textpm0.7 \\
        ProtoMAML \cite{triantafillou2019meta} & 46.5\textpm1.1 & 82.7\textpm1.0 & 75.2\textpm0.8 & 69.9\textpm1.0 & 68.3\textpm0.8 & 66.8\textpm0.9 & 42.0\textpm1.2 & 88.7\textpm0.7 \\
        CNAPS \cite{Requeima19_CNAPS} & 52.3\textpm1.0 & 88.4\textpm0.7 & 80.5\textpm0.6 & 72.2\textpm0.9 & 58.3\textpm0.7 & 72.5\textpm0.8 & 47.4\textpm1.0 & 86.0\textpm0.5 \\
        BOHB-E \cite{bohb-e-saikia2020optimized} & 55.4\textpm1.1 & 77.5\textpm1.1 & 60.9\textpm0.9 & 73.6\textpm0.8 & \textbf{72.8\textpm0.7} & 61.2\textpm0.9 & 44.5\textpm1.1 & 90.6\textpm0.6 \\
        TaskNorm \cite{Bronskill20_TaskNorm} & 50.6\textpm1.1 & 90.7\textpm0.6 & 83.8\textpm0.6 & 74.6\textpm0.8 & 62.1\textpm0.7 & 74.8\textpm0.7 & 48.7\textpm1.0 & 89.6\textpm0.5 \\
        SUR \cite{dvornik20_SUR} & 56.3\textpm1.1 & 93.1\textpm0.5 & \textbf{85.4\textpm0.7} & 71.4\textpm1.0 & \textbf{71.5\textpm0.8} & 81.3\textpm0.6 & \textbf{63.1\textpm1.0} & 82.8\textpm0.7 \\
        URT \cite{liu20_URT} & 55.7\textpm1.0 & \textbf{94.4\textpm0.4} & \textbf{85.8\textpm0.6} & \textbf{76.3\textpm0.8} & \textbf{71.8\textpm0.7} & \textbf{82.5\textpm0.6} & \textbf{63.5\textpm1.0} & 88.2\textpm0.6 \\
        \midrule
        Simple CNAPS & \textbf{58.6\textpm1.1} & 91.7\textpm0.6 & 82.4\textpm0.7 & 74.9\textpm0.8 & 67.8\textpm0.8 & 77.7\textpm0.7 & 46.9\textpm1.0 & 90.7\textpm0.5 \\
        Transductive CNAPS & \textbf{58.8\textpm1.1} & \textbf{93.9\textpm0.4} & 84.1\textpm0.6 & \textbf{76.8\textpm0.8} & 69.0\textpm0.8 & 78.6\textpm0.7 & 48.8\textpm1.1 & \textbf{91.6\textpm0.4} \\
    \end{tabular}
    \caption{Few-shot classification on in-domain Meta-Dataset tasks. Error intervals correspond to 95\% confidence intervals, and bold values indicate statistically significant SoTA performance.}
    \label{tab:results:meta-dataset-1}
\end{table*}{}

\begin{table*}
    \centering
    \scriptsize
    \tabcolsep=0.175cm
    \begin{tabular}{lcccccccc}
        & \multicolumn{5}{c}{Out-of-Domain Accuracy (\%)} & \multicolumn{3}{c}{Avg Rank} \\
        \cmidrule(lr){2-6}
        Model & \rotatebox{45}{Signs} & \rotatebox{45}{MSCOCO} & \rotatebox{45}{MNIST} & \rotatebox{45}{CIFAR10} & \rotatebox{45}{CIFAR100} & In & Out & All \\
        \midrule
        RelationNet \cite{Sung2018_RelationNet} & 37.5\textpm0.9 & 27.4\textpm0.9 & - & - & - & 10.5 & 11.0 & 10.6 \\
        MatchingNet \cite{Vinyals16_MatchingNet} & 55.6\textpm1.1 & 28.8\textpm1.0 & - & - & - & 10.1 & 8.5 & 9.8 \\
        MAML \cite{Finn17_MAML} & 42.9\textpm1.3 & 29.4\textpm1.1 & - & - & - & 9.2 & 10.5 & 9.5 \\
        ProtoNet \cite{Snell17_Proto} & 46.5\textpm1.0 & 39.9\textpm1.1 & - & - & - & 8.2 & 9.5 & 8.5 \\
        ProtoMAML \cite{triantafillou2019meta} & 52.4\textpm1.1 & 41.7\textpm1.1 & - & - & - & 7.1 & 8.0 & 7.3 \\
        CNAPS \cite{Requeima19_CNAPS} & 60.2\textpm0.9 & 42.6\textpm1.1 & 92.7\textpm0.4 & 61.5\textpm0.7 & 50.1\textpm1.0 & 6.6 & 6.0 & 6.4 \\
        BOHB-E \cite{bohb-e-saikia2020optimized} & 57.5\textpm1.0 & 51.9\textpm1.0 & - & - & - & 6.4 & 4.0 & 5.9 \\
        TaskNorm \cite{Bronskill20_TaskNorm} & 67.0\textpm0.7 & 43.4\textpm1.0 & 92.3\textpm0.4 & 69.3\textpm0.8 & 54.6\textpm1.1 & 4.7 & 4.8 & 4.8 \\
        SUR \cite{dvornik20_SUR} & 70.4\textpm0.8 & \textbf{52.4\textpm1.1} & 94.3\textpm0.4 & 66.8\textpm0.9 & 56.6\textpm1.0 & 3.1 & 2.6 & 2.9 \\
        URT \cite{liu20_URT} & 69.4\textpm0.8 & \textbf{52.2\textpm1.1} & 94.8\textpm0.4 & 67.3\textpm0.8 & 56.9\textpm1.0 & \textbf{1.7} & 2.8 & 2.2 \\
        \midrule
        Simple CNAPS & 73.5\textpm0.7 & 46.2\textpm1.1 & 93.9\textpm0.4 & 74.3\textpm0.7 & 60.5\textpm1.0 & 3.4 & 3.0 & 3.2 \\
        Transductive CNAPS & \textbf{76.1\textpm0.7} & 48.7\textpm1.0 & \textbf{95.7\textpm0.3} & \textbf{75.7\textpm0.7} & \textbf{62.9\textpm1.0} & 2.1 & \textbf{1.6} & \textbf{1.9} \\
    \end{tabular}
    \caption{Few-shot classification on out-of-domain Meta-Dataset, MNIST, and CIFAR10/100 tasks. Error intervals correspond to 95\% confidence intervals, and bold values indicate statistically significant SoTA performance. Average rank is obtained by ranking methods on each dataset and averaging the ranks.}
    \label{tab:results:meta-dataset-2}
\end{table*}{}

\begin{table}
    \centering
    \scriptsize
    \tabcolsep=0.25cm
    \begin{tabular}{lcccccccccc}
        {} & {} & \multicolumn{4}{c}{mini-ImageNet Acc (\%)} & \multicolumn{4}{c}{tiered-ImageNet Acc (\%)} \\
        \cmidrule(lr){3-6}\cmidrule(lr){7-10}
        {} & {} & \multicolumn{2}{c}{5-\textit{way}} & \multicolumn{2}{c}{10-\textit{way}} & \multicolumn{2}{c}{5-\textit{way}} & \multicolumn{2}{c}{10-\textit{way}}\\
        \cmidrule(lr){3-4}\cmidrule(lr){5-6}\cmidrule(lr){7-8}\cmidrule(lr){9-10}
        Model & T? & 1-\textit{shot} & 5-\textit{shot} & 1-\textit{shot} & 5-\textit{shot} & 1-\textit{shot} & 5-\textit{shot} & 1-\textit{shot} & 5-\textit{shot} \\
        \midrule
        MAML \cite{finn2017model} & BN & 48.7 & 63.1 & 31.3 & 46.9 & 51.7 & 70.3 & 34.4 & 53.3 \\
        MAML+ \cite{DBLP:journals/corr/abs-1805-10002-tpn} & Yes & 50.8 & 66.2 & 31.8 & 48.2 & 53.2 & 70.8 & 34.8 & 54.7 \\
        Reptile \cite{DBLP:journals/corr/abs-1803-02999-reptile} & No & 47.1 & 62.7 & 31.1 & 44.7 & 49.0 & 66.5 & 33.7 & 48.0 \\
        Reptile+BN \cite{DBLP:journals/corr/abs-1803-02999-reptile} & BN & 49.9 & 66.0 & 32.0 & 47.6 & 52.4 & 71.0 & 35.3 & 52.0 \\
        ProtoNet \cite{Snell17_Proto} & No & 46.1 & 65.8 & 32.9 & 49.3 & 48.6 & 69.6 & 37.3 & 57.8 \\
        RelationNet \cite{sung2018learning} & BN &  51.4 & 67.0 & 34.9 & 47.9 & 54.5 & 71.3 & 36.3 & 58.0 \\
        TPN \cite{DBLP:journals/corr/abs-1805-10002-tpn} & Yes & 55.5 & 69.8 & 38.4 & 52.8 & 59.9 & 73.3 & 44.8 & 59.4 \\
        AttWeightGen \cite{DBLP:journals/corr/abs-1804-09458-dynamic} & No & 56.2 & 73.0 & - & - & - & - & - & - \\
        TADAM \cite{NIPS2018_7352-tadam} & No & 58.5 & 76.7 & - & - & - & - & - & - \\
        \midrule
        Simple CNAPS & No & 53.2 & 70.8 & 37.1 & 56.7 & 63.0 & 80.0 & 48.1 & 70.2 \\
        Transductive CNAPS & Yes & 55.6 & 73.1 & \textbf{42.8} & \textbf{59.6} & 65.9 & 81.8 & \textbf{54.6} & \textbf{72.5} \\
        \midrule
        LEO \cite{DBLP:journals/corr/abs-1807-05960-leo} & No & 61.8 & 77.6 & - & - & 66.3 & 81.4 & - & - \\
        \revision{MetaOptNet \cite{Lee2019_MetaOptNet}} & \revision{No} & \revision{62.6} & \revision{78.6} & \revision{-} & \revision{-} & \revision{66.0} & \revision{81.6} & \revision{-} & \revision{-} \\
        \revision{MetaBaseline \cite{Chen2021_MetaBaseline}} & \revision{No} & \revision{63.2} & \revision{79.3} & \revision{-} & \revision{-} & \revision{68.6} & \revision{83.7} & \revision{-} & \revision{-} \\
        \revision{FEAT \cite{Ye2020_FEAT}} & \revision{No} & \revision{66.8} & \revision{82.0} & \revision{-} & \revision{-} & \revision{70.8} & \revision{84.8} & \revision{-} & \revision{-} \\
        \revision{SimpleShot \cite{Wang2019_SimpleShot}} & \revision{No} & \revision{62.8} & \revision{80.0} & \revision{-} & \revision{-} & \revision{71.3} & \revision{86.6} & \revision{-} & \revision{-} \\
        \revision{RFS \cite{Tian2020_RFS}} & \revision{No} & \revision{64.8} & \revision{82.1} & \revision{-} & \revision{-} & \revision{71.5} & \revision{86.0} & \revision{-} & \revision{-} \\
        \revision{FRN \cite{Wertheimer2021_FRN}} & \revision{No} & \revision{66.5} & \revision{82.8} & \revision{-} & \revision{-} & \revision{71.2} & \revision{86.0} & \revision{-} & \revision{-} \\
        \revision{DeepEMD \cite{Zhang2020_DeepEmd}} & \revision{No} & \revision{\textbf{68.8}} & \revision{\textbf{84.1}} & \revision{-} & \revision{-} & \revision{74.3} & \revision{87.0} & \revision{-} & \revision{-} \\
        \revision{S2M2 \cite{Mangla2020_S2M2}} & \revision{No} & \revision{64.9} & \revision{83.2} & \revision{-} & \revision{-} & \revision{73.7} & \revision{88.6} & \revision{-} & \revision{-} \\
        \revision{LR+DC \cite{Yang2021_LRDC}} & \revision{No} & \revision{\textbf{68.6}} & \revision{82.9} & \revision{-} & \revision{-} & \revision{\textbf{78.2}} & \revision{\textbf{89.9}} & \revision{-} & \revision{-} \\
    \end{tabular}
    \caption{\revision{Few-shot visual classification results on 1/5-shot 5/10-way tasks on the mini/tiered-ImageNet benchmarks. See Table \ref{tab:results:mini-tiered-imagenet-with-error-intervals} for error intervals.}}
    \label{tab:results:mini-tiered-imagenet}
\end{table}

\begin{table}
    \centering
    \scriptsize
    \tabcolsep=0.15cm
    \begin{tabular}{lcccccccccc}
        {} & {} & \multicolumn{4}{c}{mini-ImageNet Acc (\%)} & \multicolumn{4}{c}{tiered-ImageNet Acc (\%)} \\
        \cmidrule(lr){3-6}\cmidrule(lr){7-10}
        {} & {} & \multicolumn{2}{c}{5-\textit{way}} & \multicolumn{2}{c}{10-\textit{way}} & \multicolumn{2}{c}{5-\textit{way}} & \multicolumn{2}{c}{10-\textit{way}}\\
        \cmidrule(lr){3-4}\cmidrule(lr){5-6}\cmidrule(lr){7-8}\cmidrule(lr){9-10}
        Model & Transductive & 1-\textit{shot} & 5-\textit{shot} & 1-\textit{shot} & 5-\textit{shot} & 1-\textit{shot} & 5-\textit{shot} & 1-\textit{shot} & 5-\textit{shot} \\
        \revision{RS-FSL \cite{Afham2021_RS-FSL}} & \revision{No} & \revision{65.3} & \revision{-} & \revision{-} & \revision{-} & \revision{-} & \revision{-} & \revision{-} & \revision{-} \\
        \revision{AmdimNet \cite{Chen2019_AmdimNet}} & \revision{No} & \revision{76.8} & \revision{91.0} & \revision{-} & \revision{-} & \revision{-} & \revision{-} & \revision{-} & \revision{-} \\
        \midrule
        \makecell[l]{Simple CNAPS \\ + FETI} & No & 77.4 & 90.3 & 63.5 & 83.1 & 71.4 & 86.0 & 57.1 & 78.5 \\
        \makecell[l]{Transductive CNAPS \\ + FETI} & Yes & \textbf{79.9} & \textbf{91.5} & \textbf{68.5} & \textbf{85.9} & \textbf{73.8} & \textbf{87.7} & \textbf{65.1} & \textbf{80.6} \\
    \end{tabular}
    \caption{\revision{Few-shot visual classification results on 1/5-shot 5/10-way tasks on mini/tiered-ImageNet with additional training data. For our models, ``FETI'' indicates that the feature extractor used was trained on ImageNet \cite{russakovsky2015imagenet} excluding classes in the test splits of mini/tiered-ImageNet. See Table \ref{tab:results:mini-tiered-imagenet-with-error-intervals-feti} for error intervals.}}
    \label{tab:results:mini-tiered-imagenet-with-extra-data}
\end{table}

\noindent
\textbf{Evaluation on Meta-Dataset: } In-domain, out-of-domain and overall rankings on Meta-Dataset are shown in Tables \ref{tab:results:meta-dataset-1} and \ref{tab:results:meta-dataset-2}. Following  \cite{Requeima19_CNAPS}, \revision{we pretrain the feature extractor on the Meta-Dataset specified training split of the ImageNet subset of Meta-Dataset (See Requeima et al.\cite{Requeima19_CNAPS}-C.1.1 for ResNet18 training details). Note that this excludes any examples or classes present in the test/validation sets of the Meta-Dataset's ImageNet split.} As shown in Table \ref{tab:results:meta-dataset-2}, Transductive CNAPS is able to establish state-of-the-art performance with an overall rank of 1.9 while Simple CNAPS ranks fourth with an average rank of 3.2.

\revision{\vspace{0.05in}
\noindent
\textbf{Evaluation on mini/tiered-ImageNet: } We consider two feature extractor training settings on these benchmarks. First, we use the feature extractor trained on the corresponding training split of the mini/tiered-ImageNet. As shown in Table \ref{tab:results:mini-tiered-imagenet}, on tiered-ImageNet, Transductive CNAPS achieves best accuracy on both 10-way settings with Simple CNAPS ranking second-best, as compared to other previous work that report results on these settings. On 5-way tiered-ImageNet settings, however, despite outperforming a number of major baselines, both methods trail behind a number of more recent works \cite{Lee2019_MetaOptNet, Chen2021_MetaBaseline, Ye2020_FEAT, Wang2019_SimpleShot, Tian2020_RFS, Wertheimer2021_FRN, Zhang2020_DeepEmd, Mangla2020_S2M2, Yang2021_LRDC} in classification accuracy. It's also interesting to note that performance of Simple and Transductive CNAPS generally ranks higher on the tiered-ImageNet benchmark as compared to mini-ImageNet. We attribute this difference in performance between mini-ImageNet and tiered-ImageNet to the fact that mini-ImageNet only provides 38,400 training examples, compared to 448,695 examples provided by tiered-ImageNet. This results in a lower performing ResNet-18 feature extractor (which is trained in a traditional supervised manner).

This hypothesis leads us to consider a second evaluation (denoted by ``FETI'', for ``Feature Extractor Trained with ImageNet'', in Table \ref{tab:results:mini-tiered-imagenet-with-extra-data}). In this model, we train the feature extractor with a much larger subset of ImageNet, which has been carefully selected to prevent any possible overlap (in examples or classes) with the test sets of mini/tiered-ImageNet. Both Simple and Transductive CNAPS are able to take advantage of the more example-rich feature extractor, resulting in substantially better performance across the board, even as compared to other baselines that employ additional data (such as RS-FSL which pre-trains on the full ImageNet). Furthermore, Transductive CNAPS outperforms Simple CNAPS by a large margin, even when using the same example-rich feature extractor; this demonstrates that leveraging additional query set information yields gains.}

\vspace{0.05in}
\noindent
\textbf{Performance vs. Class Shot: } In Figure \ref{fig:ratios-and-shots-all-ranges-10max}, we examine the relationship between class recall (i.e. accuracy among query examples belonging to the class itself) and the number of support examples in the class (shot). As shown, Simple CNAPS outperforms CNAPS with even as few as four support examples, indicating its ability to produce useful estimates of covariance from few examples. In addition, Transductive CNAPS is very effective when class shot is below 10, showing large average recall improvements, especially at the 1-shot level. However, as the class shot increases beyond 10, performance drops compared to Simple CNAPS. This suggests that soft k-means learning of cluster parameters can be effective when very few support examples are available. Conversely, in high-shot classes, transductive updates can act as distractors.

\begin{table*}[t]
    \centering
    \scriptsize
    \tabcolsep=0.125cm
    \begin{tabular}{lcccccccc}
        & \multicolumn{8}{c}{In-Domain Accuracy (\%)} \\
        \cmidrule(lr){2-9}
        Simple CNAPS + Metric & \rotatebox{45}{ImageNet} & \rotatebox{45}{Omniglot} & \rotatebox{45}{Aircraft} & \rotatebox{45}{Birds} & \rotatebox{45}{DTD} & \rotatebox{45}{QuickDraw} & \rotatebox{45}{Fungi} & \rotatebox{45}{Flower} \\
        \midrule
        Negative Dot Product & 48.0\textpm1.1 & 83.5\textpm0.9 & 73.7\textpm0.8 & 69.0\textpm1.0 & 66.3\textpm0.6 & 66.5\textpm0.9 & 39.7\textpm1.1 & 88.6\textpm0.5 \\
        Cosine Similarity & 51.3\textpm1.1 & 89.4\textpm0.7 & 80.5\textpm0.8 & 70.9\textpm1.0 & \textbf{69.7\textpm0.7} & 72.6\textpm0.9 & 41.9\textpm1.0 & 89.3\textpm0.6 \\
        Absolute Distance ($L_1$) & 53.6\textpm1.1 & 90.6\textpm0.6 & 81.0\textpm0.7 & 73.2\textpm0.9 & 61.1\textpm0.7 & 74.1\textpm0.8 & 47.0\textpm1.0 & 87.3\textpm0.6 \\
        Squared Euclidean (${L_2}^2$) & 53.9\textpm1.1 & 90.9\textpm0.6 & 81.8\textpm0.7 & 73.1\textpm0.9 & 64.4\textpm0.7 & 74.9\textpm0.8 & 45.8\textpm1.0 & 88.8\textpm0.5 \\
        \midrule
        Squared Mahalanobis & \textbf{58.6\textpm1.1} & 91.7\textpm0.6 & 82.4\textpm0.7 & 74.9\textpm0.8 & 67.8\textpm0.8 & 77.7\textpm0.7 & 46.9\textpm1.0 & 90.7\textpm0.5 \\
    \end{tabular}
    \caption{Performance of various metric ablations on Meta-Dataset (in-domain sub-datasets). Error intervals indicate 95\% confidence intervals, and bold values indicate statistically significant state of the art performance.}
    \label{tab:results:metric-ablations-1}
\end{table*}{}

\begin{table*}[t]
    \centering
    \scriptsize
    \tabcolsep=0.15cm
    \begin{tabular}{lcccccccc}
        & \multicolumn{5}{c}{Out-of-Domain Accuracy (\%)} & \multicolumn{3}{c}{Avg Acc.} \\
        \cmidrule(lr){2-6}
        Simple CNAPS + Metric & \rotatebox{45}{Signs} & \rotatebox{45}{MSCOCO} & \rotatebox{45}{MNIST} & \rotatebox{45}{CIFAR10} & \rotatebox{45}{CIFAR100} & In & Out & All \\
        \midrule
        Negative Dot Product & 48.0\textpm1.1 & 83.5\textpm0.9 & 73.7\textpm0.8 & 69.0\textpm1.0 & 66.3\textpm0.6 & 66.5\textpm0.9 & 39.7\textpm1.1 & 88.6\textpm0.5 \\
        Cosine Similarity & 51.3\textpm1.1 & 89.4\textpm0.7 & 80.5\textpm0.8 & 70.9\textpm1.0 & \textbf{69.7\textpm0.7} & 72.6\textpm0.9 & 41.9\textpm1.0 & 89.3\textpm0.6 \\
        Absolute Distance ($L_1$) & 53.6\textpm1.1 & 90.6\textpm0.6 & 81.0\textpm0.7 & 73.2\textpm0.9 & 61.1\textpm0.7 & 74.1\textpm0.8 & 47.0\textpm1.0 & 87.3\textpm0.6 \\
        Squared Euclidean (${L_2}^2$) & 53.9\textpm1.1 & 90.9\textpm0.6 & 81.8\textpm0.7 & 73.1\textpm0.9 & 64.4\textpm0.7 & 74.9\textpm0.8 & 45.8\textpm1.0 & 88.8\textpm0.5 \\
        \midrule
        Squared Mahalanobis & \textbf{58.6\textpm1.1} & 91.7\textpm0.6 & 82.4\textpm0.7 & 74.9\textpm0.8 & 67.8\textpm0.8 & 77.7\textpm0.7 & 46.9\textpm1.0 & 90.7\textpm0.5 \\
    \end{tabular}
    \caption{Performance of various metric ablations on Meta-Dataset (out-of-domain sub-datasets and overall averages). Error intervals indicate 95\% confidence intervals, and bold values indicate statistically significant state of the art performance.}
    \label{tab:results:metric-ablations-2}
\end{table*}{}

\begin{table*}[t]
    \centering
    \scriptsize
    \tabcolsep=0.15cm
    \begin{tabular}{lcccccccccccccccccc}
        & \multicolumn{8}{c}{In-Domain Accuracy (\%)} \\
        \cmidrule(lr){2-9}
        CNAPS Model & \rotatebox{45}{ImageNet} & \rotatebox{45}{Omniglot} & \rotatebox{45}{Aircraft} & \rotatebox{45}{Birds} & \rotatebox{45}{DTD} & \rotatebox{45}{QuickDraw} & \rotatebox{45}{Fungi} & \rotatebox{45}{Flower}\\
        \midrule
        GMM & 45.3\textpm1.0 & 88.0\textpm0.9 & 80.8\textpm0.8 & 71.4\textpm0.8 & 61.1\textpm0.7 & 70.7\textpm0.8 & 42.9\textpm1.0 & 88.1\textpm0.6 \\
        GMM-EM & 52.3\textpm1.0 & 92.0\textpm0.5 & \textbf{84.3\textpm0.6} & 75.2\textpm0.8 & 64.3\textpm0.7 & 72.6\textpm0.8 & 44.6\textpm1.0 & 90.8\textpm0.5 \\
        \midrule
        Transductive+ & 53.3\textpm1.1 & 92.3\textpm0.5 & 81.2\textpm0.7 & 75.0\textpm0.8 & \textbf{72.0\textpm0.7} & 74.8\textpm0.8 & 45.1\textpm1.0 & 92.1\textpm0.4 \\
        FEOT & \textbf{57.3\textpm1.1} & 90.5\textpm0.7 & 82.9\textpm0.7 & 74.8\textpm0.8 & 67.3\textpm0.8 & 76.3\textpm0.8 & 47.7\textpm1.0 & 90.5\textpm0.5 \\
        COT & \textbf{58.8\textpm1.1} & \textbf{95.2\textpm0.3} & \textbf{84.0\textpm0.6} & \textbf{76.4\textpm0.7} & 68.5\textpm0.8 & \textbf{77.8\textpm0.7} & \textbf{49.7\textpm1.0} & \textbf{92.7\textpm0.4} \\
        \midrule
        Simple & \textbf{58.6\textpm1.1} & 91.7\textpm0.6 & 82.4\textpm0.7 & 74.9\textpm0.8 & 67.8\textpm0.8 & 77.7\textpm0.7 & 46.9\textpm1.0 & 90.7\textpm0.5 \\
        Transductive & \textbf{58.8\textpm1.1} & 93.9\textpm0.4 & \textbf{84.1\textpm0.6} & \textbf{76.8\textpm0.8} & 69.0\textpm0.8 & \textbf{78.6\textpm0.7} & \textbf{48.8\textpm1.1} & 91.6\textpm0.4 \\
    \end{tabular}
    \caption{Performance of various ablations of Transductive and Simple CNAPS on Meta-Dataset (in-domain sub-datasets). Error intervals indicate 95\% confidence intervals, and bold values indicate statistically significant state of the art performance.}
    \label{tab:results:ablations-1}
\end{table*}{}

\begin{table*}[t]
    \centering
    \scriptsize
    \tabcolsep=0.225cm
    \begin{tabular}{lcccccccccccccccccc}
        & \multicolumn{5}{c}{Out-of-Domain Accuracy (\%)} & \multicolumn{3}{c}{Avg Acc.} \\
        \cmidrule(lr){2-6}
        \rotatebox{45}{Flower} & \rotatebox{45}{Signs} & \rotatebox{45}{MSCOCO} & \rotatebox{45}{MNIST} & \rotatebox{45}{CIFAR10} & \rotatebox{45}{CIFAR100} & In & Out & All \\
        \midrule
        GMM & 68.9\textpm0.7 & 37.2\textpm0.9 & 91.4\textpm0.5 & 64.5\textpm0.7 & 46.6\textpm0.9 & 68.5 & 61.7 & 65.9 \\
        GMM-EM & 71.4\textpm0.7 & 44.7\textpm0.9 & 93.0\textpm0.4 & 71.1\textpm0.7 & 56.4\textpm0.9 & 72.0 & 67.3 & 70.2 \\
        \midrule
        Transductive+ & 71.0\textpm0.8 & 44.0\textpm1.1 & 95.9\textpm0.3 & 71.1\textpm0.7 & 57.3\textpm1.1 & 73.2 & 67.9 & 71.2 \\
        FEOT & \textbf{75.8\textpm0.7} & \textbf{47.1\textpm1.1} & 94.9\textpm0.4 & 74.3\textpm0.8 & \textbf{61.2\textpm1.0} & 73.4 & 70.7 & 72.4 \\
        COT & 70.8\textpm0.7 & \textbf{47.3\textpm1.0} & 94.2\textpm0.4 & \textbf{75.2\textpm0.7} & \textbf{61.2\textpm1.0} & \textbf{75.4} & 69.7 & \textbf{73.2} \\
        \midrule
        Simple & 73.5\textpm0.7 & 46.2\textpm1.1 & 93.9\textpm0.4 & 74.3\textpm0.7 & 60.5\textpm1.0 & 73.8 & 69.7 & 72.2 \\
        Transductive & \textbf{76.1\textpm0.7} & \textbf{48.7\textpm1.0} & \textbf{95.7\textpm0.3} & \textbf{75.7\textpm0.7} & \textbf{62.9\textpm1.0} & \textbf{75.2} & \textbf{71.8} & \textbf{73.9} \\
    \end{tabular}
    \caption{Performance of various ablations of Transductive and Simple CNAPS on Meta-Dataset (out-of-domain sub-datasets and overall averages). Error intervals indicate 95\% confidence intervals, and bold values indicate statistically significant state of the art performance.}
    \label{tab:results:ablations-2}
\end{table*}{}

\begin{table}
    \centering
    \scriptsize
    \tabcolsep=0.25cm
    \begin{tabular}{lcccccc}
        {} & \multicolumn{2}{c}{MNIST} & \multicolumn{2}{c}{CIFAR10} & \multicolumn{2}{c}{CIFAR100} \\
        \cmidrule(lr){2-3}\cmidrule(lr){4-5}\cmidrule(lr){6-7}
        Method & Multi & Single & Multi & Single & Multi & Single \\
        \midrule
        SI \cite{Zenke17_ContinualSI} & 99.3 & 57.6 & - & - & 73.2 & 22.8 \\
        EWC \cite{Kirkpatrick16_EWC} & 99.3 & 55.8 & - & - & 72.8 & 23.1 \\
        VCL \cite{Nguyen18_VCL} & \makecell{98.5\textpm0.4} & - & - & - & - & - \\
        RWalk \cite{Chaudhry18_RWalk} & \textbf{99.3} & \textbf{82.5} & - & - & 74.2 & 34.0 \\
        \makecell[l]{CNAPS \cite{Requeima19_CNAPS} \\ + Moving Encoding} & \makecell{98.9\textpm0.2} & \makecell{80.9\textpm0.9} & - & - & \makecell{\textbf{76.0}\textbf{\textpm0.5}} & \makecell{\textbf{37.2}\textbf{\textpm0.6}} \\
        \midrule
        \makecell[l]{Simple CNAPS \\ + Moving Encoding} & \makecell{83.8\textpm3.2} & \makecell{19.2\textpm0.2} & \makecell{83.0\textpm3.0} & \makecell{18.1\textpm1.2}
        & \makecell{67.7\textpm0.8} & \makecell{12.1\textpm2.0} \\
        \makecell[l]{Simple CNAPS \\ + First Encoding} & \makecell{96.7\textpm0.3} & \makecell{69.5\textpm1.0} & \makecell{\textbf{87.4}\textbf{\textpm0.8}} & \makecell{\textbf{41.9}\textbf{\textpm1.0}} & \makecell{69.0\textpm0.7} & \makecell{34.2\textpm0.8} \\
        \makecell[l]{Simple CNAPS \\ + Averaging Encoding} & \makecell{95.2\textpm0.9} & \makecell{22.1\textpm2.2} & \makecell{\textbf{86.0}\textbf{\textpm1.8}} & \makecell{27.7\textpm4.3} & \makecell{68.6\textpm0.7} & \makecell{25.9\textpm2.8} \\
        \midrule
        \makecell[l]{Transductive CNAPS \\ + First Encoding} & \makecell{90.7\textpm1.4} & \makecell{9.3\textpm0.3} & \makecell{86.0\textpm0.5} & \makecell{10.0\textpm0.3} & \makecell{66.7\textpm 0.6} & \makecell{1.0\textpm 0.1} \\
        \makecell[l]{Transductive CNAPS \\ + Averaging Encoding} & \makecell{88.7\textpm0.5} & \makecell{9.2\textpm0.3} & \makecell{85.7\textpm0.7} & \makecell{9.9\textpm0.3} & \makecell{66.8\textpm0.9} & \makecell{1.0\textpm0.1} \\
    \end{tabular}
    \caption{``Out of the box'' continual learning performance on MNIST, CIFAR10 and CIFAR100. Tasks here are generated with 100-shots per category.}
    \label{tab:continual-learning-results-table}
\end{table}

\vspace{0.05in}
\noindent
\textbf{Classification-Time Soft K-means Clustering:} We use soft k-means iterative updates of means and covariance at test-time only. It is natural to consider training the feature adaptation network end-to-end through the soft k-means transduction procedure. We provide this comparison in the bottom-half of Tables \ref{tab:results:ablations-1} and \ref{tab:results:ablations-2}, with ``Transductive+'' denoting this variation. Iterative updates during training result in an average accuracy decrease of 2.5\%, which we conjecture to be due to training instabilities caused by applying this iterative algorithm early in training on noisy features.

\vspace{0.05in}
\noindent
\textbf{Transductive Feature Extraction vs. Classification:} \label{exp:feot-vs-cot} Our approach extends Simple CNAPS in two ways: improved adaptation of the feature extractor using a transductive task-encoding, and the soft k-means iterative estimation of class means and covariances.
We perform two ablations, ``Feature Extraction Only Transductive'' (FEOT) and ``Classification Only Transductive'' (COT), to independently assess the impact of these extensions. The results are presented in Tables \ref{tab:results:ablations-1} and \ref{tab:results:ablations-2}. As shown, both extensions outperform Simple CNAPS. The transductive task-encoding is especially effective on out-of-domain tasks whereas the soft k-mean learning of class parameters boosts accuracy on in-domain tasks. Transductive CNAPS is able to leverage the best of both worlds, allowing it to achieve statistically significant gains over Simple CNAPS overall.

\vspace{0.05in}
\noindent
\textbf{Comparison to Gaussian Mixture Models:} \label{exp:gaussian-mixture-models}
We consider two GMM-based ablations of our method where the log-determinant is introduced into the weight updates (using a uniform class prior $\pi_k = 1/K$). \revision{Note that with the exception of how these ablations produce class probabilities (Equation 1 vs. Equations 18/19), all other aspects of GMM-based ablations remain identical to that of Simple and Transductive CNAPS.} Results are shown in Tables \ref{tab:results:ablations-1} and \ref{tab:results:ablations-2} where GMM and GMM-EM correspond to the GMM-based ablations of Simple and Transductive CNAPS. As demonstrated, the GMM-based variations of our method and Simple CNAPS result in a notable 4-8\% loss in overall accuracy.

\vspace{0.05in}
\noindent
\textbf{Metric Ablation:}
To test the significance of our choice of Mahalanobis distance, we substitute it within our architecture with other distance metrics - absolute difference ($L_1$), squared Euclidean ($L_2^2$), cosine similarity and negative dot-product. Performance comparisons are shown in Tables \ref{tab:results:metric-ablations-1} and \ref{tab:results:metric-ablations-2}. We observe that using the Mahalanobis distance results in the best in-domain, out-of-domain, and overall average performance on all datasets.

\vspace{0.05in}
\noindent
\textbf{Active Learning:}
We present active learning results on CIFAR10 and OMNIGLOT in Figure \ref{fig:active-learning-figure}. As shown, Transductive CNAPS and Simple CNAPS both benefit from uncertainty-based label aquisition thanks to having well-calibrated measures of uncertainty. It is interesting to note that the margin of performance gain over random selection is larger in Simple CNAPS. This can suggest that Transductive CNAPS is able to exploit much of the unlabelled information through transductive learning. It may also imply that its measure of uncertainty is less calibrated as soft-label class parameter estimates can be more noisy. \revision{We also compare both methods against fixed feature extractor (``FFE'') baselines. In these baselines, the adaptation networks are turned off and the pre-trained ResNet18 is used to produce feature vectors without any task-specific adaptation. As demonstrated, this results is lower performance across the board, signifying the role of conditional adaptation networks. Furthermore, the gap between Transductive and Simple FFE is considerably smaller, indicating that the transductive feature extractor adaptation module in Transductive CNAPS is responsible for a considerable proportion of its performance gain over Simple CNAPS.}

\vspace{0.05in}
\noindent
\textbf{Continual Learning:}
We evaluate continual learning on MNIST, CIFAR10, and CIFAR100 for both multi-head and single-head learning. In the multi-head setting, each task is assigned a separate classification head that focuses on the particular task's categories. In our work, this corresponds to only calculate distances to classes that are present in the task. In the single-head case, by contrast, we evaluate with respect on all current and previously seen classes.

As noted in Table \ref{tab:continual-learning-results-table}, while neither variation is able to achieve state of the art results on MNIST and CIFAR100, we observe that the task-encoding is extremely important in achieving competitive performance. We explore this further in Figure \ref{fig:continual-learning-charts}. As shown when using ``Moving Encoding'', the performance on previous tasks is completely lost in the single-head setting and significantly reduced in the multi-head case. This suggests that Simple CNAPS and Transductive CNAPS produce vastly different manifolds depending on the task at hand, and previously estimated class parameters may no more be valid or useful in these new feature spaces. This is further supported by the evidence we see when using ``First Encoding''. As shown, not only the best performance is achieved here, but accuracy on previously learned classes is substantially greater even in the single-head setting. Unsurprisingly, the best performance is always scene at the query of the first task, as the feature space has been adapted to maximize accuracy on that task. This shows the importance of adapting the feature space to new tasks, which in part motivates the use of the ``Averaging Encoding'' variation as a balance between the two. However, as we see empirically, the differences in the feature space manifolds generated with the averaged encoding still result in major loss of performance in the single head, although in the multi-head setting, performance matching that of ``First Encoding'' is achieved. It is interesting to note that in the single-head ``Averaging Encoding'' variation, previous task performance tends to improve as more tasks are seen. This is in part due to the fact that with more tasks, the average encoding will be more stable and less likely to change substantially with a new task, thus providing a more stable feature space across the subsequent tasks. \revision{Overall, ``out of the box'' continual learning remains an open question, even within the specific context of Simple and Transductive CNAPS models.}

\section{Discussion}

We propose two meta-learned few-shot image classification approaches, Simple and Transductive CNAPS, that focus on effective estimation of class-wise cluster parameters, including covariance. These models are able to accomplish strong performance on Meta-Dataset, mini-ImageNet and tiered-ImageNet benchmarks. We further study our methods in paradigms of active and continual learning, proposing extensions for undertaking such tasks.

Future research on better continual learning of the task-wise feature manifolds in both models can be beneficial in boosting the performance. Additional research on generating support examples through learnable augmentations, better cluster refinement algorithms, and further theoretical understanding of the Mahalanobis-based classifiers in our methods are among the many very interesting avenues for potentially fruitful future research.

\section{Acknowledgments}
We acknowledge the support of the Natural Sciences and Engineering Research Council of Canada (NSERC), the Canada Research Chairs (CRC) Program, the Canada CIFAR AI Chairs Program, Compute Canada, Intel, and DARPA under its D3M and LWLL programs. Additionally, this material is based upon work supported by the United States Air Force under Contract No.
FA8750-19-C-0515.

\newpage
\appendix

\section{Benchmarks and Training}
\label{appendix:benchmark-and-training}

\subsection{Meta-Dataset}
\label{appendix:benchmark-and-training:meta-dataset}

\begin{figure}[t]
    \centering
    \subfloat[Number of Tasks vs. Ways]{\label{fig:ways-frequency}{\includegraphics[width=0.48\textwidth]{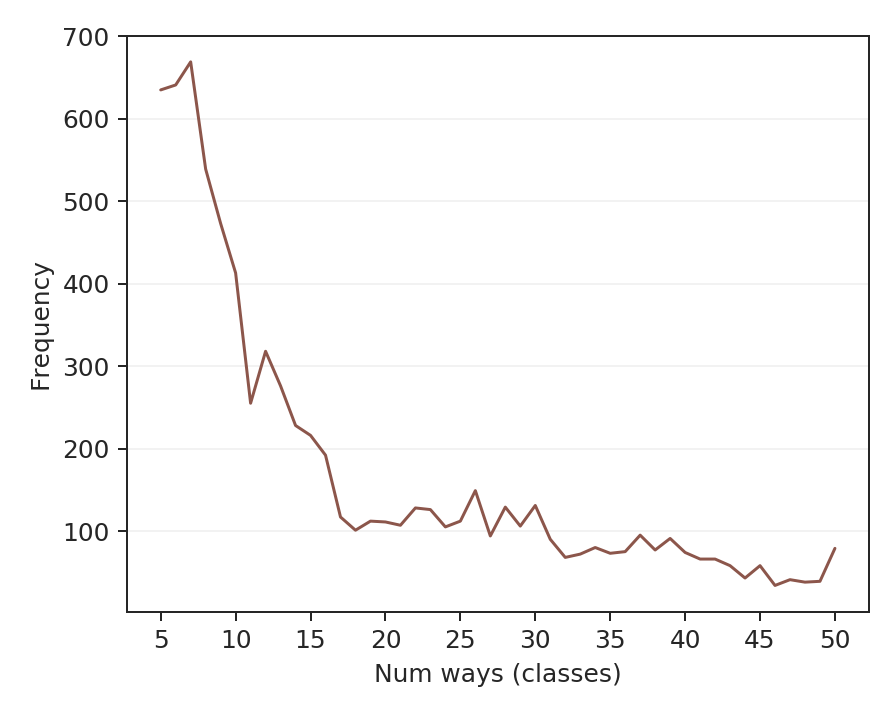} \label{fig:ways-vs-freq}}}
    \subfloat[Number of Classes vs. Shots]{\label{fig:shot-frequency}{\includegraphics[width=0.48\textwidth]{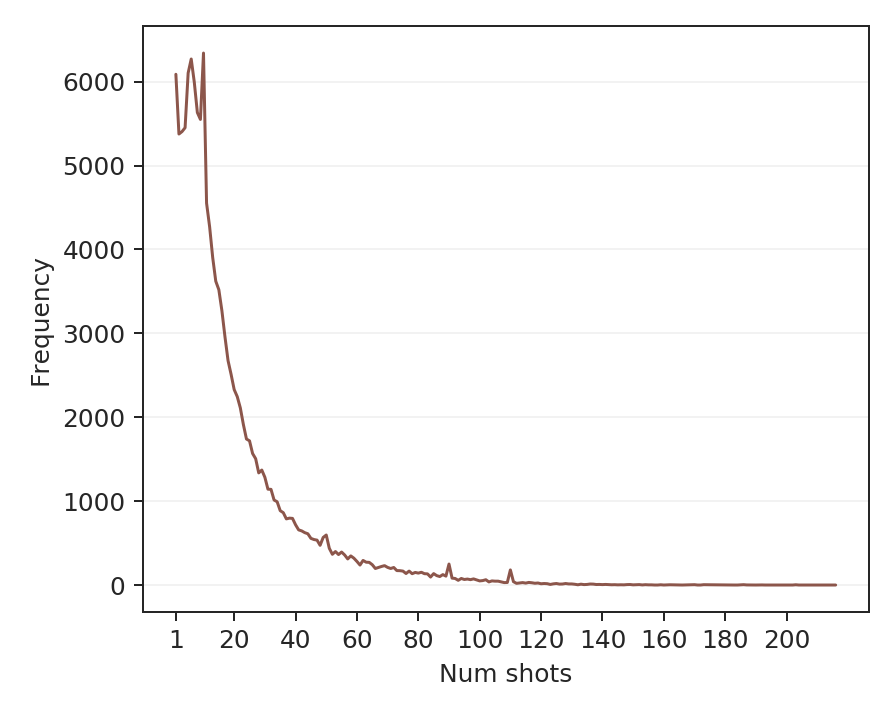} \label{fig:shots-vs-freq}}}%
    \caption{Test-time task-way and class-shot frequency graphs. As shown, most tasks have fewer than 10 classes (way) and most classes have less than 20 support examples (shot).}
\end{figure}

A brief description of the sampling procedure used in the Meta-Dataset setting is already provided in Section \ref{experiments:benchmarks}.
This sampling procedure, however, comes with additional specifications that are uniform across all tasks (such as count enforcing) and dataset specific details such as considering the class hierarchy in ImageNet tasks. 
The full algorithm for sampling is outlined in \cite{triantafillou2019meta}, and we refer the interested reader to Section 3.2 in \cite{triantafillou2019meta} for complete details. This procedure results in a task distribution where most tasks have fewer than 10 classes and each class has fewer than 20 support examples. The task frequency relative to the number of classes is presented in Figure \ref{fig:ways-frequency}, and the class frequency as compared to the class shot is presented in Figure \ref{fig:shot-frequency}. The query set contains between 1 and 10 (inclusive) examples per class for all tasks; fewer than 10 query examples occur only when there are not enough total images to support 10 query examples.

\subsection{mini/tiered-ImageNet}
\label{appendix:benchmark-and-training:minitiered}
Task sampling across both mini-ImageNet and tiered-ImageNet first starts by defining a constant number of ways and shots that will be used for each generated task. For a $L$-shot $K$-way problem setting, first $K$ classes are sampled from the dataset with uniform probability. Then, for each sampled class, $L$ of the class images are sampled with uniform probability and used as the support examples for the class. In addition, 10 query images (distinct from the support images) are sampled per class.

\begin{figure}
  \begin{center}
    \includegraphics[width=0.95\textwidth]{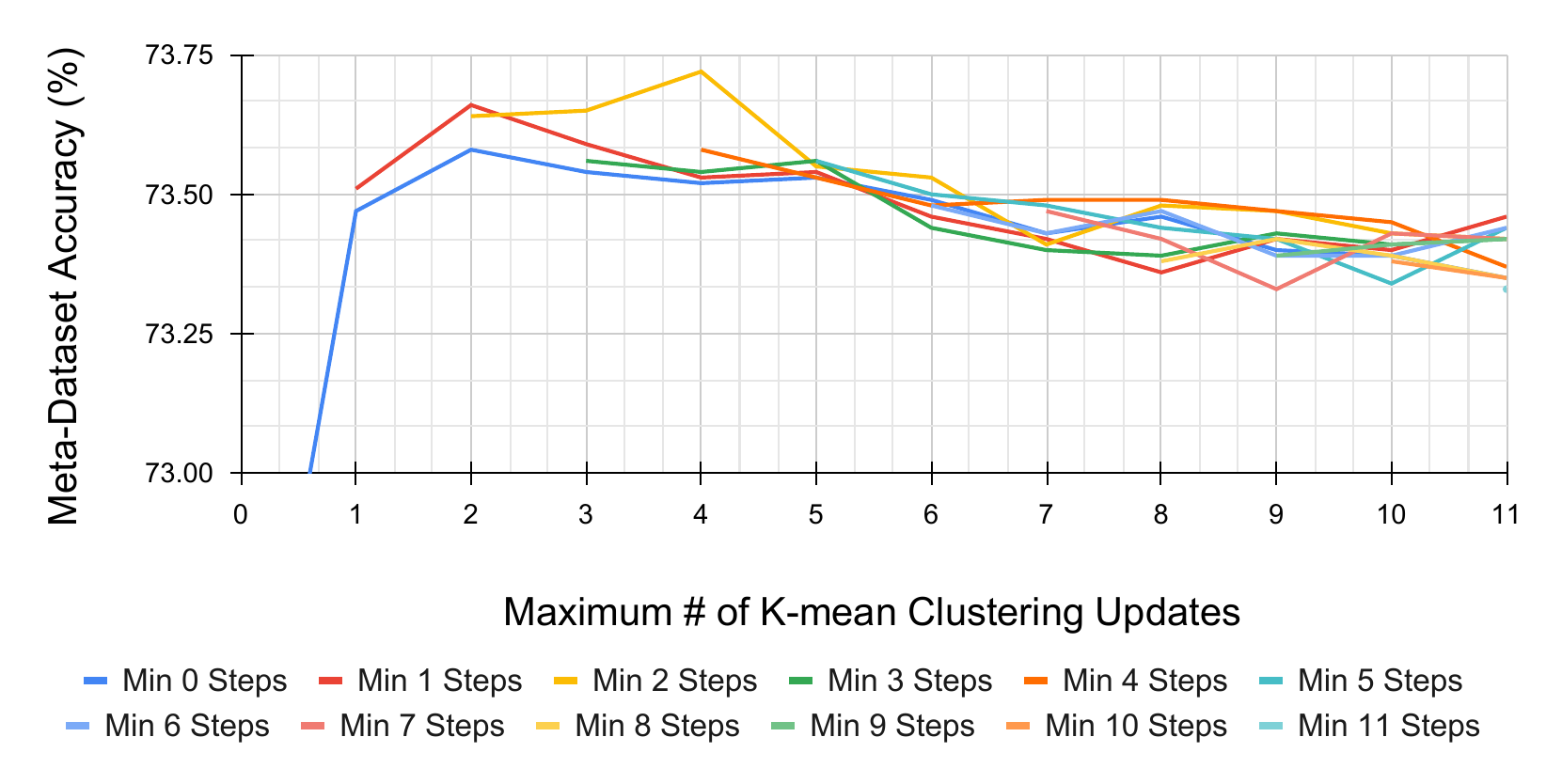}
  \end{center}
  \caption{\revision{Evaluating Transductive CNAPS on Meta-Dataset with different minimum and maximum number of steps. Performances reported stem from five run averages.}}
  \label{fig:max-and-min}
\end{figure}

\begin{table*}
    \centering
    \scriptsize
    \tabcolsep=0.085cm
    \begin{tabular}{lcccccccccc}
        {} & {} & \multicolumn{4}{c}{mini-ImageNet Accuracy (\%)} & \multicolumn{4}{c}{tiered-ImageNet Accuracy (\%)} \\
        \cmidrule(lr){3-6}\cmidrule(lr){7-10}
        {} & {} & \multicolumn{2}{c}{5-\textit{way}} & \multicolumn{2}{c}{10-\textit{way}} & \multicolumn{2}{c}{5-\textit{way}} & \multicolumn{2}{c}{10-\textit{way}}\\
        \cmidrule(lr){3-4}\cmidrule(lr){5-6}\cmidrule(lr){7-8}\cmidrule(lr){9-10}
        Model & Transductive & 1-\textit{shot} & 5-\textit{shot} & 1-\textit{shot} & 5-\textit{shot} & 1-\textit{shot} & 5-\textit{shot} & 1-\textit{shot} & 5-\textit{shot} \\
        \midrule
        MAML \cite{finn2017model} & BN & 48.7\textpm1.8 & 63.1\textpm0.9 & 31.3\textpm1.1 & 46.9\textpm1.2 & 51.7\textpm1.8 & 70.3\textpm1.7 & 34.4\textpm1.2 & 53.3\textpm1.3 \\
        MAML+ \cite{DBLP:journals/corr/abs-1805-10002-tpn} & Yes & 50.8\textpm1.8 & 66.2\textpm1.8 & 31.8\textpm0.4 & 48.2\textpm1.3 & 53.2\textpm1.8 & 70.8\textpm1.8 & 34.8\textpm1.2 & 54.7\textpm1.3 \\
        Reptile \cite{DBLP:journals/corr/abs-1803-02999-reptile} & No & 47.1\textpm0.3 & 62.7\textpm0.4 & 31.1\textpm0.3 & 44.7\textpm0.3 & 49.0\textpm0.2 & 66.5\textpm0.2 & 33.7\textpm0.3 & 48.0\textpm0.3 \\
        Reptile+BN \cite{DBLP:journals/corr/abs-1803-02999-reptile} & BN & 49.9\textpm0.3 & 66.0\textpm0.6 & 32.0\textpm0.3 & 47.6\textpm0.3 & 52.4\textpm0.2 & 71.0\textpm0.2 & 35.3\textpm0.3 & 52.0\textpm0.3 \\
        ProtoNet \cite{Snell17_Proto} & No & 46.1\textpm0.8 & 65.8\textpm0.7 & 32.9\textpm0.5 & 49.3\textpm0.4 & 48.6\textpm0.9 & 69.6\textpm0.7 & 37.3\textpm0.6 & 57.8\textpm0.5 \\
        RelationNet \cite{sung2018learning} & BN &  51.4\textpm0.8 & 67.0\textpm0.7 & 34.9\textpm0.5 & 47.9\textpm0.4 & 54.5\textpm0.9 & 71.3\textpm0.8 & 36.3\textpm0.6 & 58.0\textpm0.6 \\
        TPN \cite{DBLP:journals/corr/abs-1805-10002-tpn} & Yes & 55.5\textpm0.8 & 69.8\textpm0.7 & 38.4\textpm0.5 & 52.8\textpm0.4 & 59.9\textpm0.9 & 73.3\textpm0.7 & 44.8\textpm0.6 & 59.4\textpm0.5 \\
        AttWeightGen \cite{DBLP:journals/corr/abs-1804-09458-dynamic} & No & 56.2\textpm0.9 & 73.0\textpm0.6 & - & - & - & - & - & - \\
        TADAM \cite{NIPS2018_7352-tadam} & No & 58.5\textpm0.3 & 76.7\textpm0.3 & - & - & - & - & - & - \\
        \midrule
        Simple CNAPS & No & 53.2\textpm0.9 & 70.8\textpm0.7 & 37.1\textpm0.5 & 56.7\textpm0.5 & 63.0\textpm1.0 & 80.0\textpm0.8 & 48.1\textpm0.7 & 70.2\textpm0.6 \\
        Transductive CNAPS & Yes & 55.6\textpm0.9 & 73.1\textpm0.7 & \textbf{42.8\textpm0.7} & \textbf{59.6\textpm0.5} & 65.9\textpm1.0 & 81.8\textpm0.7 & \textbf{54.6\textpm0.8} & \textbf{72.5\textpm0.6} \\
        \midrule
        LEO \cite{DBLP:journals/corr/abs-1807-05960-leo} & No & 61.8\textpm0.1 & 77.6\textpm0.1 & - & - & 66.3\textpm0.1 & 81.4\textpm0.1 & - & - \\
        \revision{MetaOptNet \cite{Lee2019_MetaOptNet}} & \revision{No} & \revision{62.6\textpm0.6} & \revision{78.6\textpm0.5} & \revision{-} & \revision{-} & \revision{66.0\textpm0.7} & \revision{81.6\textpm0.5} & \revision{-} & \revision{-} \\
        \revision{MetaBaseline \cite{Chen2021_MetaBaseline}} & \revision{No} & \revision{63.2\textpm0.2}  & \revision{79.3\textpm0.2} & \revision{-} & \revision{-} & \revision{68.6\textpm0.3} & \revision{83.7\textpm0.2} & \revision{-} & \revision{-} \\
        \revision{FEAT \cite{Ye2020_FEAT}} & \revision{No} & \revision{66.8\textpm0.2} & \revision{82.0\textpm0.1} & \revision{-} & \revision{-} & \revision{70.8\textpm0.2} & \revision{84.8\textpm0.2} & \revision{-} & \revision{-} \\
        \revision{SimpleShot \cite{Wang2019_SimpleShot}} & \revision{No} & \revision{62.8\textpm0.2} & \revision{80.0\textpm0.1} & \revision{-} & \revision{-} & \revision{71.3\textpm0.2} & \revision{86.6\textpm0.2} & \revision{-} & \revision{-} \\
        \revision{RFS \cite{Tian2020_RFS}} & \revision{No} & \revision{64.8\textpm0.6} & \revision{82.1\textpm0.4} & \revision{-} & \revision{-} & \revision{71.5\textpm0.7} & \revision{86.0\textpm0.5} & \revision{-} & \revision{-} \\
        \revision{FRN \cite{Wertheimer2021_FRN}} & \revision{No} & \revision{66.5\textpm0.2} & \revision{82.8\textpm0.1} & \revision{-} & \revision{-} & \revision{71.2\textpm0.2} &  \revision{86.0\textpm0.2} & \revision{-} & \revision{-} \\
        \revision{DeepEMD \cite{Zhang2020_DeepEmd}} & \revision{No} & \revision{\textbf{68.8\textpm0.3}} & \revision{\textbf{84.1\textpm0.5}} & \revision{-} & \revision{-} & \revision{74.3\textpm0.3} & \revision{87.0\textpm0.6} & \revision{-} & \revision{-} \\
        \revision{S2M2 \cite{Mangla2020_S2M2}} & \revision{No} & \revision{64.9\textpm0.2} & \revision{83.2\textpm0.1} & \revision{-} & \revision{-} & \revision{73.7\textpm0.2} & \revision{88.6\textpm0.1} & \revision{-} & \revision{-} \\
        \revision{LR+DC \cite{Yang2021_LRDC}} & \revision{No} & \revision{\textbf{68.6\textpm0.6}} & \revision{82.9\textpm0.4} & \revision{-} & \revision{-} & \revision{\textbf{78.2\textpm0.3}} &  \revision{\textbf{89.9\textpm0.4}} & \revision{-} & \revision{-} \\
    \end{tabular}
    \caption{\revision{Few-shot visual classification results on 1/5-shot 5/10-way few-shot tasks on mini/tiered-ImageNet. Error intervals denote 95\% confidence interval.}}
    \label{tab:results:mini-tiered-imagenet-with-error-intervals}
\end{table*}

\begin{table*}
    \centering
    \scriptsize
    \tabcolsep=0.085cm
    \begin{tabular}{lcccccccccc}
        {} & {} & \multicolumn{4}{c}{mini-ImageNet Accuracy (\%)} & \multicolumn{4}{c}{tiered-ImageNet Accuracy (\%)} \\
        \cmidrule(lr){3-6}\cmidrule(lr){7-10}
        {} & {} & \multicolumn{2}{c}{5-\textit{way}} & \multicolumn{2}{c}{10-\textit{way}} & \multicolumn{2}{c}{5-\textit{way}} & \multicolumn{2}{c}{10-\textit{way}}\\
        \cmidrule(lr){3-4}\cmidrule(lr){5-6}\cmidrule(lr){7-8}\cmidrule(lr){9-10}
        Model & T? & 1-\textit{shot} & 5-\textit{shot} & 1-\textit{shot} & 5-\textit{shot} & 1-\textit{shot} & 5-\textit{shot} & 1-\textit{shot} & 5-\textit{shot} \\
        \revision{RS-FSL \cite{Afham2021_RS-FSL}} & \revision{No} & \revision{65.3\textpm0.8} & \revision{-} & \revision{-} & \revision{-} & \revision{-} & \revision{-} & \revision{-} & \revision{-} \\
        \revision{AmdimNet \cite{Chen2019_AmdimNet}} & \revision{No} & \revision{76.8\textpm0.2} & \revision{91.0\textpm0.1} & \revision{-} & \revision{-} & \revision{-} & \revision{-} & \revision{-} & \revision{-} \\
        \midrule
        Simple CNAPS \cite{bateni2019improved} + FETI & No & 77.4\textpm0.8 & 90.3\textpm0.4 & 63.5\textpm0.6 & 83.1\textpm0.4 & 71.4\textpm1.0 & 86.0\textpm0.6 & 57.1\textpm0.7 & 78.5\textpm0.5 \\
        Transductive CNAPS + FETI & Yes & \textbf{79.9\textpm0.8} & \textbf{91.5\textpm0.4} & \textbf{68.5\textpm0.6} & \textbf{85.9\textpm0.3} & \textbf{73.8\textpm1.0} & \textbf{87.7\textpm0.6} & \textbf{65.1\textpm0.8} & \textbf{80.6\textpm0.5} \\
    \end{tabular}
    \caption{\revision{Few-shot visual classification results on 1/5-shot 5/10-way few-shot on mini/tiered-ImageNet with additional training data. For CNAP-based model, ``FETI'' indicates that the feature extractor used has been trained on ImageNet \cite{russakovsky2015imagenet} exluding classes within the test splits of mini/tiered-ImageNet. Error intervals denote 95\% confidence interval.}}
    \label{tab:results:mini-tiered-imagenet-with-error-intervals-feti}
\end{table*}

\begin{table*}[t]
    \centering
    \scriptsize
    \tabcolsep=0.08cm
    \begin{tabular}{lccccccccccccccccc}
        {} & {} & {} & \multicolumn{6}{c}{Out-of-Domain Accuracy (\%)} \\
        \cmidrule(lr){4-10}
        Model & \rotatebox{45}{Backbone} & \rotatebox{45}{ImageNet} & \rotatebox{45}{Omniglot} & \rotatebox{45}{Aircraft} & \rotatebox{45}{Birds} & \rotatebox{45}{DTD} & \rotatebox{45}{QuickDraw} & \rotatebox{45}{Fungi} & \rotatebox{45}{Flower} \\
        \midrule
        \revision{RelationNet \cite{Sung2018_RelationNet}} & \revision{ResNet18} & \revision{34.7\textpm1.0} & \revision{45.3\textpm1.4} & \revision{40.7\textpm0.8} & \revision{49.5\textpm1.0} & \revision{53.0\textpm0.7} & \revision{43.3\textpm1.1} & \revision{30.5\textpm1.0} & \revision{68.7\textpm0.8} \\
        \revision{MatchingNet \cite{Vinyals16_MatchingNet}} & \revision{ResNet18} & \revision{45.0\textpm1.1} & \revision{52.3\textpm1.3} & \revision{49.0\textpm0.9} & \revision{62.2\textpm0.9} & \revision{64.1\textpm0.8} & \revision{42.9\textpm1.1} & \revision{34.0\textpm1.0} & \revision{80.1\textpm0.7} \\
        \revision{ProtoNet \cite{Snell17_Proto}} & \revision{ResNet18} & \revision{50.5\textpm1.1} & \revision{60.0\textpm1.3} & \revision{53.1\textpm1.0} & \revision{68.8\textpm1.0} & \revision{66.6\textpm0.8} & \revision{49.0\textpm1.1} & \revision{39.7\textpm1.1} & \revision{85.3\textpm0.8} \\
        \revision{MAML \cite{Finn17_MAML}} & \revision{ResNet18} & \revision{45.5±1.1} & \revision{55.5\textpm1.5} & \revision{56.2\textpm1.1} & \revision{63.6\textpm1.1} & \revision{68.0\textpm0.8} & \revision{44.0\textpm1.3} & \revision{32.1\textpm1.1} & \revision{81.7\textpm0.8} \\
        \revision{ProtoMAML \cite{triantafillou2019meta}} & \revision{ResNet18} & \revision{49.5\textpm1.0} & \revision{63.4\textpm1.3} & \revision{56.0\textpm1.0} & \revision{\textbf{68.7\textpm1.0}} &	\revision{66.5\textpm0.8} & \revision{51.5\textpm1.0} & \revision{40.0\textpm1.1} & \revision{87.1\textpm0.7} \\
        \revision{BOHB-E \cite{bohb-e-saikia2020optimized}} & \revision{ResNet18} & \revision{51.9\textpm1.0} & \revision{67.6\textpm1.2} & \revision{54.1\textpm0.9} & \revision{70.7\textpm0.9} & \revision{68.3\textpm0.8} & \revision{50.3\textpm1.0} & \revision{41.4\textpm1.1} & \revision{87.3\textpm0.6} \\
        \revision{ProtoNet (Large) \cite{Doersch2020_CrossTransformers}} & \revision{ResNet34} & \revision{53.7\textpm1.1} & \revision{68.5\textpm1.3} & \revision{58.0\textpm1.0} & \revision{74.1\textpm0.9} & \revision{68.8\textpm0.8} & \revision{53.3\textpm1.0} & \revision{40.7\textpm1.1} & \revision{87.0\textpm0.7} \\
        \revision{CTX \cite{Doersch2020_CrossTransformers}} & \revision{ResNet34} & \revision{\textbf{62.8\textpm1.0}} & \revision{\textbf{82.2\textpm1.0}} & \revision{\textbf{79.5\textpm0.9}} & \revision{\textbf{80.6\textpm0.9}} & \revision{\textbf{75.6\textpm0.6}} & \revision{\textbf{72.7\textpm0.8}} & \revision{\textbf{51.6\textpm1.1}} & \revision{\textbf{95.3\textpm0.4}} \\
        \midrule
        \revision{Simple CNAPS} & \revision{ResNet18} & \revision{54.8\textpm1.2} & \revision{62.0\textpm1.3} & \revision{49.2\textpm0.9} & \revision{66.5\textpm1.0} & \revision{71.6\textpm0.7} & \revision{56.6\textpm1.0} & \revision{37.5\textpm1.2} & \revision{82.1\textpm0.9} \\
        \revision{Transductive CNAPS} & \revision{ResNet18} & \revision{54.1\textpm1.1} & \revision{62.9\textpm1.3} & \revision{48.4\textpm0.9} & \revision{67.3\textpm0.9} & \revision{72.5\textpm0.7} & \revision{58.0\textpm1.0} & \revision{37.7\textpm1.1} & \revision{82.8\textpm0.8} \\
    \end{tabular}
    \caption{\revision{Few-shot classification on Meta-Dataset (ImageNet, Omniglot, Aircraft, Birds, DTD, QuickDraw, Fungi, Flower) with ImageNet-only training (all remaining datasets were kept for out-of-domain evaluation). Error intervals correspond to 95\% confidence intervals, and bold values indicate statistically significant SoTA performance. *Accuracy averages for Simple and Transductive CNAPS exclude MNIST, CIFAR10 and CIFAR100 as other baselines do not report performance on these datasets.}}
    \label{tab:results:meta-dataset-imagenet-only-1}
\end{table*}{}

\begin{table*}[t]
    \centering
    \scriptsize
    \tabcolsep=0.15cm
    \begin{tabular}{lccccccccccccccccc}
        {} & {} & \multicolumn{5}{c}{Out-of-Domain Accuracy (\%)} \\
        \cmidrule(lr){3-7}
        Model & \rotatebox{45}{Backbone} & \rotatebox{45}{Signs} & \rotatebox{45}{MSCOCO} & \rotatebox{45}{MNIST} & \rotatebox{45}{CIFAR10} & \rotatebox{45}{CIFAR100} & In & Out & All \\
        \midrule
        \revision{RelationNet \cite{Sung2018_RelationNet}} & \revision{ResNet18} & \revision{33.7\textpm1.0} & \revision{29.1\textpm1.0} & \revision{-} & \revision{-} & \revision{-} & \revision{34.7} & \revision{43.8} & \revision{42.9} \\
        \revision{MatchingNet \cite{Vinyals16_MatchingNet}} & \revision{ResNet18} & \revision{47.8\textpm1.1} & \revision{35.0\textpm1.0} & \revision{-} & \revision{-} & \revision{-} & \revision{45.0} & \revision{51.9} & \revision{51.2} \\
        \revision{ProtoNet \cite{Snell17_Proto}} & \revision{ResNet18} & \revision{47.1\textpm1.1} & \revision{41.0\textpm1.1} & \revision{-} & \revision{-} & \revision{-} & \revision{50.5} & \revision{56.7} & \revision{56.1} \\
        \revision{MAML \cite{Finn17_MAML}} & \revision{ResNet18} & \revision{50.9\textpm1.5} & \revision{35.3\textpm1.2} & \revision{-} & \revision{-} & \revision{-} & \revision{45.5} & \revision{54.2} & \revision{53.3} \\
        \revision{ProtoMAML \cite{triantafillou2019meta}} & \revision{ResNet18} & \revision{48.8\textpm1.1} & \revision{43.7\textpm1.1} & \revision{-} & \revision{-} & \revision{-} & \revision{49.5} & \revision{58.4} & \revision{57.5} \\
        \revision{BOHB-E \cite{bohb-e-saikia2020optimized}} & \revision{ResNet18} & \revision{51.8\textpm1.0} & \revision{48.0\textpm1.0} & \revision{-} & \revision{-} & \revision{-} & \revision{51.9} & \revision{59.9} & \revision{59.2} \\
        \revision{ProtoNet (Large) \cite{Doersch2020_CrossTransformers}} & \revision{ResNet34} & \revision{58.1\textpm1.0} & \revision{41.7\textpm1.1} & \revision{-} & \revision{-} & \revision{-} & \revision{53.7} & \revision{61.1} & \revision{60.4} \\
        \revision{CTX \cite{Doersch2020_CrossTransformers}} & \revision{ResNet34} & \revision{\textbf{82.6\textpm0.8}} & \revision{\textbf{59.9\textpm1.0}} & \revision{-} & \revision{-} & \revision{-} & \revision{\textbf{62.8}} & \revision{\textbf{75.6}} & \revision{\textbf{74.3}} \\
        \midrule
        \revision{Simple CNAPS} & \revision{ResNet18} & \revision{63.1\textpm1.1} & \revision{45.8\textpm1.0} & \revision{81.2\textpm0.6} & \revision{\textbf{69.9\textpm0.8}} & \revision{\textbf{59.4\textpm1.0}} & \revision{54.8*} & \revision{59.4*} & \revision{58.9*} \\
        \revision{Transductive CNAPS} & \revision{ResNet18} & \revision{61.8\textpm1.1} & \revision{45.8\textpm1.0} & \revision{\textbf{83.9\textpm0.7}} & \revision{\textbf{68.9\textpm0.8}} & \revision{\textbf{60.0\textpm1.1}} & \revision{54.1*} & \revision{59.7*} & \revision{59.1*} \\
    \end{tabular}
    \caption{\revision{Few-shot classification on Meta-Dataset (Signs, MSCOCO), MNIST, and CIFAR10/100 with ImageNet-only training (all remaining datasets were kept for out-of-domain evaluation. Error intervals correspond to 95\% confidence intervals, and bold values indicate statistically significant SoTA performance. *Accuracy averages for Simple/Transductive CNAPS exclude MNIST, CIFAR10 and CIFAR100 as other baselines do not report performance on these datasets.}}
    \label{tab:results:meta-dataset-imagenet-only-2}
\end{table*}{}

\begin{table*}[t]
    \centering
    \scriptsize
    \tabcolsep=0.15cm
    \begin{tabular}{lcccccccccccccccccc}
        & \multicolumn{8}{c}{In-Domain Accuracy (\%)} \\
        \cmidrule(lr){2-9}
        Simple CNAPS + Metric & \rotatebox{45}{ImageNet} & \rotatebox{45}{Omniglot} & \rotatebox{45}{Aircraft} & \rotatebox{45}{Birds} & \rotatebox{45}{DTD} & \rotatebox{45}{QuickDraw} & \rotatebox{45}{Fungi} & \rotatebox{45}{Flower} \\
        \midrule
        \revision{Root Riemannian*} & \revision{48.2\textpm1.0} & \revision{88.4\textpm0.7} & \revision{78.0\textpm0.7} & \revision{70.2\textpm0.8} & \revision{61.7\textpm0.7} & \revision{74.5\textpm0.7} & \revision{44.3\textpm1.0} & \revision{79.9\textpm0.7} \\
        \revision{Squared Mahalanobis} & \revision{\textbf{58.6\textpm1.1}} & \revision{\textbf{91.7\textpm0.6}} & \revision{\textbf{82.4\textpm0.7}} & \revision{\textbf{74.9\textpm0.8}} & \revision{\textbf{67.8\textpm0.8}} & \revision{\textbf{77.7\textpm0.7}} & \revision{\textbf{46.9\textpm1.0}} & \revision{\textbf{90.7\textpm0.5}} \\
    \end{tabular}
    \caption{\revision{Performance of root Riemannian ablation vs. squared Mahalanobis on in-domain Meta-Dataset tasks. Error intervals indicate 95\% confidence intervals, and bold values indicate statistically significant state of the art performance. *Note that the ``Root Riemannian'' ablation was trained with gradient clipping (L2-norm capped at 5.0) to become numerically stable during training.}}
    \label{tab:results:squared-mahalanobis-vs-root-riemannian-1}
\end{table*}{}

\begin{table*}[t]
    \centering
    \scriptsize
    \tabcolsep=0.225cm
    \begin{tabular}{lcccccccccccccccccc}
        & \multicolumn{5}{c}{Out-of-Domain Accuracy (\%)} & \multicolumn{3}{c}{Avg Acc.} \\
        \cmidrule(lr){2-6}
        Simple CNAPS + Metric & \rotatebox{45}{Signs} & \rotatebox{45}{MSCOCO} & \rotatebox{45}{MNIST} & \rotatebox{45}{CIFAR10} & \rotatebox{45}{CIFAR100} & In & Out & All \\
        \midrule
        \revision{Root Riemannian*} & \revision{60.4\textpm1.1} & \revision{31.1\textpm1.1} & \revision{94.1\textpm0.4} & \revision{62.4\textpm0.8} & \revision{54.7\textpm1.0} & \revision{68.2} & \revision{60.5} & \revision{65.2} \\
        \revision{Squared Mahalanobis} & \revision{\textbf{73.5\textpm0.7}} & \revision{\textbf{46.2\textpm1.1}} & \revision{\textbf{93.9\textpm0.4}} & \revision{\textbf{74.3\textpm0.7}} & \revision{\textbf{60.5\textpm1.0}} & \revision{\textbf{73.8}} & \revision{\textbf{69.7}} & \revision{\textbf{72.2}} \\
    \end{tabular}
    \caption{\revision{Performance of root Riemannian ablation vs. squared Mahalanobis on out-of-domain Meta-Dataset tasks and MNIST, CIFAR10 and CIFAR100 datasets. Error intervals indicate 95\% confidence intervals, and bold values indicate statistically significant state of the art performance. *Note that the ``Root Riemannian'' ablation was trained with gradient clipping (L2-norm capped at 5.0) to become numerically stable during training.}}
    \label{tab:results:squared-mahalanobis-vs-root-riemannian-2}
\end{table*}{}

\begin{table*}
    \centering
    \scriptsize
    \tabcolsep=0.15cm
    \begin{tabular}{lcccccccccccccccc}
        & \multicolumn{8}{c}{In-Domain Accuracy (\%)} \\
        \cmidrule(lr){2-9}
        Model & \rotatebox{45}{ImageNet} & \rotatebox{45}{Omniglot} & \rotatebox{45}{Aircraft} & \rotatebox{45}{Birds} & \rotatebox{45}{DTD} & \rotatebox{45}{QuickDraw} & \rotatebox{45}{Fungi} & \rotatebox{45}{Flower} \\
        \midrule
        Simple CNAPS & \textbf{58.6\textpm1.1} & 91.7\textpm0.6 & 82.4\textpm0.7 & 74.9\textpm0.8 & \textbf{67.8\textpm0.8} & \textbf{77.7\textpm0.7} & 46.9\textpm1.0 & 90.7\textpm0.5 \\
        \midrule
        \revision{No Refinements} & \revision{\textbf{57.3\textpm1.1}} & \revision{90.5\textpm0.7} & \revision{82.9\textpm0.7} & \revision{74.8\textpm0.8} & \revision{\textbf{67.3\textpm0.8}} & \revision{76.3\textpm0.8} & \revision{47.7\textpm1.0} & \revision{90.5\textpm0.5} \\
        \revision{No Min/Max} & \revision{\textbf{58.7\textpm1.1}} & \revision{\textbf{94.0\textpm0.4}} & \revision{\textbf{84.0\textpm0.6}} & \revision{\textbf{76.4\textpm0.8}} & \revision{\textbf{68.9\textpm0.8}} & \revision{\textbf{77.9\textpm0.7}} & \revision{48.0\textpm1.0} & \revision{91.6\textpm0.5} \\
        \revision{Min 2 Max 4} & \revision{\textbf{58.8\textpm1.1}} & \revision{\textbf{93.9\textpm0.4}} & \revision{\textbf{84.1\textpm0.6}} & \revision{\textbf{76.8\textpm0.8}} & \revision{\textbf{69.0\textpm0.8}} & \revision{\textbf{78.6\textpm0.7}} & \revision{48.8\textpm1.1} & \revision{\textbf{91.6\textpm0.4}} \\
    \end{tabular}
    \caption{\revision{Evaluating min/max refinement restrictions in Transductive CNAPS on in-domain Meta-Dataset sub-datasets.}}
    \label{results:max-min-ablation-1}
\end{table*}{}

\begin{table*}
    \centering
    \scriptsize
    \tabcolsep=0.25cm
    \begin{tabular}{lcccccccccccccccc}
        & \multicolumn{5}{c}{Out-of-Domain Accuracy (\%)} & \multicolumn{3}{c}{Avg Accuracy} \\
        \cmidrule(lr){2-6}
        Model & \rotatebox{45}{Signs} & \rotatebox{45}{MSCOCO} & \rotatebox{45}{MNIST} & \rotatebox{45}{CIFAR10} & \rotatebox{45}{CIFAR100} & In & Out & All \\
        \midrule
        Simple CNAPS & 73.5\textpm0.7 & 46.2\textpm1.1 & 93.9\textpm0.4 & 74.3\textpm0.7 & 60.5\textpm1.0 & 73.8 & 69.7 & 72.2 \\
        \midrule
        \revision{No Refinements} & \revision{\textbf{75.8\textpm0.7}} & \revision{\textbf{47.1\textpm1.1}} & \revision{94.9\textpm0.4} & \revision{74.3\textpm0.8} & \revision{\textbf{61.2\textpm1.0}} & \revision{73.4} & \revision{70.7} & \revision{72.4} \\
        \revision{No Min/Max} & \revision{74.0\textpm0.8} & \revision{\textbf{48.3\textpm1.0}} & \revision{95.7\textpm0.3} & \revision{75.5\textpm0.7} & \revision{61.3\textpm1.0} & \revision{74.9} & \revision{71.0} & \revision{73.4} \\
        \revision{Min 2 Max 4} & \revision{\textbf{76.1\textpm0.7}} & \revision{\textbf{48.7\textpm1.0}} & \revision{\textbf{95.7\textpm0.3}} & \revision{\textbf{75.7\textpm0.7}} & \revision{\textbf{62.9\textpm1.0}} & \revision{\textbf{75.2}} & \revision{\textbf{71.8}} & \revision{\textbf{73.9}} \\
    \end{tabular}
    \caption{\revision{Evaluating min/max refinement restrictions in Transductive CNAPS on out-of-domain Meta-Dataset sub-datasets, MNIST, and CIFAR10/100.}}
    \label{results:max-min-ablation-2}
\end{table*}{}

\subsection{Meta-Dataset Training/Testing}
\label{appendix:benchmark-and-training:meta-dataset-training-and-testing}

Following \cite{bateni2019improved} and \cite{Requeima19_CNAPS}, we train our ResNet18 feature extractor as a supervised multi-class image classifier on the training split of the ImageNet subset of the Meta-Dataset. We directly follow the procedure described in C.1.1 of \cite{Requeima19_CNAPS} and use their released checkpoint. Images from the 712 ImageNet classes designated for training by Meta-Dataset \cite{triantafillou2019meta} are first resized to 84x84. The ResNet18 in Transductive CNAPS is then trained as a 712-class image recognition task using this data. Training is done using the cross-entropy loss for 125 epochs using SGD with momentum of 0.9, weight decay of 0.0001, batch size of 256, and a learning rate of 0.1 that is reduced by a factor of 10 every 25 epochs. During training, the dataset was augmented with random crops, horizontal flips, and color jitter.

Once the ResNet18 is trained, we freeze the parameters and proceed to train the adaptation network using Episodic training \cite{Snell17_Proto, finn2017model} where tasks themselves are used as training examples.
For each iteration of Episodic training, a task (with additional ground truth query labels) is generated, and the adaptation network is trained to minimize classification error (cross entropy) of the query set given the task.

We train for a total of 110K tasks, with 16 tasks per batch, resulting in 6875 gradient updates. We train using Adam optimizer with learning rate of $5 \times 10^{-4}$. We evaluate on the validation splits of all 8 in-domain and 1 out-of-domain, namely MSCOCO, datasets, saving the best performing checkpoint for test-time evaluation.

\subsection{mini/tiered-ImageNet Training/Testing}
\label{appendix:benchmark-and-training:minitiered-training-and-testing}
Similar to Meta-Dataset, we first train the ResNet18 feature extractor. This is done with respect to two settings: first, we directly use the training data from mini-ImageNet and tiered-ImageNet, with 38,400 and 448,695 images respectively. Second, we consider a larger training set of 825 classes from ImageNet \cite{russakovsky2015imagenet} that don't overlap with the test sets of the benchmarks. In both cases, we follow the same procedure as the one described for Meta-Dataset in \ref{appendix:benchmark-and-training:meta-dataset-training-and-testing} with the exception of training for 90 epochs and reducing the learning rate every 30 epochs. 

After training the ResNet18, the weights are frozen while we train the adaptation network using Episodic training: at each iteration, a task is generated, and we backpropagate the query set classification loss through the adaptation network. For mini/tiered-ImageNet, we train for a total of 20K tasks, validating performance every 2K tasks and saving the best checkpoint for test-time evaluation. We, similarly, use the Adam optimiser with learning rate of $5 \times 10^{-4}$, and use a batch size of 16, for a total of 1250 gradient steps.

\begin{figure}[t]
    \centering
    \subfloat[\revision{5-way 1-shot}]{\label{fig:query-num-vs-performance}{\includegraphics[width=0.65\textwidth]{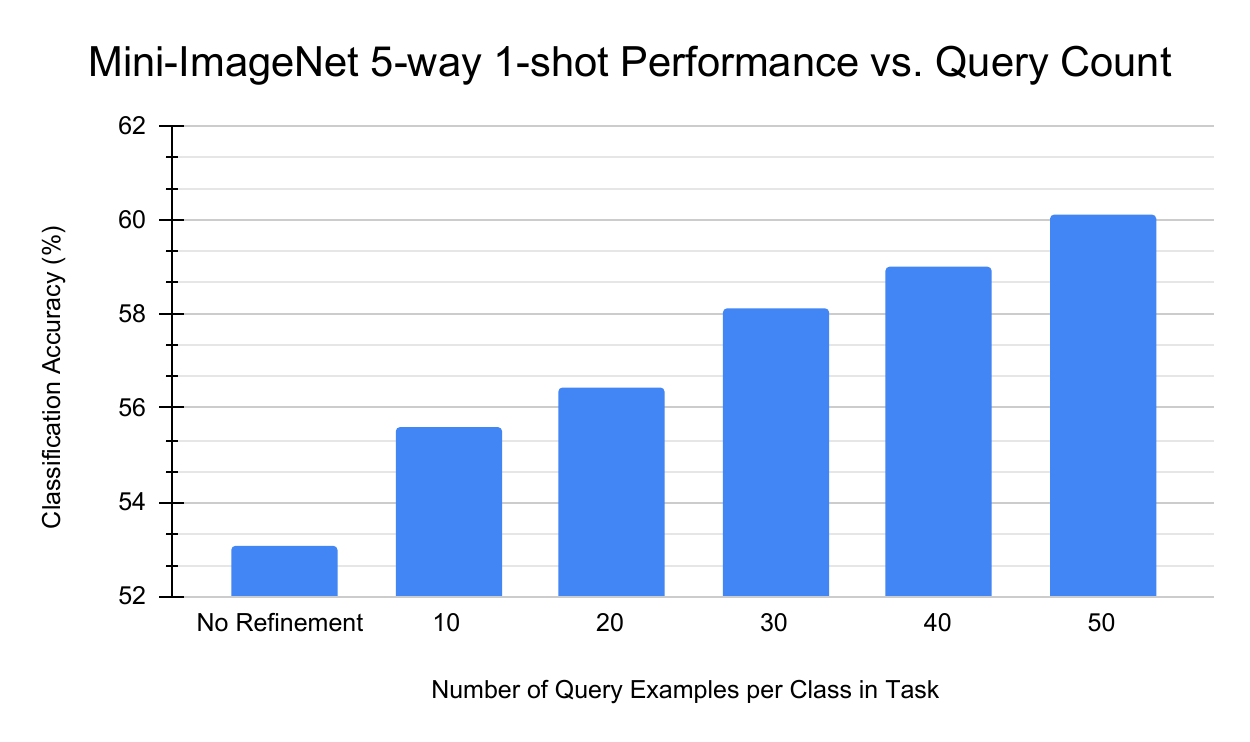}}}\\
    \subfloat[\revision{5-way 5-shot}]{\label{fig:query-num-vs-performance}{\includegraphics[width=0.65\textwidth]{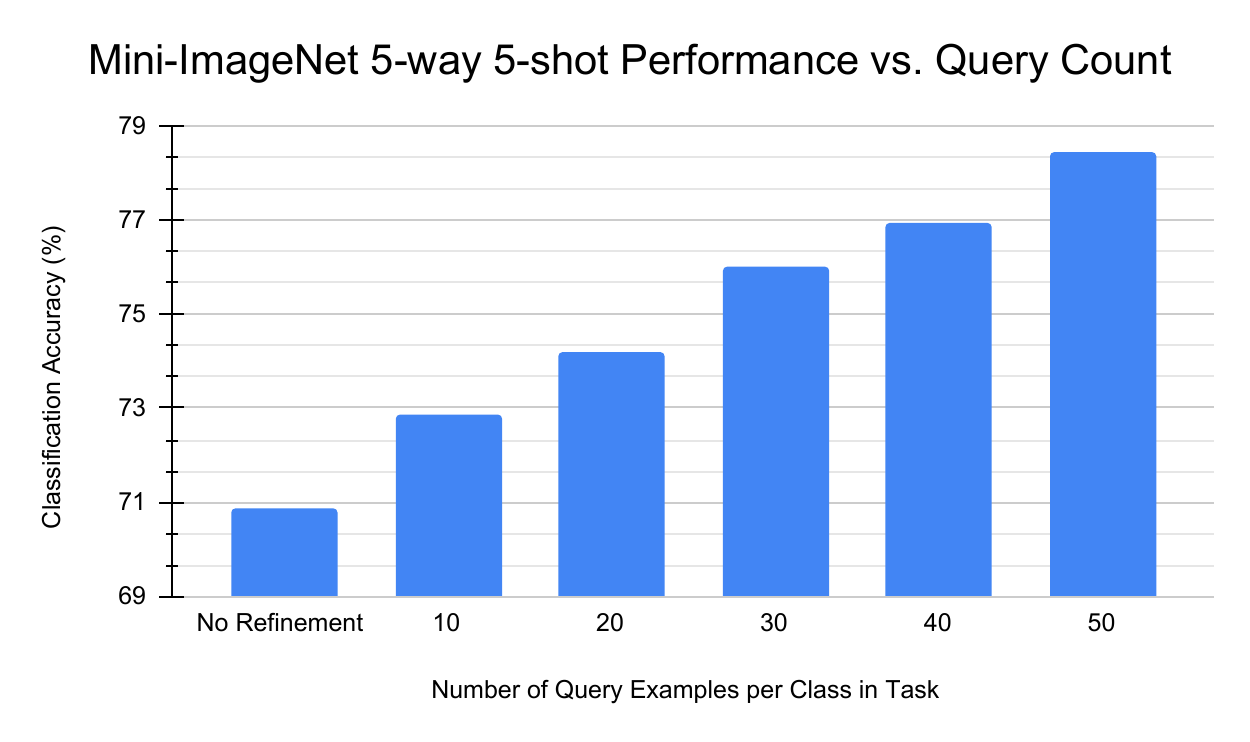}}}%
    \caption{\revision{Mini-ImageNet performance as the number of query examples per class increases on 1/5-shot 5-way tasks.}}
    \label{fig:query-num-vs-performance}
\end{figure}

\revision{
\section{Additional Experimental Results}}

\begin{figure*}
    \centering
    \subfloat[Atlantean]{{\includegraphics[width=0.24\textwidth]{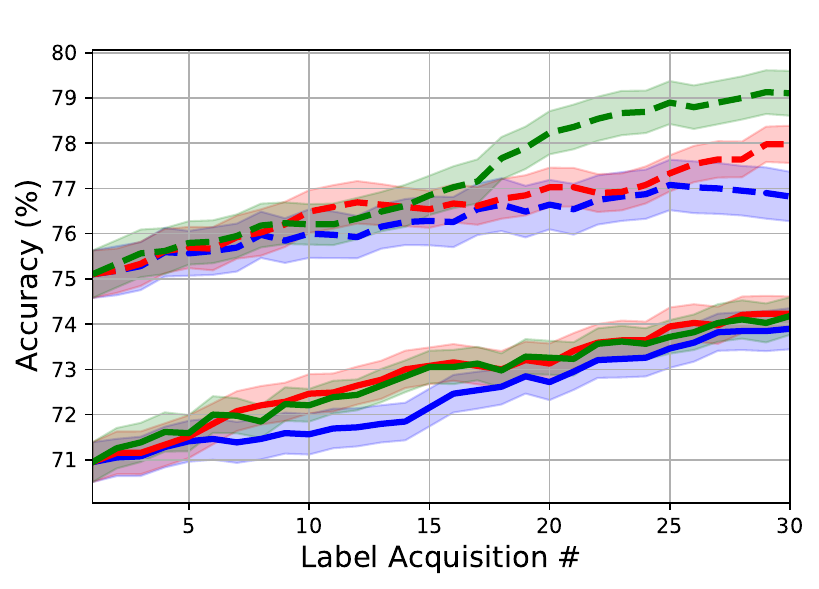}}}
    \subfloat[\revision{Atlantean (FFE)}]{{\includegraphics[width=0.24\textwidth]{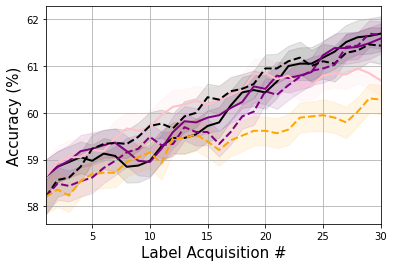}}}
    \subfloat[Aurek-Besh]{{\includegraphics[width=0.24\textwidth]{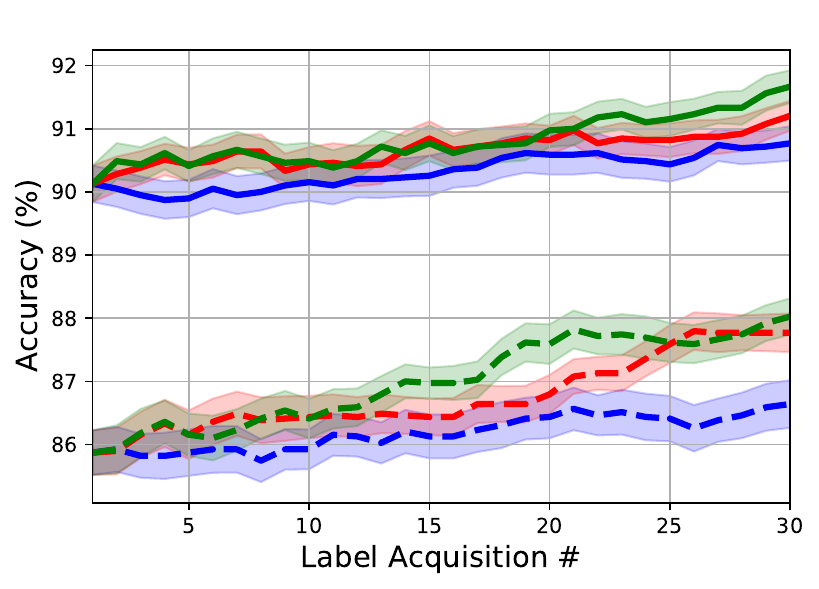}}}
    \subfloat[\revision{Aurek-Besh (FFE)}]{{\includegraphics[width=0.24\textwidth]{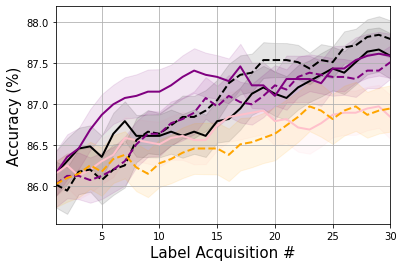}}}
    \\
    \vspace{-0.1in}
    \subfloat[Avesta]{{\includegraphics[width=0.24\textwidth]{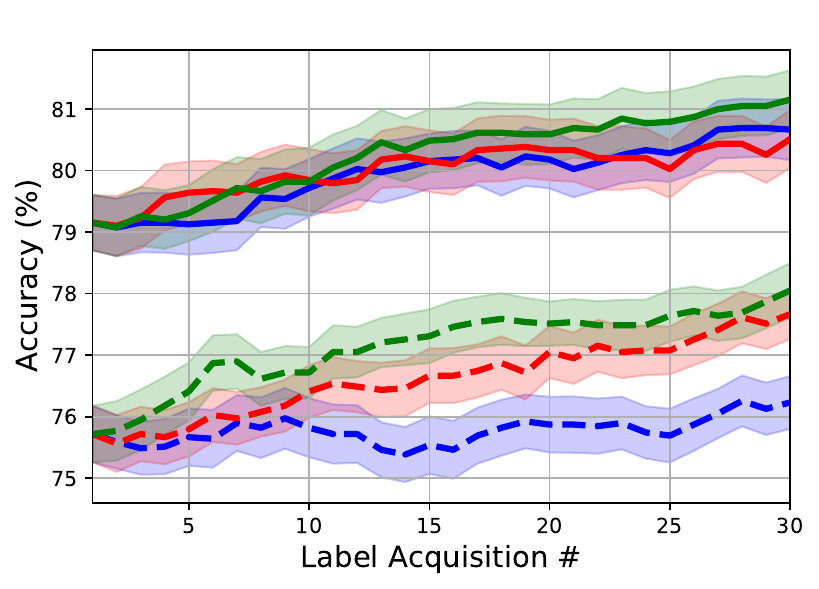}}}
    \subfloat[\revision{Avesta (FFE)}]{{\includegraphics[width=0.24\textwidth]{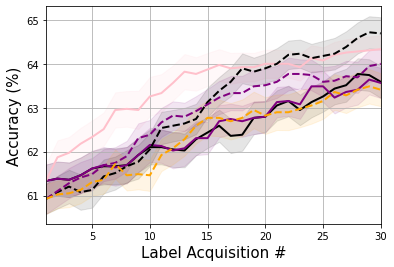}}}
    \subfloat[Ge Ez]{{\includegraphics[width=0.24\textwidth]{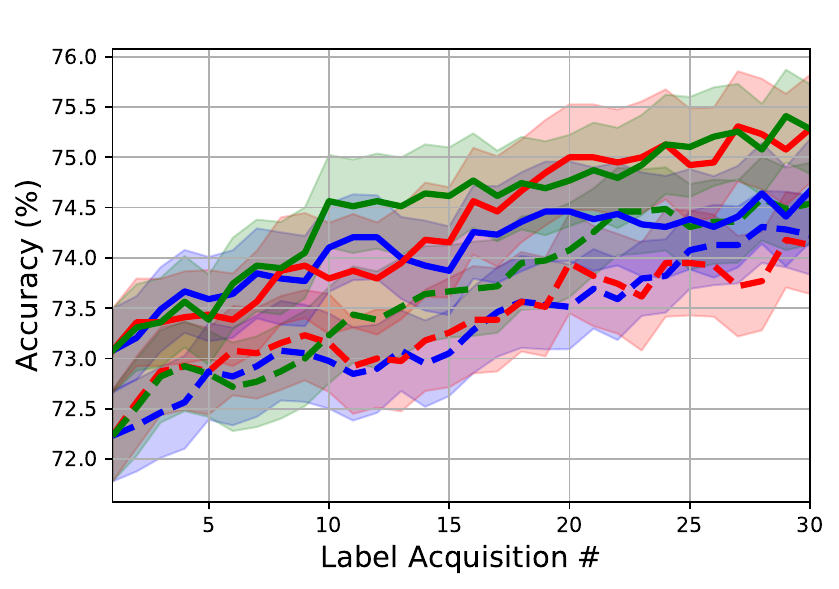}}}
    \subfloat[\revision{Ge Ez (FFE)}]{{\includegraphics[width=0.24\textwidth]{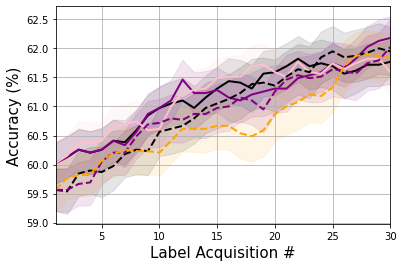}}}
    \\
    \vspace{-0.1in}
    \subfloat[Glagolitic]{{\includegraphics[width=0.24\textwidth]{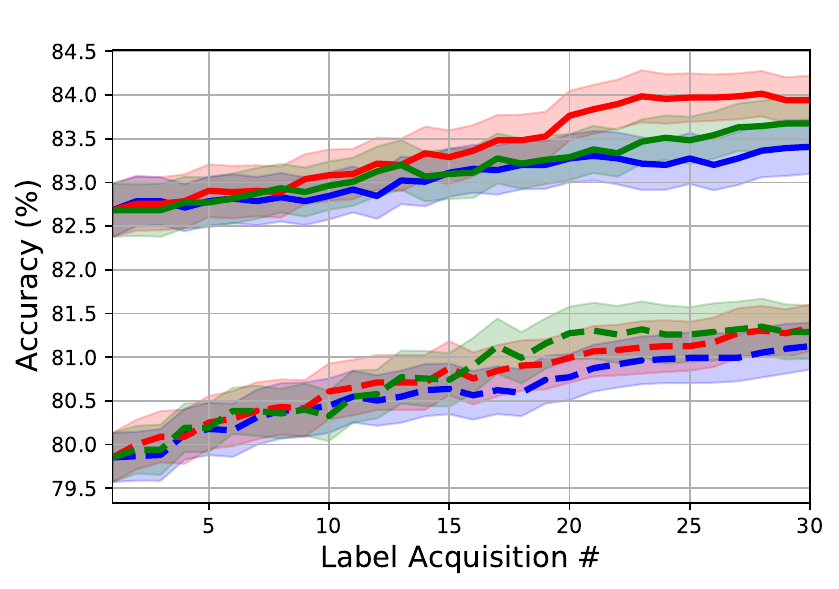}}}
    \subfloat[\revision{Glagolitic (FFE)}]{{\includegraphics[width=0.24\textwidth]{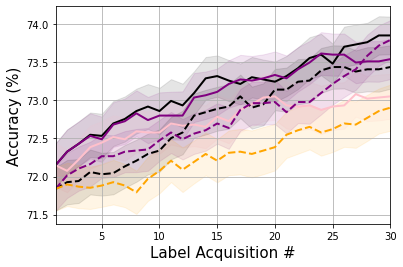}}}
    \subfloat[Gurmukhi]{{\includegraphics[width=0.24\textwidth]{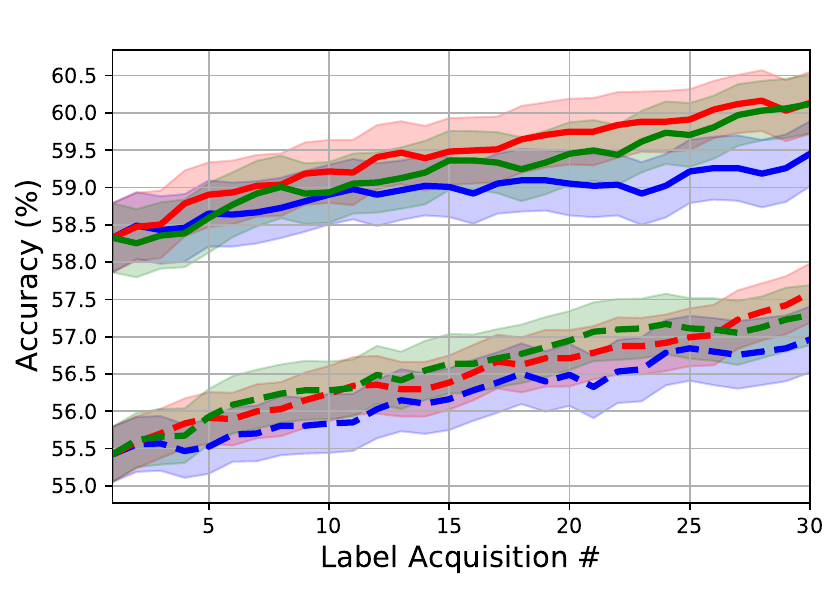}}}\subfloat[\revision{Gurmukhi (FFE)}]{{\includegraphics[width=0.24\textwidth]{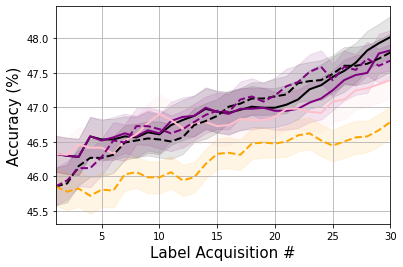}}}
    \\
    \vspace{-0.1in}
    \subfloat[Kannada]{{\includegraphics[width=0.24\textwidth]{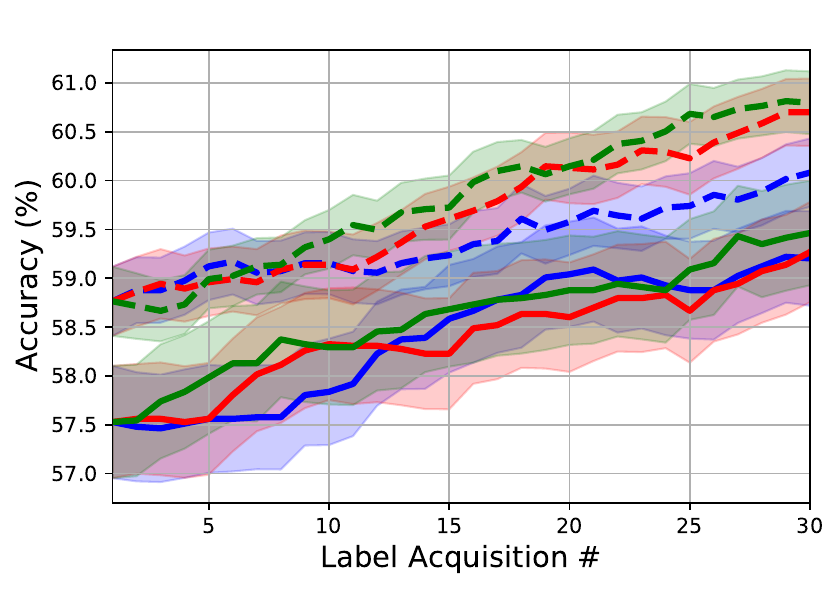}}}
    \subfloat[\revision{Kannada (FFE)}]{{\includegraphics[width=0.24\textwidth]{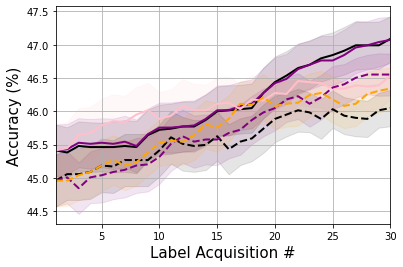}}}
    \subfloat[Keble]{{\includegraphics[width=0.24\textwidth]{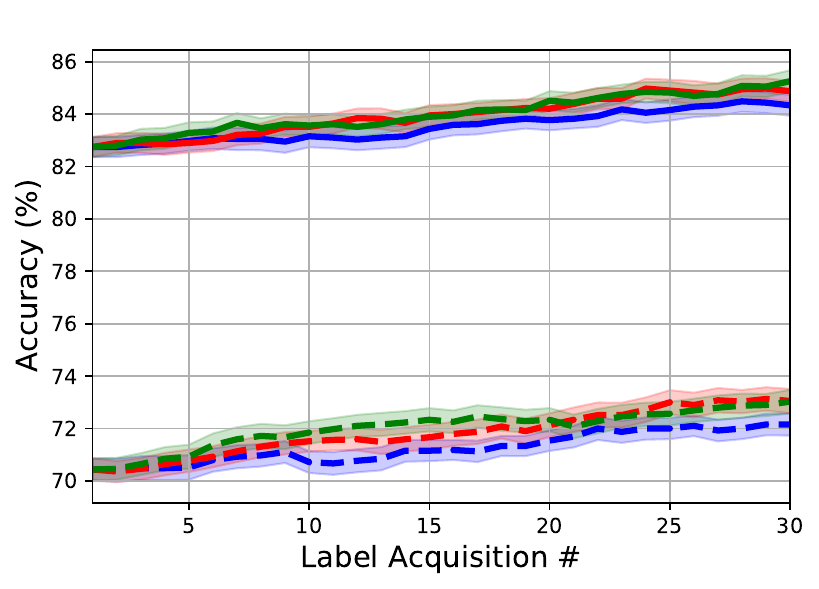}}}
    \subfloat[\revision{Keble (FFE)}]{{\includegraphics[width=0.24\textwidth]{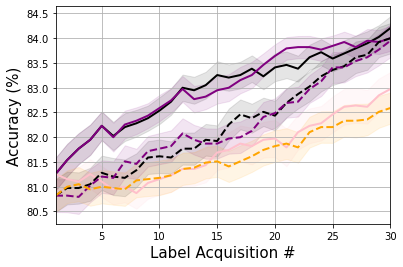}}}
    \\
    \vspace{-0.1in}
    \subfloat[Malayalam]{{\includegraphics[width=0.24\textwidth]{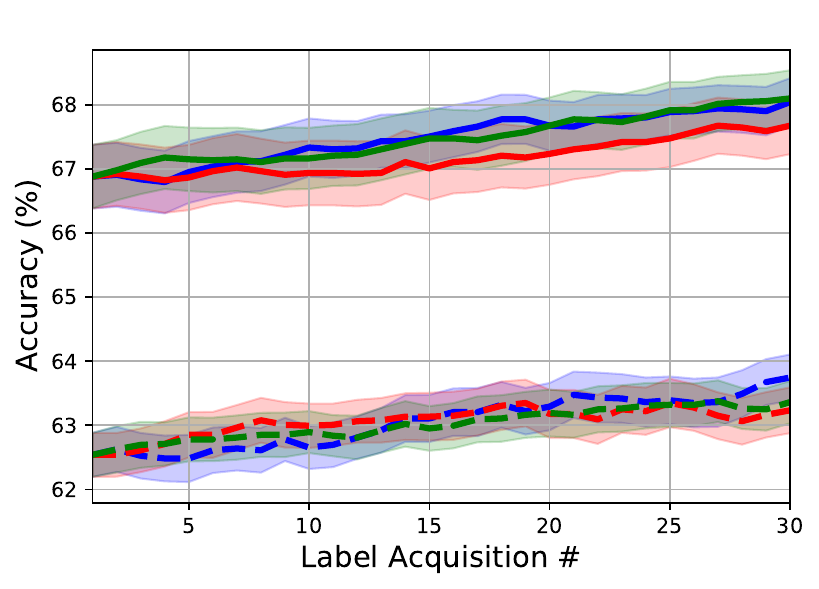}}}
    \subfloat[\revision{Malayalam (FFE)}]{{\includegraphics[width=0.24\textwidth]{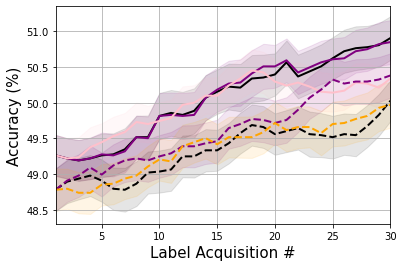}}}
    \subfloat[Manipuri]{{\includegraphics[width=0.24\textwidth]{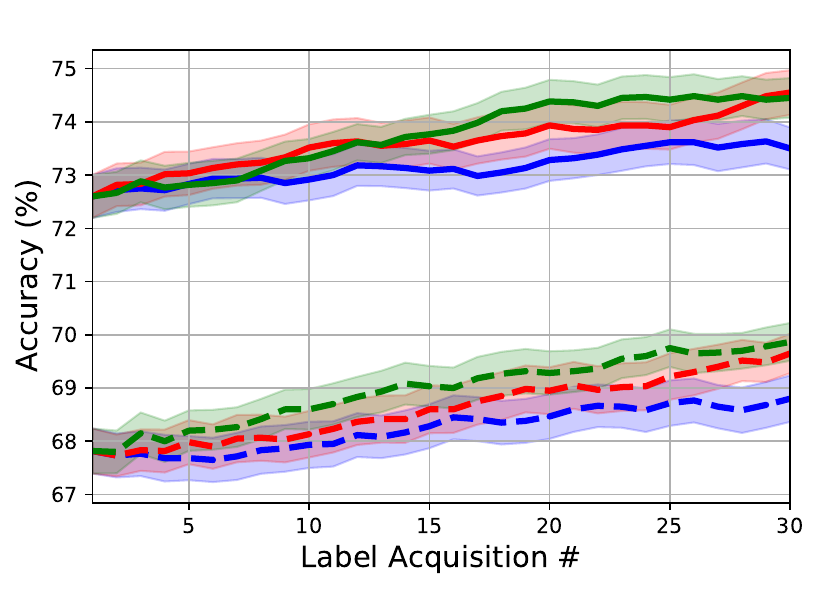}}}
    \subfloat[\revision{Manipuri (FFE)}]{{\includegraphics[width=0.24\textwidth]{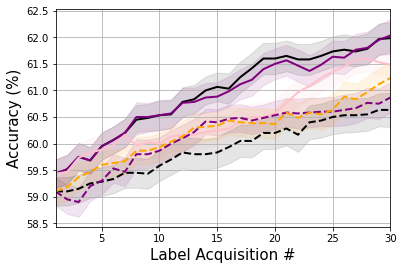}}}
    \\
    \vspace{-0.1in}
    \subfloat[Mongolian]{{\includegraphics[width=0.24\textwidth]{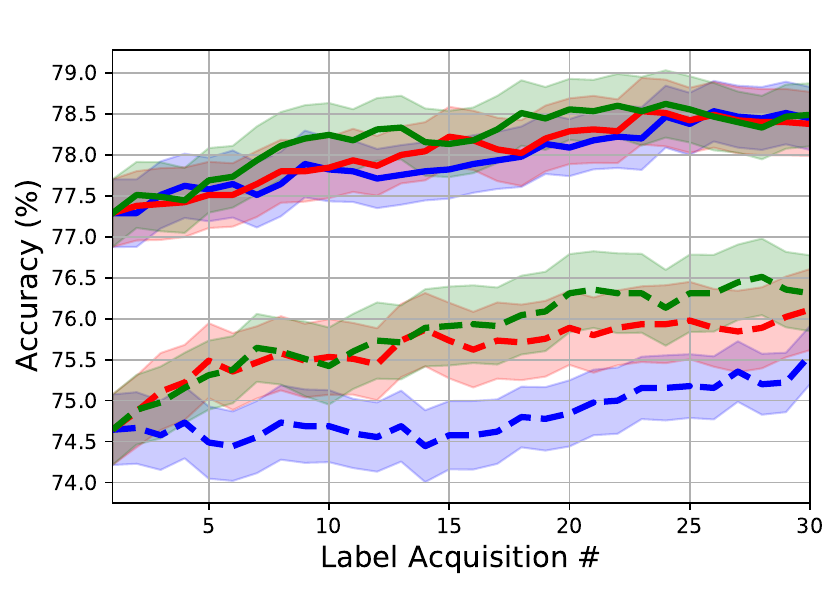}}}
    \subfloat[\revision{Mongolian (FFE)}]{{\includegraphics[width=0.24\textwidth]{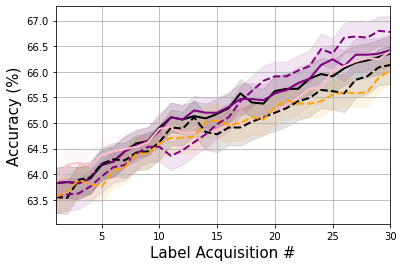}}}
    \subfloat[Old Church]{{\includegraphics[width=0.24\textwidth]{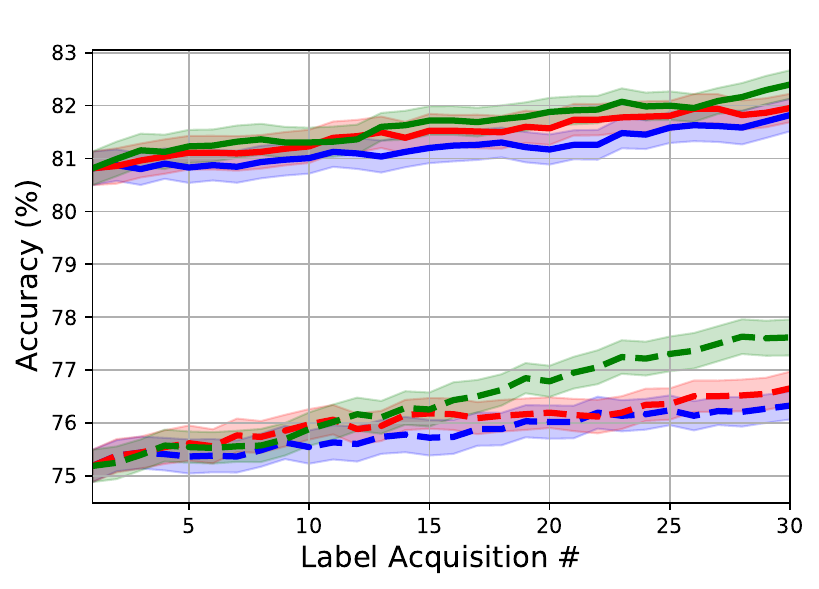}}}
    \subfloat[\revision{Old Church (FFE)}]{{\includegraphics[width=0.24\textwidth]{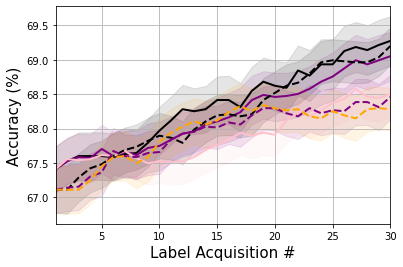}}}
    \\
    \caption{``Out of the box'' active learning results on 12 of the remaining OMNIGLOT languages.}
    \label{fig:active-learning:additional-results-first-batch}
\end{figure*}

\revision{\subsection{Evaluation on mini/tiered-ImageNet}}

\revision{Few-shot image classification results with error intervals on mini-ImageNet and tiered-ImageNet in Table \ref{tab:results:mini-tiered-imagenet-with-error-intervals}. Classification results with error intervals when additional data is present on both mini- and tiered-ImageNet benchmarks is shown in Table \ref{tab:results:mini-tiered-imagenet-with-error-intervals-feti}.}

\revision{\subsection{Active Learning}}

\begin{figure*}[t]
    \centering
    \subfloat[Oriya]{{\includegraphics[width=0.24\textwidth]{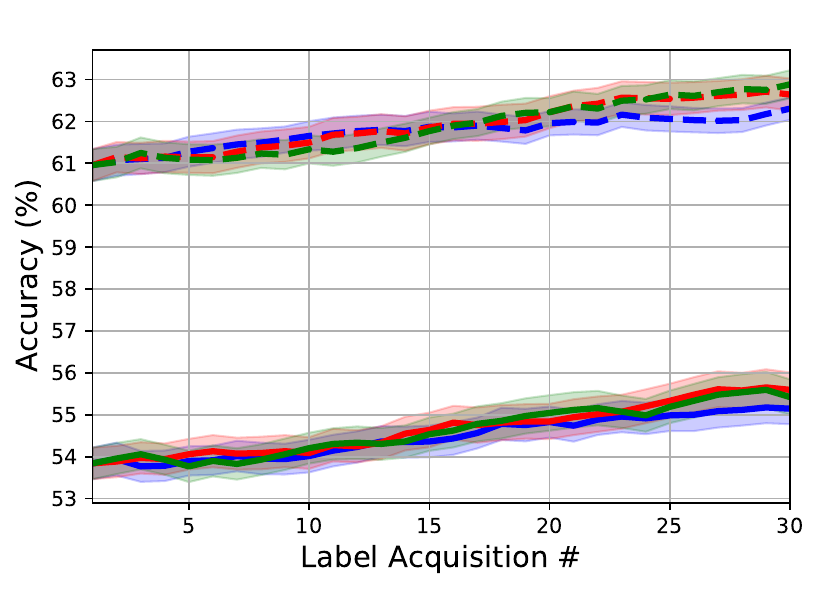}}}
    \subfloat[\revision{Oriya (FFE)}]{{\includegraphics[width=0.24\textwidth]{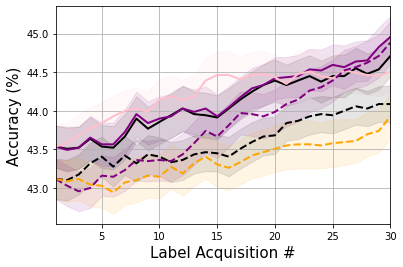}}}
    \subfloat[Sylheti]{{\includegraphics[width=0.24\textwidth]{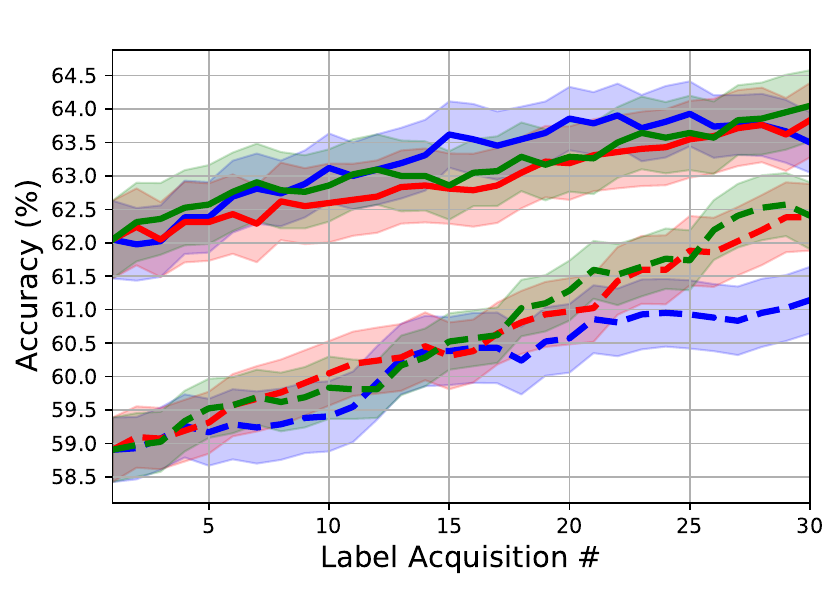}}}
    \subfloat[\revision{Sylheti (FFE)}]{{\includegraphics[width=0.24\textwidth]{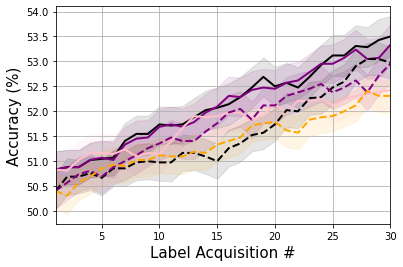}}}
    \\
    \vspace{-0.1in}
    \subfloat[Syriac (Serto)]{{\includegraphics[width=0.24\textwidth]{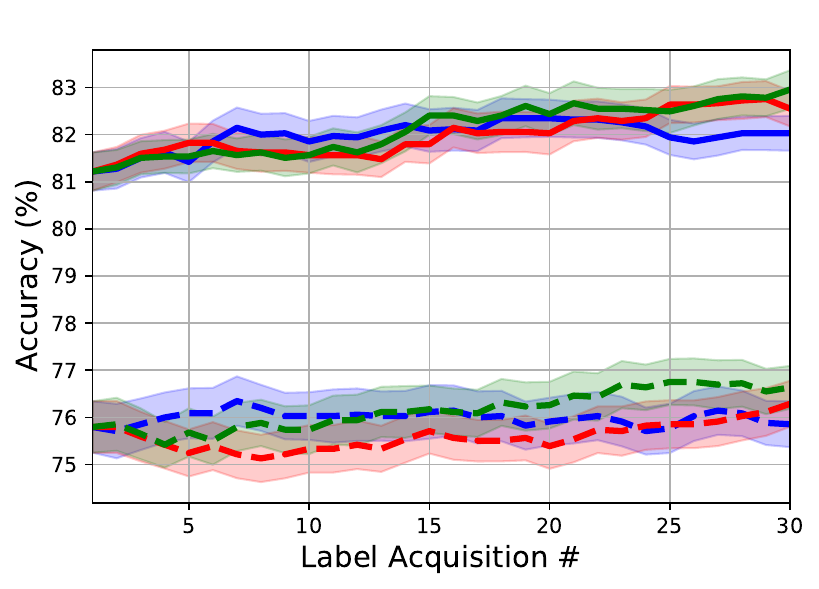}}}
    \subfloat[\revision{Syriac (Serto) (FFE)}]{{\includegraphics[width=0.24\textwidth]{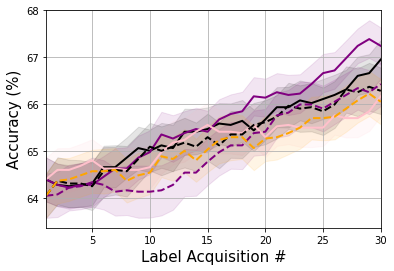}}}
    \subfloat[Tengwar]{{\includegraphics[width=0.24\textwidth]{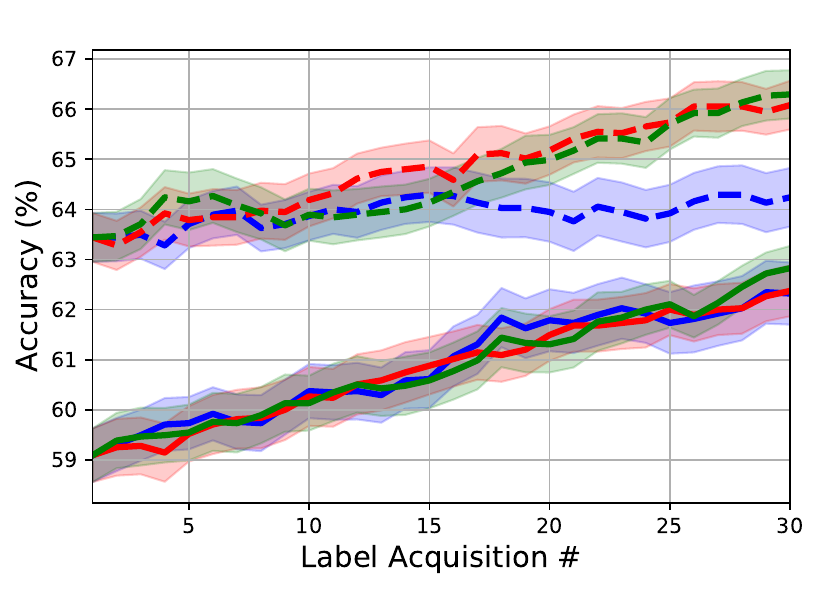}}}
    \subfloat[\revision{Tengwar (FFE)}]{{\includegraphics[width=0.24\textwidth]{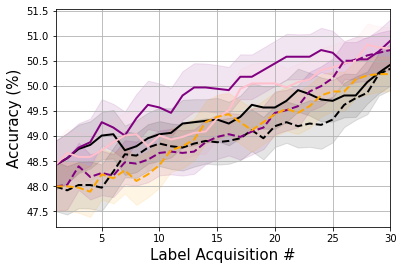}}}
    \\
    \vspace{-0.1in}
    \subfloat[Tibetan]{{\includegraphics[width=0.24\textwidth]{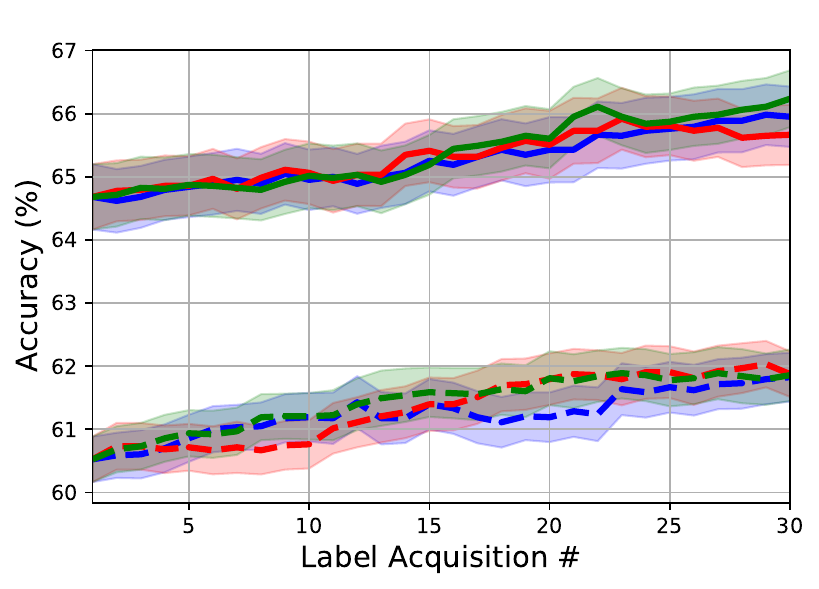}}}
    \subfloat[\revision{Tibetan (FFE)}]{{\includegraphics[width=0.24\textwidth]{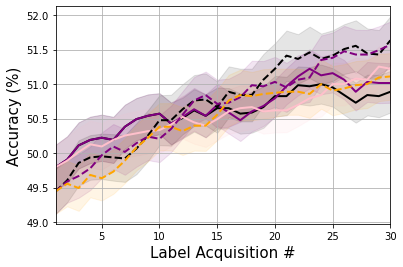}}}
    \subfloat[ULOG]{{\includegraphics[width=0.24\textwidth]{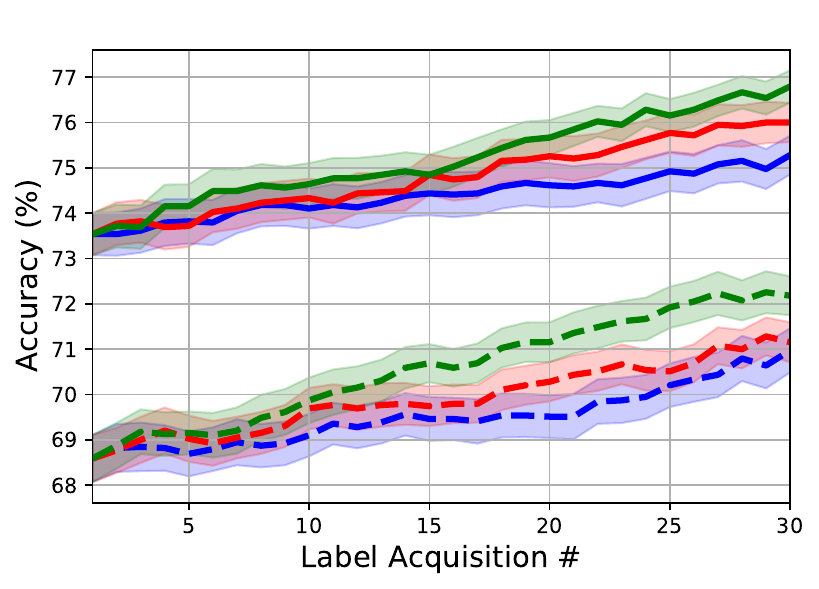}}}
    \subfloat[\revision{ULOG (FFE)}]{{\includegraphics[width=0.24\textwidth]{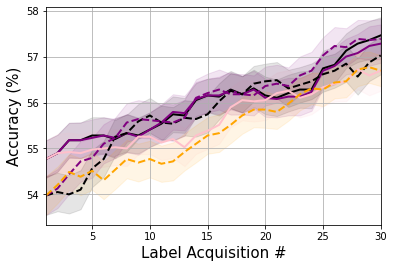}}}
    \\
    \caption{``Out of the box'' active learning results on the last 6 remaining OMNIGLOT languages.}
    \label{fig:active-learning:additional-results-second-batch}
\end{figure*}

\revision{``Out of the box'' active learning results on additional OMNIGLOT language have been provided in Figures \ref{fig:active-learning:additional-results-first-batch} and \ref{fig:active-learning:additional-results-second-batch}.}

\revision{\subsection{Squared Mahalanobis vs. Root Riemannian}}

\revision{Given the theoretical grounding of our methods in Riemannian metric learning, the root Riemannian distance is a natural baseline to consider. That is, instead of producing class probabilities using Equation \ref{simple-cnaps-eq:simple-cnaps-mahalanobis-classifier}, we employ the Mahalanobis distance itself as oppose to the squared Mahalanobis distance, such that}

\revision{\begin{equation}
    p(y^* = k \mid \mathbf z^*) \propto \exp
    \big(
      -\sqrt{(\mathbf z^* - \bm{\mu}_k)^T\mathbf{Q}_k^{-1}(\mathbf z^* - \bm{\mu}_k)}
    \big).
    \label{root-riemannian-equation}
\end{equation}}

\revision{At test time, given the same model checkpoints, both variations perform the same as the decision boundaries remain intact because they are not impacted by the square root function. The main difference observed is during training where the square root function results in a differing optimization manifold, and by consequence, gradient decent.}

\revision{Extensive experiments were conducted exploring this and the root Riemannian variation was observed to be incredibly unstable numerically. Standard settings for training Simple and Transductive CNAPS resulted in exploding gradients during backpropagation, and reducing step size proved ineffective. Stable runs were only achieved with extreme gradient clipping (L2-norm capped at 5.0). As demonstrated in Tables \ref{tab:results:squared-mahalanobis-vs-root-riemannian-1} and \ref{tab:results:squared-mahalanobis-vs-root-riemannian-2}, this results in significantly worse performance as compared to the squared Mahalanobis distance.}

\revision{\subsection{ImageNet-Only Training on Meta-Dataset}}

\revision{The Meta-Dataset benchmark also proposes an alternative experimental setting where only the ImageNet training split is used for episodic training. All remaining sub-datasets are kept for out-of-domain evaluation. This variation has received comparatively little attention as it lacks the examples diversity provided by the original Meta-Dataset setup during training.}

\revision{We report ImageNet-only training results on the Meta-Dataset in Tables \ref{tab:results:meta-dataset-imagenet-only-1} and \ref{tab:results:meta-dataset-imagenet-only-2}. As shown, Simple and Transductive CNAPS demonstrate competitive performances, ranking second to that of CTX \cite{Doersch2020_CrossTransformers} whilst matching BOHB-E \cite{bohb-e-saikia2020optimized} and ResNet34 Prototypical Networks \cite{Doersch2020_CrossTransformers}. When compared to other methods with ResNet18 backbones, our models match the SoTA performance of BOHB-E.}

\revision{\subsection{Performance vs. Number of Query Examples}}

\revision{Figure \ref{fig:query-num-vs-performance} shows the performance of Transductive CNAPS on 1/5-shot 5-way Mini-ImageNet tasks as the number of query examples per category increases from 10 to 50. As shown, with a greater number of query examples, performance improves as our method is able to exploit more unlabelled data.}

\revision{\subsection{Max and Min Number of Refinements in Transductive CNAPS}}

\revision{In our experiments, we use a minimum number of 2 refinement steps of class parameters, with the maximum set to 4 on the Meta-Dataset and 10 on the mini- and tiered-ImageNet benchmarks. As shown in Tables \ref{results:max-min-ablation-1} and \ref{results:max-min-ablation-2}, the refinement criteria itself, without any step constraints, results in a significant performance gain as compared to performing no refinements. In fact, it accounts for the majority of the accuracy gain for Transductive CNAPS. We further explore the impact of these step-hyperparameters on the performance on Transductive CNAPS on the Meta-Dataset in Figure \ref{fig:max-and-min}.}

\revision{As shown, requiring the same number of refinement steps for every task results in sub-optimal performance. This is demonstrated by the fact that the peak performance for each minimum number of steps is achieved with larger number of maximum steps, showcasing the importance of allowing different numbers of refinement steps depending on the task. In addition, we observe that as the number of minimum refinement steps increases, the performance improves up to two steps while declining after. This suggests that, unlike \cite{DBLP:journals/corr/abs-1803-00676-tieredimagenet} where only a single refinement step leads to the best performance, our Mahalanobis-based approach can leverage extra steps to further refining the class parameters. We do see a decline in performance with a higher number of steps; this suggests that while our refinement criteria can be effective at performing different number of steps depending on the task, it can potentially lead to over-fitting, justifying the need for a well chosen maximum number of steps.}

\newpage

\bibliographystyle{elsarticle-num} 
\bibliography{main}

\end{document}